\documentclass[]{onethree}

\usepackage[toc,page,header]{appendix}
\usepackage{minitoc}
\usepackage{amsmath,amssymb}
\usepackage{bm}
\usepackage{wrapfig}
\usepackage{enumitem}
\usepackage{etoolbox}
\usepackage{pifont}
\usepackage{adjustbox}
\DeclareTextFontCommand{\paperbold}{\rmfamily\bfseries}
\DeclareFontShape{T1}{bytesans}{b}{n}{<->s*[1]assets/bytesans}{}
\DeclareFontShape{T1}{bytesans}{bx}{n}{<->s*[1]assets/bytesans}{}
\hypersetup{
  pdftitle={OmniHarness: Harnessing Generalizable Visual Generation via Symbolic Policy Learning},
  pdfauthor={Xu Xu, Jinxiu Liu, Zhangbo Qiao, Cheng Song, Jiaxing Lu, Xiangyu Zhang, Yubin Gu, Fangwei Ning, Yan Shi}
}

\usepackage{etoc}

\usepackage{algorithm}
\usepackage{algpseudocode}

\usepackage{tabularray}
\UseTblrLibrary{booktabs}
\algrenewcommand\algorithmiccomment[1]{\hfill$\triangleright$~#1}

\definecolor{OmniPromptPink}{HTML}{FAF5F6}
\definecolor{OmniPromptBlue}{HTML}{EEF6F8}

\renewcommand{\beginappendix}{\appendix}

\usepackage{float}
\usepackage{tabularx}
\usepackage{array}
\newcolumntype{Y}{>{\centering\arraybackslash}X}
\usepackage{colortbl}

\definecolor{comfyagentbg}{HTML}{F6F8FA}
\definecolor{comfymindbg}{HTML}{EFF3F7}
\definecolor{symbomnibg}{HTML}{E5EDF5}
\definecolor{catsbg}{HTML}{D8E5F1}
\definecolor{omniharnessbg}{HTML}{C9DDEC}

\newcommand{\bestscore}[1]{{\rmfamily\bfseries #1}}
\newcommand{\secondscore}[1]{\underline{#1}}

\definecolor{gegain}{HTML}{940001}

\newcommand{\gecell}[2]{%
    \mbox{#1\kern0.4pt{%
        \normalfont\fontsize{5.5pt}{7pt}\selectfont #2%
    }}%
}

\newcommand{\geup}[1]{%
    \textcolor{gegain}{\ensuremath{\uparrow}#1}%
}

\definecolor{geloss}{HTML}{006400}

\newcommand{\gedown}[1]{%
    \textcolor{geloss}{\ensuremath{\downarrow}#1}%
}

\title{OmniHarness: Harnessing Generalizable Visual Generation via Symbolic Policy Learning}

\renewcommand{\authorlist}{%
  \begin{tabular}{@{}c@{}}
    \authorformat[1,*]{Xu Xu},
    \authorformat[2]{Jinxiu Liu},
    \authorformat[1]{Zhangbo Qiao},
    \authorformat[3]{Cheng Song},
    \authorformat[1]{Jiaxing Lu}
    \\[3pt]
    \authorformat[1]{Xiangyu Zhang},
    \authorformat[3]{Yubin Gu},
    \authorformat[1]{Fangwei Ning},
    \authorformat[1]{Yan Shi}
  \end{tabular}%
}

\affiliation[1]{Beihang University}
\affiliation[2]{The Chinese University of Hong Kong}
\affiliation[3]{National University of Singapore}

\contribution[*]{%
  Corresponding author:
  Xu Xu (\email{xuxu1@buaa.edu.cn})
}

\abstract{
Unified multimodal large language models and multi-agent systems have advanced visual generation. However, several limitations remain. (1) Experience acquisition is typically reactive, occurring only in response to downstream tasks. (2) Accumulated knowledge varies in relevance and reliability, making efficient and effective reuse challenging. To address these limitations, we introduce \paperbold{OmniHarness}, a framework for generalizable visual generation via \paperbold{\textit{symbolic policy learning}}. Before downstream objectives are specified, self-directed inquiry autonomously generates and executes practice tasks, with selection guided by capability novelty and the current competence frontier. Verified workflows are abstracted into symbolic policies that encode experience for task families. Symbolic retrieval searches workflow metadata in stages rather than full parameterized workflows, using reliability tiers to prioritize reliable knowledge. Execution feedback updates these policies after each task, guiding subsequent retrieval and reuse while model parameters remain fixed. Experiments across six benchmarks demonstrate strong performance and generalizable visual generation capabilities. On ComfyBench's Creative tasks, OmniHarness achieves a 95.0\% resolve rate, exceeding the strongest baseline by 27.5 percentage points.
\vspace{-0.1cm}
}

\checkdata[Resources]{%
  \href{https://omniharness.github.io/}{%
    \raisebox{-1ex}{%
      \includegraphics[height=4.2ex]{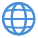}%
    }~Project Page%
  }%
  \hspace{2em}%
  \href{https://github.com/OmniHarness/OmniHarness}{%
    \raisebox{-1ex}{%
      \includegraphics[height=4.2ex]{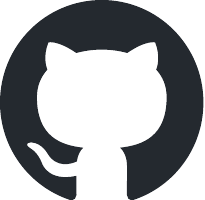}%
    }~Code%
  }%
}

\makeatletter

\newcommand{\OmniHarnessInsertCover}{%
  \par
  \noindent
  \begin{minipage}{\linewidth}
    \centering

    \includegraphics[
      width=\linewidth,
      keepaspectratio
    ]{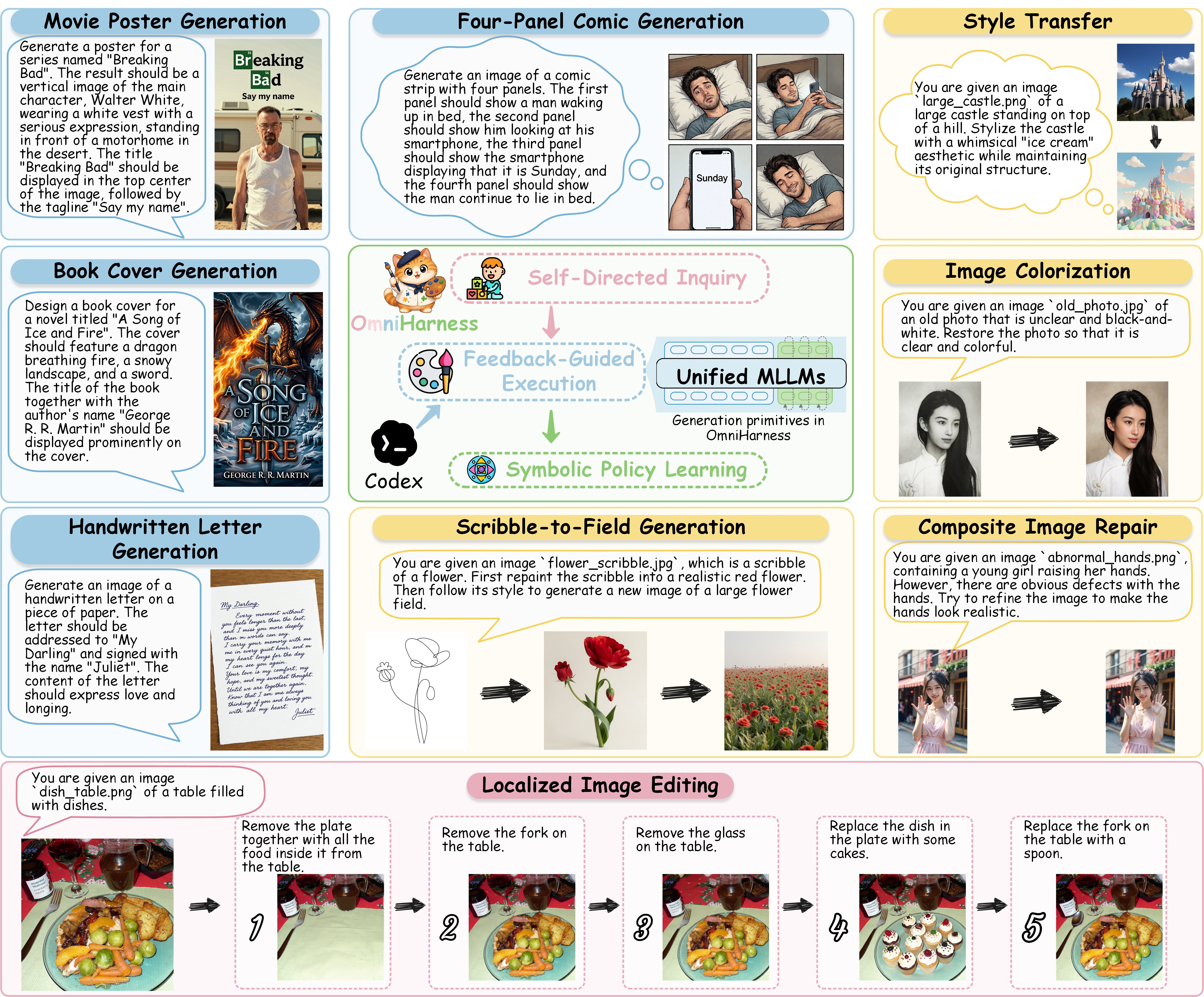}

    \captionof{figure}{Overview of OmniHarness and its diverse visual generation capabilities.}
    \label{fig:1}
  \end{minipage}
  \par

  \thispagestyle{firststyle}
  \clearpage
}

\patchcmd{\mymaketitle}
  {\thispagestyle{firststyle}}
  {}
  {}
  {%
    \PackageError{cover-layout}
      {First-page style not found}
      {Check the title structure in onethree.cls.}%
  }

\patchcmd{\mymaketitle}
  {\tcbset{enhanced,frame hidden}}
  {\OmniHarnessInsertCover\tcbset{enhanced,frame hidden}}
  {}
  {%
    \PackageError{cover-layout}
      {Abstract box not found}
      {Check the abstract-box code in onethree.cls.}%
  }

\makeatother

\begin{document}

\maketitle

\section{Introduction}

\begin{figure}[!htbp]
\centering
\includegraphics[width=\textwidth]{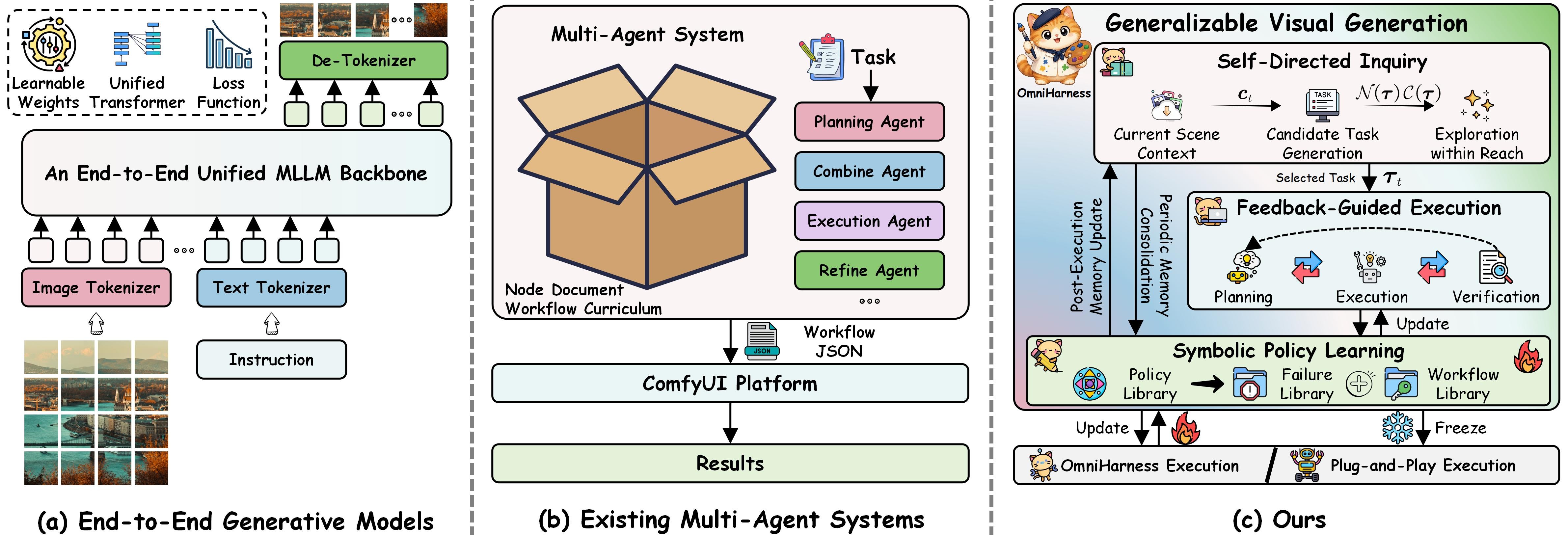}
\caption{Comparison of visual generation paradigms. (a) End-to-end generative models. (b) Existing multi-agent systems for visual generation. (c) OmniHarness.}
\label{fig:2}
\end{figure}

Visual intelligence is advancing along two complementary paths for effective scaling. The first develops end-to-end unified multimodal large language models (MLLMs) for visual perception \citep{llamagen,bagel,got,LWM}, multimodal reasoning \citep{wen2024realtime,zhao-etal-2026-safety,hao-etal-2026-recreate}, and image generation \citep{paper2figure,coDrawAgents,genevolve,PixelCraft}. However, as shown in Figure~\ref{fig:2}(a), this approach relies on large-scale training data. Standalone models offer limited support for explicit verification and self-correction and can struggle with complex reasoning tasks. These limitations motivate the second path, which explores MLLM-based multi-agent systems (MAS) \citep{MAS2,AgenTracer,IEA,IntentEdit}.

As illustrated in Figure~\ref{fig:2}(b), existing MAS for visual generation coordinate agents for planning, workflow composition, execution, and refinement \citep{comfybench,comfygpt}. They convert execution traces into workflows and store them in knowledge bases for reuse \citep{comfyclaw}. Such knowledge is often specific to a particular task, limiting its generalizability. Prior work has also explored symbolic workflow abstraction, experience accumulation, and recovery guided by execution feedback \citep{comfymind,symbomni}. However, these methods treat stored knowledge uniformly, without explicit retrieval priorities or dynamic maintenance during reuse. As knowledge bases expand through continued task execution, retrieval can become more costly, while the lack of prioritization and maintenance can reduce retrieval quality. Moreover, these methods learn from experience only during downstream task execution. Their initial knowledge bases rely on manually composed workflows, whose quality and effectiveness vary.

Addressing these limitations requires considering two questions.
\ding{182}~\textit{Must learning be passive?}
\paperbold{\emph{Seek and you shall find.}}
However, repeated attempts alone do not ensure effective exploration. Motivated by \emph{the investigation of things and the extension of knowledge} \citep{wangyangming}, exploration should take place within the bounds of current capabilities.
\ding{183}~\textit{Is all accumulated knowledge useful?}
\paperbold{\emph{All that glitters is not gold.}}
Each execution can be distilled into experience for reuse, but not all such experience is reliable. Effective reuse therefore requires \emph{separating the wheat from the chaff}.
These challenges motivate a central question:
\paperbold{\textit{How can we build a visual generation system that learns through effective autonomous exploration and selectively applies the knowledge it acquires?}}

To address this central question, we introduce \paperbold{OmniHarness}, a framework for generalizable visual generation via \paperbold{\textit{symbolic policy learning}}.
As shown in Figure~\ref{fig:2}(c), before downstream task execution, OmniHarness uses self-directed inquiry to generate and execute practice tasks that probe its capability limits. It abstracts parameterized workflows from verified executions into symbolic policies that represent experience for families of visual generation tasks. Retrieval operates on these symbolic representations and prioritizes relevant, reliable experience. Execution feedback updates the policies after each task to guide subsequent retrieval and workflow construction. This combination of symbolic abstraction and continual updates improves retrieval efficiency and quality, supporting effective reuse across related tasks.

Our contributions are summarized as follows:
\begin{itemize}
    \item \paperbold{Self-Directed Inquiry.}
    We introduce self-directed inquiry that jointly considers capability novelty and the current competence frontier when selecting practice tasks. These criteria balance exploration of underexplored capabilities with estimated task learnability, enabling effective exploration.
    \item \paperbold{Symbolic Policy Learning.}
    We abstract parameterized workflows into symbolic policies that encode experience for task families. Symbolic retrieval searches workflow metadata in stages rather than full workflows, prioritizing policies according to their reliability tiers. Execution feedback updates these policies after each task to guide subsequent execution. This design enables more efficient and effective knowledge reuse.
    \item \paperbold{Experimental Evaluation.}
    Experiments across six benchmarks demonstrate strong performance, continual capability expansion, and transfer across visual agent frameworks.
    On ComfyBench's Creative tasks, OmniHarness achieves a 95.0\% resolve rate, exceeding the state-of-the-art baseline by 27.5 percentage points.
\end{itemize}

\section{Related Work}

\paperbold{End-to-End Unified MLLMs.} Unified MLLMs integrate visual understanding and generation within a single architecture \citep{emu3,seedx,Show-o2}. Recent work advances reasoning through cross-modal chain-of-thought \citep{MM-R1,UniT}, multi-representation mutual reinforcement \citep{su2026generationenhancesunderstandingunified}, and shared-context visual tokenization \citep{peng2026unifiedmultimodalautoregressivemodeling}. However, standard inference offers limited support for explicit verification, failure recovery, and persistent workflow reuse.

\paperbold{Agentic Systems.}
ComfyBench evaluates autonomous workflow construction in ComfyUI \citep{comfybench}, while ComfyGPT coordinates specialized agents to generate and refine workflows from task instructions \citep{comfygpt}. ComfyMind combines semantic workflow interfaces with search tree planning and local execution feedback \citep{comfymind}. SymbOmni abstracts execution experience into reusable symbolic concepts that combine semantic descriptions with parameterized workflow instructions \citep{symbomni}. These methods primarily construct workflows and accumulate experience in response to supplied tasks. Proactive experience acquisition and effective knowledge reuse remain important challenges.

\begin{figure}[!t]
    \centering
    \includegraphics[width=\textwidth]{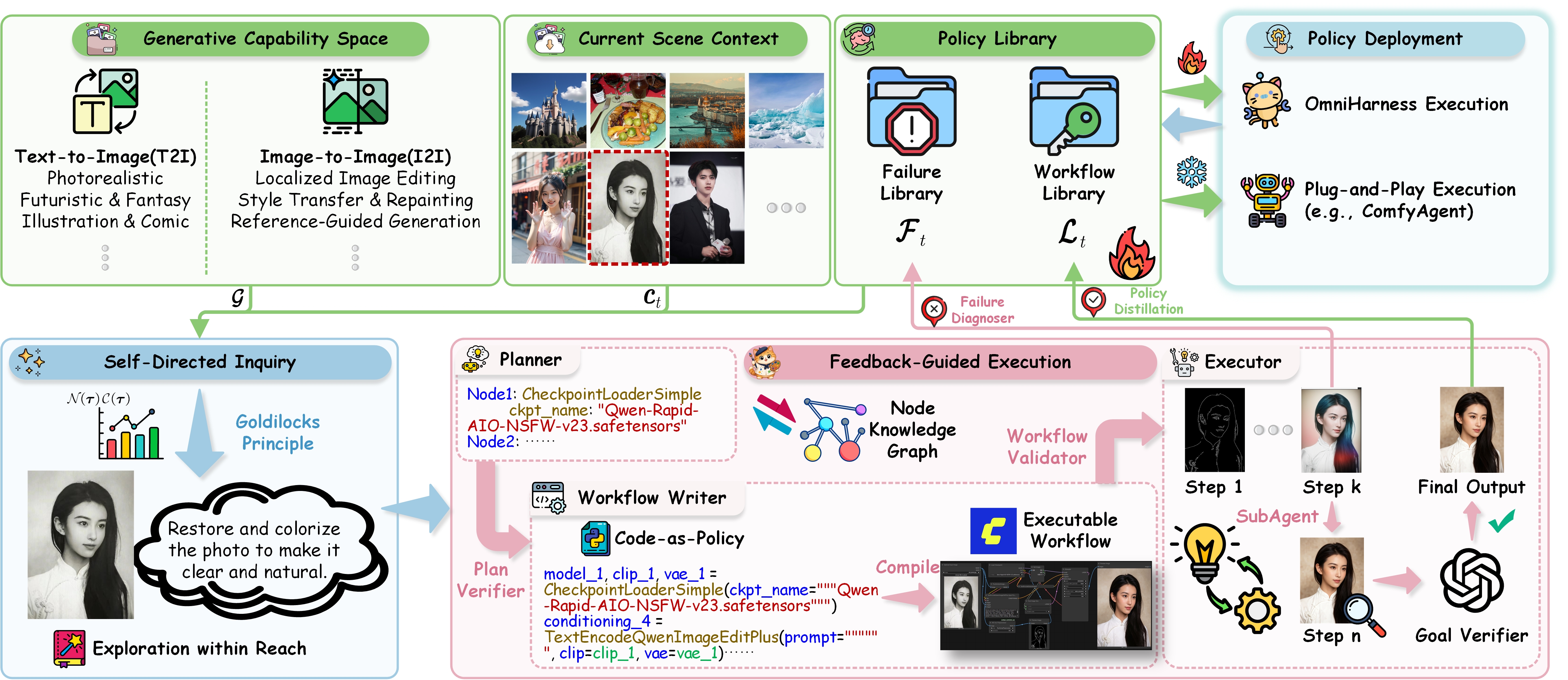}
    \caption{Architecture of OmniHarness. Self-directed inquiry enables autonomous exploration, while symbolic policy learning abstracts verified executions into symbolic policies for task families. Execution feedback continually updates the policies to guide subsequent tasks.}
    \label{fig:3}
\end{figure}

\section{OmniHarness: Symbolic Policy Learning}

As illustrated in Figure~\ref{fig:3}, \paperbold{OmniHarness} is a Code-as-Policy harness for visual generation. Before downstream objectives are specified, \paperbold{\textit{self-directed inquiry}} autonomously selects practice tasks near the current competence frontier. \paperbold{\textit{Symbolic retrieval}} guides planning, execution, and verification for each selected task. After execution, \paperbold{\textit{symbolic abstraction}} converts verified workflows into policies for task families, while execution feedback updates the policy library. The updated policies guide retrieval and execution in the next round.

\subsection{Self-Directed Inquiry}

At iteration $t$, OmniHarness selects a task from the candidate set $\mathcal{T}_t$ by balancing capability novelty $\mathcal{N}(\tau)$ and the competence frontier score $\mathcal{C}(\tau)$:
\begin{equation}
\resizebox{0.92\linewidth}{!}{$\displaystyle
\mathcal{N}(\tau)
=\frac{1}{|\mathcal{G}_\tau|}
\sum_{g\in\mathcal{G}_\tau}
\frac{1}{\sqrt{n_t(z_\tau,g)+1}},
\quad
\mathcal{C}(\tau)
=4\bar{r}_t(\tau)\left(1-\bar{r}_t(\tau)\right),
\quad
\tau_t
=\operatorname*{arg\,max}_{\tau\in\mathcal{T}_t}
\mathcal{N}(\tau)\mathcal{C}(\tau).
$}
\end{equation}
Here, $\mathcal{G}_\tau$ denotes the capabilities required by task $\tau$, and $n_t(z_\tau,g)$ counts previous practice attempts involving capability $g$ in context $z_\tau$, including unsuccessful attempts. The context is a shared token for text-to-image tasks and the source image for image-to-image tasks. Thus, novelty favors underexplored combinations of context and capability. To estimate task competence, OmniHarness takes the highest reliability among applicable, non-suspended workflows supporting each required capability, assigning $0.05$ when none are available. Workflow reliability is the lower endpoint of a 95\% Wilson confidence interval computed from usage and verified success counts. The minimum estimate across the required capabilities defines $\bar{r}_t(\tau)$. Following the Goldilocks principle~\citep{BARANES201349,10.1371/journal.pone.0036399}, $\mathcal{C}(\tau)$ peaks at $\bar{r}_t(\tau)=0.5$ and downweights tasks with very low or high estimated competence. The product of the two scores balances capability novelty with estimated task learnability.

For the selected task, symbolic retrieval and the ComfyUI knowledge graph guide planning. The plan is translated into a Code-as-Policy program and compiled into an executable workflow. Intermediate and final verification guide local failure recovery. Practice attempts and execution outcomes drive continual updates to the symbolic policy library, enabling the system to evolve.

\subsection{Symbolic Abstraction}

OmniHarness represents each verified workflow as a symbolic policy $w$ in the workflow library $\mathcal{L}_t$:
\begin{equation}
w=\bigl(m_w,\mathcal{G}_w,T_w,\nu_w,d_w,P_w,E_w,D_w,u_w,s_w,\sigma_w\bigr).
\end{equation}
Here, $m_w$, $\mathcal{G}_w$, $T_w$, $\nu_w$, and $d_w$ denote \emph{Modality}, \emph{Capability Categories}, \emph{Workflow Template}, \emph{Workflow Name}, and \emph{Workflow Description}.
$P_w$, $E_w$, and $D_w$ denote \emph{Preconditions}, \emph{Expected Effects}, and \emph{Dependencies}.
$u_w$, $s_w$, and $\sigma_w$ record \emph{Usage Count}, \emph{Success Count}, and \emph{Reliability Tier}.

$T_w$ retains reusable workflow structure, shared parameters, and execution settings, with placeholders for task-specific inputs. Together with $d_w$ and $\mathcal{G}_w$, it provides an \emph{instance-agnostic abstraction} for a task family. $\nu_w$ identifies the policy, while $m_w$ and $\mathcal{G}_w$ support retrieval by task type. $P_w$, $D_w$, and $E_w$ specify input and task requirements, execution dependencies, and intended effects, respectively. \emph{Adaptive instantiation} binds inputs and adjusts template parameters subject to $P_w$ and $D_w$. \emph{Workflow composition} uses $\mathcal{G}_w$ to select components, $E_w$ and $P_w$ to check their compatibility, and $D_w$ to verify resource availability. Each invocation increments $u_w$, while verified success increments $s_w$. The tier $\sigma_w$ is initialized as \emph{Provisional} and updated using the empirical success rate $p_w=s_w/u_w$. It becomes \emph{Validated} when $u_w\geq3$ and $p_w\geq0.5$, or \emph{Suspended} when $u_w\geq10$ and $p_w\leq0.2$. Retrieval prioritizes validated policies and deprioritizes suspended ones.

\subsection{Symbolic Retrieval}

\begin{figure}[!t]
    \centering
    \includegraphics[width=\textwidth]{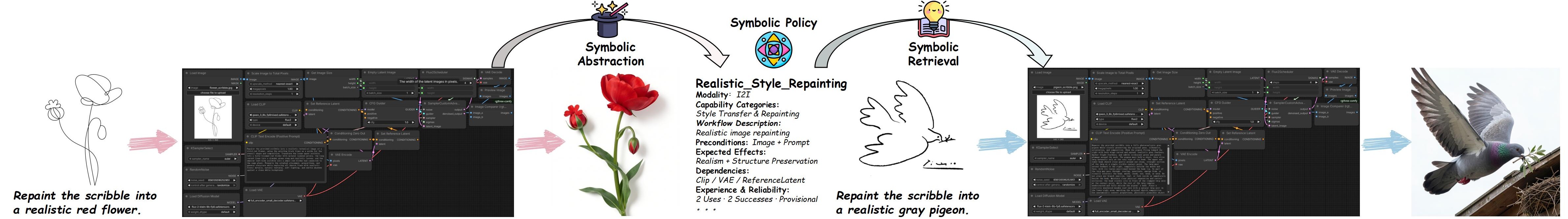}
    \caption{An example of cross-task reuse through symbolic abstraction and retrieval.}
    \label{fig:cross_task_reuse}
\end{figure}

OmniHarness filters candidate policies by modality $m_w$ and capability categories $\mathcal{G}_w$. Retrieval operates on metadata, reducing contextual redundancy and the need to infer intended effects from full parameterized workflows. It shortlists the top five candidates using workflow names $\nu_w$, then selects the best match using workflow descriptions $d_w$ and expected effects $E_w$. Reliability tiers $\sigma_w$ guide this process by prioritizing \emph{Validated} policies and deprioritizing \emph{Suspended} ones. Only after selection does the harness load the full template $T_w$ to construct a workflow for the current task, subject to preconditions $P_w$ and dependencies $D_w$ and guided by corrective knowledge from the failure library $\mathcal{F}_t$. Execution feedback updates the symbolic policy library after each task, guiding subsequent retrieval and reuse. A cross-task reuse example is shown in Figure~\ref{fig:cross_task_reuse}.

\section{Experiments}

\subsection{Experimental Setup and Evaluation Protocol}

All OmniHarness experiments are conducted in ComfyUI. Unless otherwise specified, OmniHarness uses Codex GPT-4o with \texttt{temperature}=0 and \texttt{top\_p}=1. Model parameters remain fixed throughout. For fair comparison, OmniHarness and the compared agents use a maximum of four retries \citep{symbomni} and the same underlying generative models within each evaluation setting \citep{comfybench,comfymind,symbomni}. Unified generative models and published baselines follow their respective benchmark protocols. Before each benchmark evaluation, the workflow and failure libraries are reset to the same snapshot obtained after 50 rounds of self-directed inquiry. Inquiry does not access benchmark evaluation tasks. Within each benchmark, execution feedback continually updates policies and associated records, allowing subsequent tasks to reuse the accumulated knowledge. These updates do not carry over to other benchmarks.

\subsection{Comparative Evaluation on Image Generation Benchmarks}

\begin{figure}[!htb]
\centering
\includegraphics[width=\linewidth]{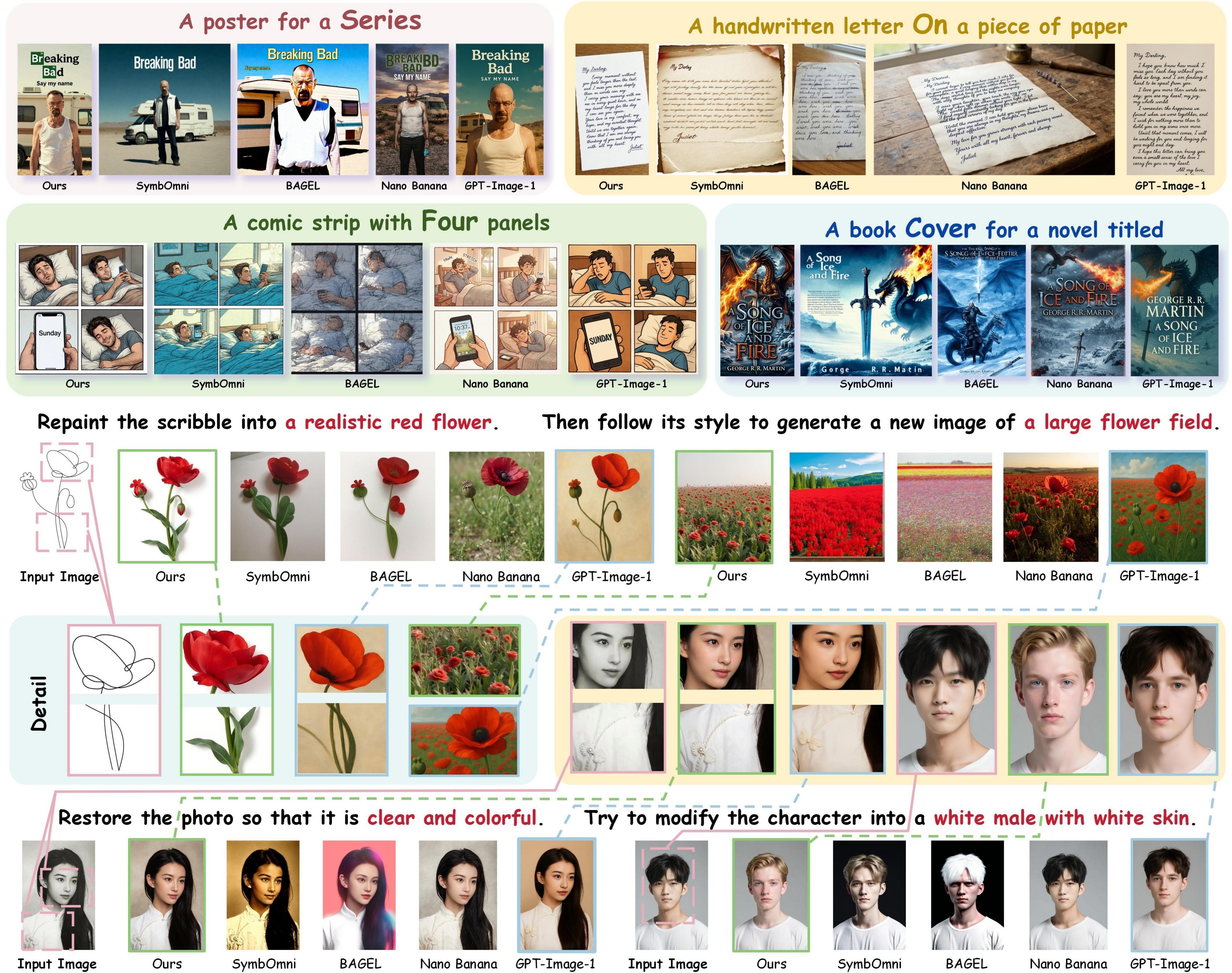}
\caption{Qualitative comparison on challenging tasks from the Complex and
Creative subsets.}
\label{fig:11}
\end{figure}

\paperbold{Autonomous Workflow Construction.}
We evaluate autonomous workflow construction on ComfyBench~\citep{comfybench}, where agents construct executable ComfyUI workflows that satisfy task requirements. As shown in Table~\ref{tab:comfybench_workflow}, OmniHarness achieves a 100.0\% Pass rate across all subsets and a 92.5\% Total Resolve rate, exceeding SymbOmni by 6.5 percentage points. The largest gain occurs on Creative tasks, where its 95.0\% Resolve rate exceeds SymbOmni by 27.5 percentage points. Figure~\ref{fig:11} provides qualitative comparisons.

\begin{table}[!htb]
\caption{Quantitative comparison on ComfyBench~\citep{comfybench}.}
\label{tab:comfybench_workflow}
\centering

\begingroup
\fontsize{8pt}{10pt}\selectfont
\setlength{\tabcolsep}{1.0pt}
\renewcommand{\arraystretch}{0.90}
\setlength{\extrarowheight}{0.2pt}
\setlength{\aboverulesep}{0.6pt}
\setlength{\belowrulesep}{0.6pt}
\setlength{\cmidrulesep}{0.5pt}
\renewcommand{\tabularxcolumn}[1]{m{#1}}

\definecolor{cbLight}{HTML}{F6F8FA}
\definecolor{cbDark}{HTML}{C9DDEC}
\definecolor{cbGainColor}{HTML}{940001}

\newcommand{\cbHead}[1]{%
    {\fontsize{6pt}{7.4pt}\selectfont\bfseries\mbox{#1}}%
}
\newcommand{\cbCell}[2]{%
    \mbox{#1\kern0.25pt%
    {\normalfont\fontsize{6pt}{7.4pt}\selectfont #2}}%
}
\newcommand{\cbGain}[2]{%
    \cbCell{#1}{%
        \textcolor{cbGainColor}{\ensuremath{\uparrow}#2}}%
}
\newcommand{\cbEqual}[1]{%
    \cbCell{#1}{\textcolor{gray}{\ensuremath{=}0.0}}%
}

\begin{tabularx}{\linewidth}{
    l
    *{8}{>{\centering\arraybackslash}X}
}
\toprule
\multicolumn{1}{c}{\multirow[c]{2}{*}{\paperbold{Agent}}}
& \multicolumn{2}{c}{\paperbold{Vanilla}}
& \multicolumn{2}{c}{\paperbold{Complex}}
& \multicolumn{2}{c}{\paperbold{Creative}}
& \multicolumn{2}{c}{\paperbold{Total}} \\
\cmidrule(lr){2-3}
\cmidrule(lr){4-5}
\cmidrule(lr){6-7}
\cmidrule(lr){8-9}
& \cbHead{Pass\(\uparrow\)}
& \cbHead{Res.\(\uparrow\)}
& \cbHead{Pass\(\uparrow\)}
& \cbHead{Res.\(\uparrow\)}
& \cbHead{Pass\(\uparrow\)}
& \cbHead{Res.\(\uparrow\)}
& \cbHead{Pass\(\uparrow\)}
& \cbHead{Res.\(\uparrow\)} \\
\midrule

GPT-4o + Zero-shot
& 0.0 & 0.0 & 0.0 & 0.0
& 0.0 & 0.0 & 0.0 & 0.0 \\

GPT-4o + Few-shot~\citep{Few-Shot}
& \cbGain{32.0}{32.0}
& \cbGain{27.0}{27.0}
& \cbGain{16.7}{16.7}
& \cbGain{8.3}{8.3}
& \cbGain{7.5}{7.5}
& \cbEqual{0.0}
& \cbGain{22.5}{22.5}
& \cbGain{16.0}{16.0} \\

GPT-4o + CoT~\citep{CoT}
& \cbGain{44.0}{44.0}
& \cbGain{29.0}{29.0}
& \cbGain{11.7}{11.7}
& \cbGain{8.3}{8.3}
& \cbGain{12.5}{12.5}
& \cbEqual{0.0}
& \cbGain{28.0}{28.0}
& \cbGain{17.0}{17.0} \\

GPT-4o + CoT-SC~\citep{cot-sc}
& \cbGain{45.0}{45.0}
& \cbGain{34.0}{34.0}
& \cbGain{11.7}{11.7}
& \cbGain{5.0}{5.0}
& \cbGain{15.0}{15.0}
& \cbEqual{0.0}
& \cbGain{29.0}{29.0}
& \cbGain{18.5}{18.5} \\

Claude-3.5-Sonnet + RAG~\citep{RAG}
& \cbGain{27.0}{27.0}
& \cbGain{13.0}{13.0}
& \cbGain{23.0}{23.0}
& \cbGain{6.7}{6.7}
& \cbGain{7.5}{7.5}
& \cbEqual{0.0}
& \cbGain{22.0}{22.0}
& \cbGain{8.5}{8.5} \\

Llama-3.1-70B + RAG
& \cbGain{58.0}{58.0}
& \cbGain{32.0}{32.0}
& \cbGain{23.0}{23.0}
& \cbGain{10.0}{10.0}
& \cbGain{15.0}{15.0}
& \cbGain{5.0}{5.0}
& \cbGain{39.0}{39.0}
& \cbGain{20.0}{20.0} \\

GPT-4o + RAG
& \cbGain{62.0}{62.0}
& \cbGain{41.0}{41.0}
& \cbGain{45.0}{45.0}
& \cbGain{21.7}{21.7}
& \cbGain{\underline{40.0}}{40.0}
& \cbGain{7.5}{7.5}
& \cbGain{52.0}{52.0}
& \cbGain{23.0}{23.0} \\

o1-mini + RAG
& \cbGain{32.0}{32.0}
& \cbGain{16.0}{16.0}
& \cbGain{21.7}{21.7}
& \cbGain{8.3}{8.3}
& \cbGain{12.5}{12.5}
& \cbGain{7.5}{7.5}
& \cbGain{25.0}{25.0}
& \cbGain{12.0}{12.0} \\

o1-preview + RAG
& \cbGain{\underline{70.0}}{70.0}
& \cbGain{46.0}{46.0}
& \cbGain{\underline{48.3}}{48.3}
& \cbGain{23.3}{23.3}
& \cbGain{30.0}{30.0}
& \cbGain{12.5}{12.5}
& \cbGain{55.5}{55.5}
& \cbGain{32.5}{32.5} \\

\midrule

\rowcolor{cbDark!0!cbLight}
Llama-3.1-70B + ComfyAgent~\citep{comfybench}
& 63.0 & 35.0 & 26.7 & 18.3
& 20.0 & 5.0 & 43.5 & 24.0 \\

\rowcolor{cbDark!17!cbLight}
GPT-4o + ComfyAgent
& \cbGain{67.0}{4.0}
& \cbGain{46.0}{11.0}
& \cbGain{\underline{48.3}}{21.6}
& \cbGain{21.7}{3.4}
& \cbGain{\underline{40.0}}{20.0}
& \cbGain{15.0}{10.0}
& \cbGain{\underline{56.0}}{12.5}
& \cbGain{32.5}{8.5} \\

\rowcolor{cbDark!33!cbLight}
GPT-4o + ComfyMind~\citep{comfymind}
& \cbGain{\paperbold{100.0}}{37.0}
& \cbGain{92.0}{57.0}
& \cbGain{\paperbold{100.0}}{73.3}
& \cbGain{\paperbold{85.0}}{66.7}
& \cbGain{\paperbold{100.0}}{80.0}
& \cbGain{57.5}{52.5}
& \cbGain{\paperbold{100.0}}{56.5}
& \cbGain{83.0}{59.0} \\

\rowcolor{cbDark!50!cbLight}
DeepSeek-V3 + ComfyMind
& \cbGain{\paperbold{100.0}}{37.0}
& \cbGain{90.0}{55.0}
& \cbGain{\paperbold{100.0}}{73.3}
& \cbGain{71.7}{53.4}
& \cbGain{\paperbold{100.0}}{80.0}
& \cbGain{60.0}{55.0}
& \cbGain{\paperbold{100.0}}{56.5}
& \cbGain{78.5}{54.5} \\

\rowcolor{cbDark!67!cbLight}
Gemini-2.5-Flash + SymbOmni~\citep{symbomni}
& \cbGain{\paperbold{100.0}}{37.0}
& \cbGain{\underline{95.0}}{60.0}
& \cbGain{\paperbold{100.0}}{73.3}
& \cbGain{\underline{83.3}}{65.0}
& \cbGain{\paperbold{100.0}}{80.0}
& \cbGain{\underline{67.5}}{62.5}
& \cbGain{\paperbold{100.0}}{56.5}
& \cbGain{86.0}{62.0} \\

\rowcolor{cbDark!83!cbLight}
\paperbold{GPT-4o + OmniHarness}
& \cbGain{\paperbold{100.0}}{37.0}
& \cbGain{\underline{95.0}}{60.0}
& \cbGain{\paperbold{100.0}}{73.3}
& \cbGain{76.7}{58.4}
& \cbGain{\paperbold{100.0}}{80.0}
& \cbGain{\paperbold{95.0}}{90.0}
& \cbGain{\paperbold{100.0}}{56.5}
& \cbGain{\underline{89.5}}{65.5} \\

\rowcolor{cbDark!100!cbLight}
\paperbold{Codex GPT-4o + OmniHarness}
& \cbGain{\paperbold{100.0}}{37.0}
& \cbGain{\paperbold{97.0}}{62.0}
& \cbGain{\paperbold{100.0}}{73.3}
& \cbGain{\underline{83.3}}{65.0}
& \cbGain{\paperbold{100.0}}{80.0}
& \cbGain{\paperbold{95.0}}{90.0}
& \cbGain{\paperbold{100.0}}{56.5}
& \cbGain{\paperbold{92.5}}{68.5} \\

\bottomrule
\end{tabularx}

\endgroup
\end{table}

\begin{figure}[!htb]
\centering
\includegraphics[width=\textwidth]{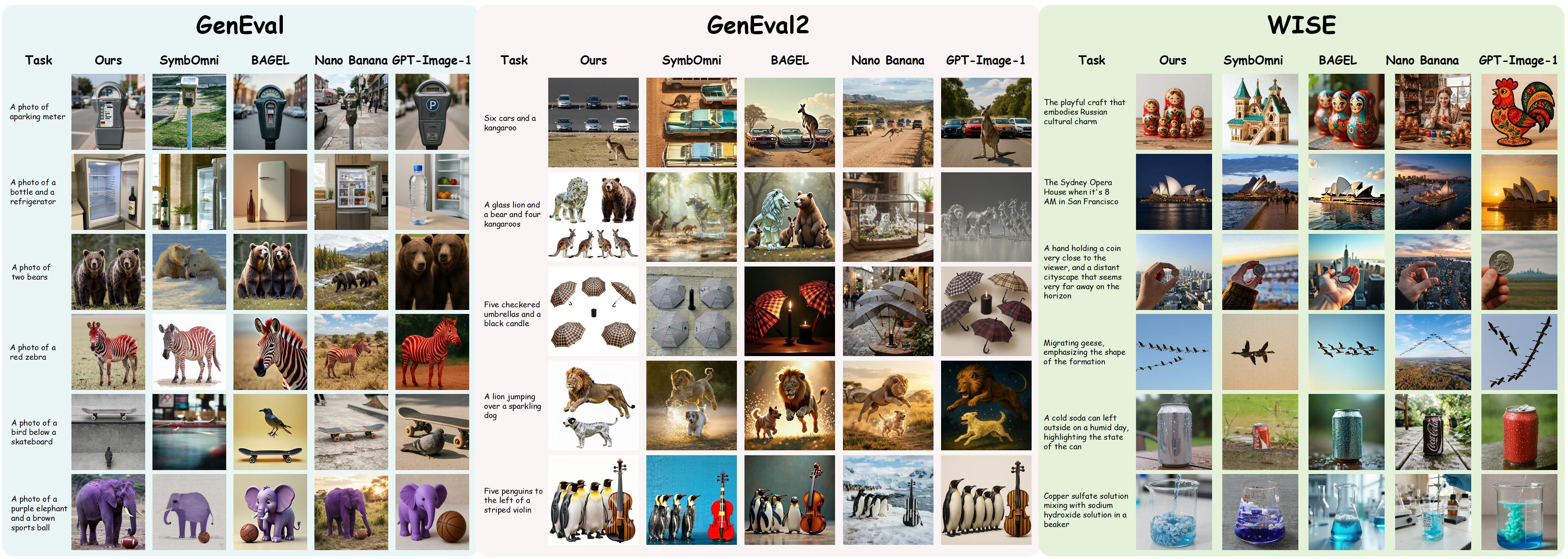}
\caption{Qualitative comparison on representative text-to-image tasks.}
\label{fig:text_to_image_qualitative}
\end{figure}

\newpage
\begingroup
\setlength{\intextsep}{4.5pt plus 1pt minus 1pt}
\paperbold{Text-to-Image Generation.}
We evaluate text-to-image generation on GenEval~\citep{geneval}, GenEval2~\citep{geneval2}, and WISE~\citep{WISE}. GenEval measures six object-centric compositional skills. GenEval2 tests fine-grained attributes, counting, and spatial and transitive verb relations, while WISE assesses knowledge-informed synthesis across cultural, spatiotemporal, and scientific domains. As shown in Tables~\ref{tab:geneval}, \ref{tab:geneval2}, and~\ref{tab:wise}, OmniHarness achieves the highest overall scores on GenEval (0.997) and WISE (0.86). On GenEval2, it leads in Attribute, Count, Position, and Verb, with the largest gain of 19.2 points in Count over the best competing score. Figure~\ref{fig:text_to_image_qualitative} presents qualitative results.

\begin{table}[!htb]
\caption{Quantitative comparison on GenEval~\citep{geneval}.}
\label{tab:geneval}
\centering

\begingroup
\fontsize{8pt}{10pt}\selectfont
\setlength{\tabcolsep}{1.0pt}
\renewcommand{\arraystretch}{0.90}
\setlength{\extrarowheight}{0.2pt}
\setlength{\aboverulesep}{0.6pt}
\setlength{\belowrulesep}{0.6pt}
\setlength{\cmidrulesep}{0.5pt}
\renewcommand{\tabularxcolumn}[1]{m{#1}}

\definecolor{gvLight}{HTML}{F6F8FA}
\definecolor{gvDark}{HTML}{C9DDEC}
\definecolor{gvGainColor}{HTML}{940001}
\definecolor{gvLossColor}{HTML}{006400}

\newcommand{\gvHead}[1]{\mbox{\paperbold{#1}}}

\newcommand{\gvCell}[2]{%
    \mbox{#1\kern0.25pt%
    {\normalfont\fontsize{6pt}{7.4pt}\selectfont #2}}%
}
\newcommand{\gvGain}[2]{%
    \gvCell{#1}{%
        \textcolor{gvGainColor}{\ensuremath{\uparrow}#2}}%
}
\newcommand{\gvLoss}[2]{%
    \gvCell{#1}{%
        \textcolor{gvLossColor}{\ensuremath{\downarrow}#2}}%
}

\begin{tabularx}{\linewidth}{
    l
    *{7}{>{\centering\arraybackslash}X}
}
\toprule
\multicolumn{1}{c}{\paperbold{Method}}
& \gvHead{Single Obj.\(\uparrow\)}
& \gvHead{Two Obj.\(\uparrow\)}
& \gvHead{Counting\(\uparrow\)}
& \gvHead{Colors\(\uparrow\)}
& \gvHead{Position\(\uparrow\)}
& \gvHead{Attr. Bind.\(\uparrow\)}
& \gvHead{Overall\(\uparrow\)} \\
\midrule

\multicolumn{8}{l}{
    \textit{Frozen Text-Encoder Mapping Methods}
} \\

DALL-E 2~\citep{DALLE}
& 0.94
& 0.66
& 0.49
& 0.77
& 0.10
& 0.19
& 0.52 \\

SD-XL~\citep{SDXL}
& \gvGain{0.98}{0.04}
& \gvGain{0.74}{0.08}
& \gvLoss{0.39}{0.10}
& \gvGain{0.85}{0.08}
& \gvGain{0.15}{0.05}
& \gvGain{0.23}{0.04}
& \gvGain{0.55}{0.03} \\

SD3-Medium~\citep{SD3}
& \gvGain{\underline{0.99}}{0.05}
& \gvGain{\underline{0.94}}{0.28}
& \gvGain{0.72}{0.23}
& \gvGain{0.89}{0.12}
& \gvGain{0.33}{0.23}
& \gvGain{0.60}{0.41}
& \gvGain{0.74}{0.22} \\

\midrule

\multicolumn{8}{l}{
    \textit{Unified Multimodal Models}
} \\

LlamaGen~\citep{llamagen}
& 0.71
& 0.34
& 0.21
& 0.58
& 0.07
& 0.04
& 0.32 \\

LWM~\citep{LWM}
& \gvGain{0.93}{0.22}
& \gvGain{0.41}{0.07}
& \gvGain{0.46}{0.25}
& \gvGain{0.79}{0.21}
& \gvGain{0.09}{0.02}
& \gvGain{0.15}{0.11}
& \gvGain{0.47}{0.15} \\

SEED-X~\citep{seedx}
& \gvGain{0.97}{0.26}
& \gvGain{0.58}{0.24}
& \gvGain{0.26}{0.05}
& \gvGain{0.80}{0.22}
& \gvGain{0.19}{0.12}
& \gvGain{0.14}{0.10}
& \gvGain{0.49}{0.17} \\

Emu3-Gen~\citep{emu3}
& \gvGain{0.98}{0.27}
& \gvGain{0.71}{0.37}
& \gvGain{0.34}{0.13}
& \gvGain{0.81}{0.23}
& \gvGain{0.17}{0.10}
& \gvGain{0.21}{0.17}
& \gvGain{0.54}{0.22} \\

Janus~\citep{janus}
& \gvGain{0.97}{0.26}
& \gvGain{0.68}{0.34}
& \gvGain{0.30}{0.09}
& \gvGain{0.84}{0.26}
& \gvGain{0.46}{0.39}
& \gvGain{0.42}{0.38}
& \gvGain{0.61}{0.29} \\

JanusFlow~\citep{JanusFlow}
& \gvGain{0.97}{0.26}
& \gvGain{0.59}{0.25}
& \gvGain{0.45}{0.24}
& \gvGain{0.83}{0.25}
& \gvGain{0.53}{0.46}
& \gvGain{0.42}{0.38}
& \gvGain{0.63}{0.31} \\

Janus-Pro-7B~\citep{Janus-Pro}
& \gvGain{\underline{0.99}}{0.28}
& \gvGain{0.89}{0.55}
& \gvGain{0.59}{0.38}
& \gvGain{0.90}{0.32}
& \gvGain{0.79}{0.72}
& \gvGain{0.66}{0.62}
& \gvGain{0.80}{0.48} \\

GoT~\citep{got}
& \gvGain{\underline{0.99}}{0.28}
& \gvGain{0.69}{0.35}
& \gvGain{0.67}{0.46}
& \gvGain{0.85}{0.27}
& \gvGain{0.34}{0.27}
& \gvGain{0.27}{0.23}
& \gvGain{0.64}{0.32} \\

Bagel~\citep{bagel}
& \gvGain{0.98}{0.27}
& \gvGain{\underline{0.94}}{0.60}
& \gvGain{0.76}{0.55}
& \gvGain{0.91}{0.33}
& \gvGain{0.69}{0.62}
& \gvGain{0.70}{0.66}
& \gvGain{0.78}{0.46} \\

GPT-Image-1
& \gvGain{\underline{0.99}}{0.28}
& \gvGain{0.92}{0.58}
& \gvGain{0.85}{0.64}
& \gvGain{0.92}{0.34}
& \gvGain{0.75}{0.68}
& \gvGain{0.61}{0.57}
& \gvGain{0.84}{0.52} \\

\midrule

\multicolumn{8}{l}{
    \textit{Collaborative AI Systems}
} \\

\rowcolor{gvDark!0!gvLight}
ComfyAgent~\citep{comfybench}
& 0.69
& 0.30
& 0.33
& 0.50
& 0.04
& 0.04
& 0.32 \\

\rowcolor{gvDark!33!gvLight}
ComfyMind~\citep{comfymind}
& \gvGain{\paperbold{1.00}}{0.31}
& \gvGain{\paperbold{1.00}}{0.70}
& \gvGain{0.96}{0.63}
& \gvGain{0.97}{0.47}
& \gvGain{0.63}{0.59}
& \gvGain{0.81}{0.77}
& \gvGain{0.90}{0.58} \\

\rowcolor{gvDark!67!gvLight}
SymbOmni~\citep{symbomni}
& \gvGain{\paperbold{1.00}}{0.31}
& \gvGain{\paperbold{1.00}}{0.70}
& \gvGain{\underline{0.99}}{0.66}
& \gvGain{\underline{0.98}}{0.48}
& \gvGain{\underline{0.97}}{0.93}
& \gvGain{\underline{0.95}}{0.91}
& \gvGain{\underline{0.98}}{0.66} \\

\rowcolor{gvDark!100!gvLight}
\paperbold{OmniHarness}
& \gvGain{\paperbold{1.00}}{0.31}
& \gvGain{\paperbold{1.00}}{0.70}
& \gvGain{\paperbold{1.00}}{0.67}
& \gvGain{\paperbold{1.00}}{0.50}
& \gvGain{\paperbold{1.00}}{0.96}
& \gvGain{\paperbold{0.98}}{0.94}
& \gvGain{\paperbold{0.997}}{0.677} \\

\bottomrule
\end{tabularx}

\endgroup
\end{table}

\begin{table}[!htb]
\caption{Quantitative comparison on GenEval2~\citep{geneval2}.}
\label{tab:geneval2}
\centering

\begingroup
\fontsize{8pt}{10pt}\selectfont
\setlength{\tabcolsep}{1.0pt}
\renewcommand{\arraystretch}{0.90}
\setlength{\extrarowheight}{0.2pt}
\setlength{\aboverulesep}{0.6pt}
\setlength{\belowrulesep}{0.6pt}
\setlength{\cmidrulesep}{0.5pt}
\renewcommand{\tabularxcolumn}[1]{m{#1}}

\definecolor{gevLight}{HTML}{E5EDF5}
\definecolor{gevDark}{HTML}{C9DDEC}
\definecolor{gevGainColor}{HTML}{940001}

\newcommand{\gevHead}[1]{\mbox{\paperbold{#1}}}

\newcommand{\gevCell}[2]{%
    \mbox{#1\kern0.25pt%
    {\normalfont\fontsize{6pt}{7.4pt}\selectfont #2}}%
}
\newcommand{\gevGain}[2]{%
    \gevCell{#1}{%
        \textcolor{gevGainColor}{\ensuremath{\uparrow}#2}}%
}
\newcommand{\gevEqual}[1]{%
    \gevCell{#1}{\textcolor{gray}{\ensuremath{=}0.0}}%
}

\begin{tabularx}{\linewidth}{
    l
    *{6}{>{\centering\arraybackslash}X}
}
\toprule
\multicolumn{1}{c}{\paperbold{Method}}
& \gevHead{Object\(\uparrow\)}
& \gevHead{Attribute\(\uparrow\)}
& \gevHead{Count\(\uparrow\)}
& \gevHead{Position\(\uparrow\)}
& \gevHead{Verb\(\uparrow\)}
& \gevHead{Overall\(\uparrow\)} \\
\midrule

\multicolumn{7}{l}{
    \textit{Stable Diffusion Model Series}
} \\

SDXL~\citep{SDXL}
& 74.1
& 42.4
& 28.7
& 16.0
& 28.9
& 38.02 \\

SD3~\citep{SD3}
& \gevGain{87.0}{12.9}
& \gevGain{65.5}{23.1}
& \gevGain{49.9}{21.2}
& \gevGain{41.0}{25.0}
& \gevGain{46.7}{17.8}
& \gevGain{58.02}{20.00} \\

SD3.5-Large~\citep{SD3}
& \gevGain{91.6}{17.5}
& \gevGain{70.3}{27.9}
& \gevGain{52.2}{23.5}
& \gevGain{39.5}{23.5}
& \gevGain{55.6}{26.7}
& \gevGain{61.84}{23.82} \\

\midrule

\multicolumn{7}{l}{
    \textit{State-of-the-Art Text-to-Image Models}
} \\

FLUX.1-dev~\citep{flux.1-dev}
& 88.4
& 68.3
& 55.6
& 37.0
& 44.4
& 58.74 \\

Bagel + CoT~\citep{bagel}
& \gevGain{92.9}{4.5}
& \gevGain{75.9}{7.6}
& \gevEqual{55.6}
& \gevGain{50.6}{13.6}
& \gevGain{57.8}{13.4}
& \gevGain{66.56}{7.82} \\

Qwen-Image~\citep{qwenimage}
& \gevGain{\paperbold{99.1}}{10.7}
& \gevGain{85.6}{17.3}
& \gevGain{70.3}{14.7}
& \gevGain{60.2}{23.2}
& \gevGain{71.1}{26.7}
& \gevGain{77.26}{18.52} \\

Gemini 2.5 Flash Image~\citep{gemini25}
& \gevGain{\underline{99.0}}{10.6}
& \gevGain{\underline{91.4}}{23.1}
& \gevGain{70.1}{14.5}
& \gevGain{\underline{70.2}}{33.2}
& \gevGain{\underline{86.7}}{42.3}
& \gevGain{\underline{83.48}}{24.74} \\

\midrule

\multicolumn{7}{l}{
    \textit{Collaborative AI Systems}
} \\

\rowcolor{gevLight}
SymbOmni~\citep{symbomni}
& 95.0
& 83.6
& \underline{74.8}
& 68.8
& 64.5
& 77.34 \\

\rowcolor{gevDark}
\paperbold{OmniHarness}
& \gevEqual{95.0}
& \gevGain{\paperbold{94.0}}{10.4}
& \gevGain{\paperbold{94.0}}{19.2}
& \gevGain{\paperbold{76.9}}{8.1}
& \gevGain{\paperbold{89.0}}{24.5}
& \gevGain{\paperbold{89.78}}{12.44} \\

\bottomrule
\end{tabularx}

\endgroup
\end{table}

\endgroup
\begin{table}[!htb]
\caption{Quantitative comparison on WISE~\citep{WISE}.}
\label{tab:wise}
\centering

\begingroup
\fontsize{8pt}{10pt}\selectfont
\setlength{\tabcolsep}{1.0pt}
\renewcommand{\arraystretch}{0.90}
\setlength{\extrarowheight}{0.2pt}
\setlength{\aboverulesep}{0.6pt}
\setlength{\belowrulesep}{0.6pt}
\setlength{\cmidrulesep}{0.5pt}
\renewcommand{\tabularxcolumn}[1]{m{#1}}

\definecolor{wsComfyMind}{HTML}{EFF3F7}
\definecolor{wsSymbOmni}{HTML}{E5EDF5}
\definecolor{wsOmniHarness}{HTML}{C9DDEC}
\definecolor{wsGainColor}{HTML}{940001}
\definecolor{wsLossColor}{HTML}{006400}

\newcommand{\wsHead}[1]{\mbox{\paperbold{#1}}}

\newcommand{\wsCell}[2]{%
    \mbox{#1\kern0.25pt%
    {\normalfont\fontsize{6pt}{7.4pt}\selectfont #2}}%
}
\newcommand{\wsGain}[2]{%
    \wsCell{#1}{%
        \textcolor{wsGainColor}{\ensuremath{\uparrow}#2}}%
}
\newcommand{\wsLoss}[2]{%
    \wsCell{#1}{%
        \textcolor{wsLossColor}{\ensuremath{\downarrow}#2}}%
}
\newcommand{\wsEqual}[1]{%
    \wsCell{#1}{\textcolor{gray}{\ensuremath{=}0.00}}%
}

\begin{tabularx}{\linewidth}{
    l
    *{7}{>{\centering\arraybackslash}X}
}
\toprule
\multicolumn{1}{c}{\paperbold{Method}}
& \wsHead{Cultural\(\uparrow\)}
& \wsHead{Time\(\uparrow\)}
& \wsHead{Space\(\uparrow\)}
& \wsHead{Biology\(\uparrow\)}
& \wsHead{Physics\(\uparrow\)}
& \wsHead{Chemistry\(\uparrow\)}
& \wsHead{Overall\(\uparrow\)} \\
\midrule

\multicolumn{8}{l}{\textit{Dedicated T2I Models}} \\

SD-XL~\citep{SDXL}
& 0.43 & 0.48 & 0.47 & 0.44
& 0.45 & 0.27 & 0.43 \\

SD3-Medium~\citep{SD3}
& \wsLoss{0.42}{0.01}
& \wsLoss{0.44}{0.04}
& \wsGain{0.48}{0.01}
& \wsLoss{0.39}{0.05}
& \wsGain{0.47}{0.02}
& \wsGain{0.29}{0.02}
& \wsLoss{0.42}{0.01} \\

SD3.5-Medium~\citep{SD3}
& \wsEqual{0.43}
& \wsGain{0.50}{0.02}
& \wsGain{0.52}{0.05}
& \wsLoss{0.41}{0.03}
& \wsGain{0.53}{0.08}
& \wsGain{0.33}{0.06}
& \wsGain{0.45}{0.02} \\

SD3.5-Large~\citep{SD3}
& \wsGain{0.44}{0.01}
& \wsGain{0.50}{0.02}
& \wsGain{0.58}{0.11}
& \wsEqual{0.44}
& \wsGain{0.52}{0.07}
& \wsGain{0.31}{0.04}
& \wsGain{0.46}{0.03} \\

PixArt-Alpha~\citep{PixArt}
& \wsGain{0.45}{0.02}
& \wsGain{0.50}{0.02}
& \wsGain{0.48}{0.01}
& \wsGain{0.49}{0.05}
& \wsGain{0.56}{0.11}
& \wsGain{0.34}{0.07}
& \wsGain{0.47}{0.04} \\

Playground-v2.5~\citep{playgroundv25}
& \wsGain{0.49}{0.06}
& \wsGain{0.58}{0.10}
& \wsGain{0.55}{0.08}
& \wsLoss{0.43}{0.01}
& \wsGain{0.48}{0.03}
& \wsGain{0.33}{0.06}
& \wsGain{0.49}{0.06} \\

FLUX.1-schnell~\citep{flux.1-dev}
& \wsLoss{0.39}{0.04}
& \wsLoss{0.44}{0.04}
& \wsGain{0.50}{0.03}
& \wsLoss{0.31}{0.13}
& \wsLoss{0.44}{0.01}
& \wsLoss{0.26}{0.01}
& \wsLoss{0.40}{0.03} \\

FLUX.1-dev~\citep{flux.1-dev}
& \wsGain{0.48}{0.05}
& \wsGain{0.58}{0.10}
& \wsGain{0.62}{0.15}
& \wsLoss{0.42}{0.02}
& \wsGain{0.51}{0.06}
& \wsGain{0.35}{0.08}
& \wsGain{0.50}{0.07} \\

\midrule

\multicolumn{8}{l}{\textit{Unified MLLM Models}} \\

Janus-1.3B~\citep{janus}
& 0.16 & 0.26 & 0.35 & 0.28
& 0.30 & 0.14 & 0.23 \\

JanusFlow-1.3B~\citep{JanusFlow}
& \wsLoss{0.13}{0.03}
& \wsEqual{0.26}
& \wsLoss{0.28}{0.07}
& \wsLoss{0.20}{0.08}
& \wsLoss{0.19}{0.11}
& \wsLoss{0.11}{0.03}
& \wsLoss{0.18}{0.05} \\

Janus-Pro-1B~\citep{Janus-Pro}
& \wsGain{0.20}{0.04}
& \wsGain{0.28}{0.02}
& \wsGain{0.45}{0.10}
& \wsLoss{0.24}{0.04}
& \wsGain{0.32}{0.02}
& \wsGain{0.16}{0.02}
& \wsGain{0.26}{0.03} \\

Janus-Pro-7B~\citep{Janus-Pro}
& \wsGain{0.30}{0.14}
& \wsGain{0.37}{0.11}
& \wsGain{0.49}{0.14}
& \wsGain{0.36}{0.08}
& \wsGain{0.42}{0.12}
& \wsGain{0.26}{0.12}
& \wsGain{0.35}{0.12} \\

Show-o~\citep{show-o}
& \wsGain{0.28}{0.12}
& \wsGain{0.36}{0.10}
& \wsGain{0.40}{0.05}
& \wsLoss{0.23}{0.05}
& \wsGain{0.33}{0.03}
& \wsGain{0.22}{0.08}
& \wsGain{0.30}{0.07} \\

Show-o-512~\citep{show-o}
& \wsGain{0.28}{0.12}
& \wsGain{0.40}{0.14}
& \wsGain{0.48}{0.13}
& \wsGain{0.30}{0.02}
& \wsGain{0.46}{0.16}
& \wsGain{0.30}{0.16}
& \wsGain{0.35}{0.12} \\

VILA-U-7B~\citep{VILA-U}
& \wsGain{0.26}{0.10}
& \wsGain{0.33}{0.07}
& \wsGain{0.37}{0.02}
& \wsGain{0.35}{0.07}
& \wsGain{0.39}{0.09}
& \wsGain{0.23}{0.09}
& \wsGain{0.31}{0.08} \\

Orthus-7B-base~\citep{orthus}
& \wsLoss{0.07}{0.09}
& \wsLoss{0.10}{0.16}
& \wsLoss{0.12}{0.23}
& \wsLoss{0.15}{0.13}
& \wsLoss{0.15}{0.15}
& \wsLoss{0.10}{0.04}
& \wsLoss{0.10}{0.13} \\

Orthus-7B-instruct~\citep{orthus}
& \wsGain{0.23}{0.07}
& \wsGain{0.31}{0.05}
& \wsGain{0.38}{0.03}
& \wsEqual{0.28}
& \wsGain{0.31}{0.01}
& \wsGain{0.20}{0.06}
& \wsGain{0.27}{0.04} \\

Emu3~\citep{emu3}
& \wsGain{0.34}{0.18}
& \wsGain{0.45}{0.19}
& \wsGain{0.48}{0.13}
& \wsGain{0.41}{0.13}
& \wsGain{0.45}{0.15}
& \wsGain{0.27}{0.13}
& \wsGain{0.39}{0.16} \\

BAGEL~\citep{bagel}
& \wsGain{0.44}{0.28}
& \wsGain{0.55}{0.29}
& \wsGain{0.68}{0.33}
& \wsGain{0.44}{0.16}
& \wsGain{0.60}{0.30}
& \wsGain{0.39}{0.25}
& \wsGain{0.52}{0.29} \\

BAGEL + CoT~\citep{bagel}
& \wsGain{0.76}{0.60}
& \wsGain{0.69}{0.43}
& \wsGain{0.75}{0.40}
& \wsGain{0.65}{0.37}
& \wsGain{0.75}{0.45}
& \wsGain{0.58}{0.44}
& \wsGain{0.70}{0.47} \\

\midrule

\multicolumn{8}{l}{\textit{Closed-Source Models}} \\

GPT-Image-1
& 0.81
& \underline{0.71}
& \paperbold{0.89}
& \underline{0.83}
& \underline{0.79}
& 0.74
& \underline{0.80} \\

\midrule

\multicolumn{8}{l}{\textit{Collaborative AI Systems}} \\

\rowcolor{wsComfyMind}
ComfyMind~\citep{comfymind}
& 0.85
& 0.66
& 0.72
& 0.67
& 0.70
& \underline{0.78}
& 0.76 \\

\rowcolor{wsSymbOmni}
SymbOmni~\citep{symbomni}
& \wsGain{\paperbold{0.90}}{0.05}
& \wsGain{0.70}{0.04}
& \wsGain{0.74}{0.02}
& \wsGain{0.75}{0.08}
& \wsGain{0.74}{0.04}
& \wsEqual{\underline{0.78}}
& \wsGain{\underline{0.80}}{0.04} \\

\rowcolor{wsOmniHarness}
\paperbold{OmniHarness}
& \wsGain{\underline{0.88}}{0.03}
& \wsGain{\paperbold{0.85}}{0.19}
& \wsGain{\underline{0.86}}{0.14}
& \wsGain{\paperbold{0.84}}{0.17}
& \wsGain{\paperbold{0.82}}{0.12}
& \wsGain{\paperbold{0.86}}{0.08}
& \wsGain{\paperbold{0.86}}{0.10} \\

\bottomrule
\end{tabularx}

\endgroup
\end{table}

\begin{figure}[!htb]
\centering
\includegraphics[width=\textwidth]{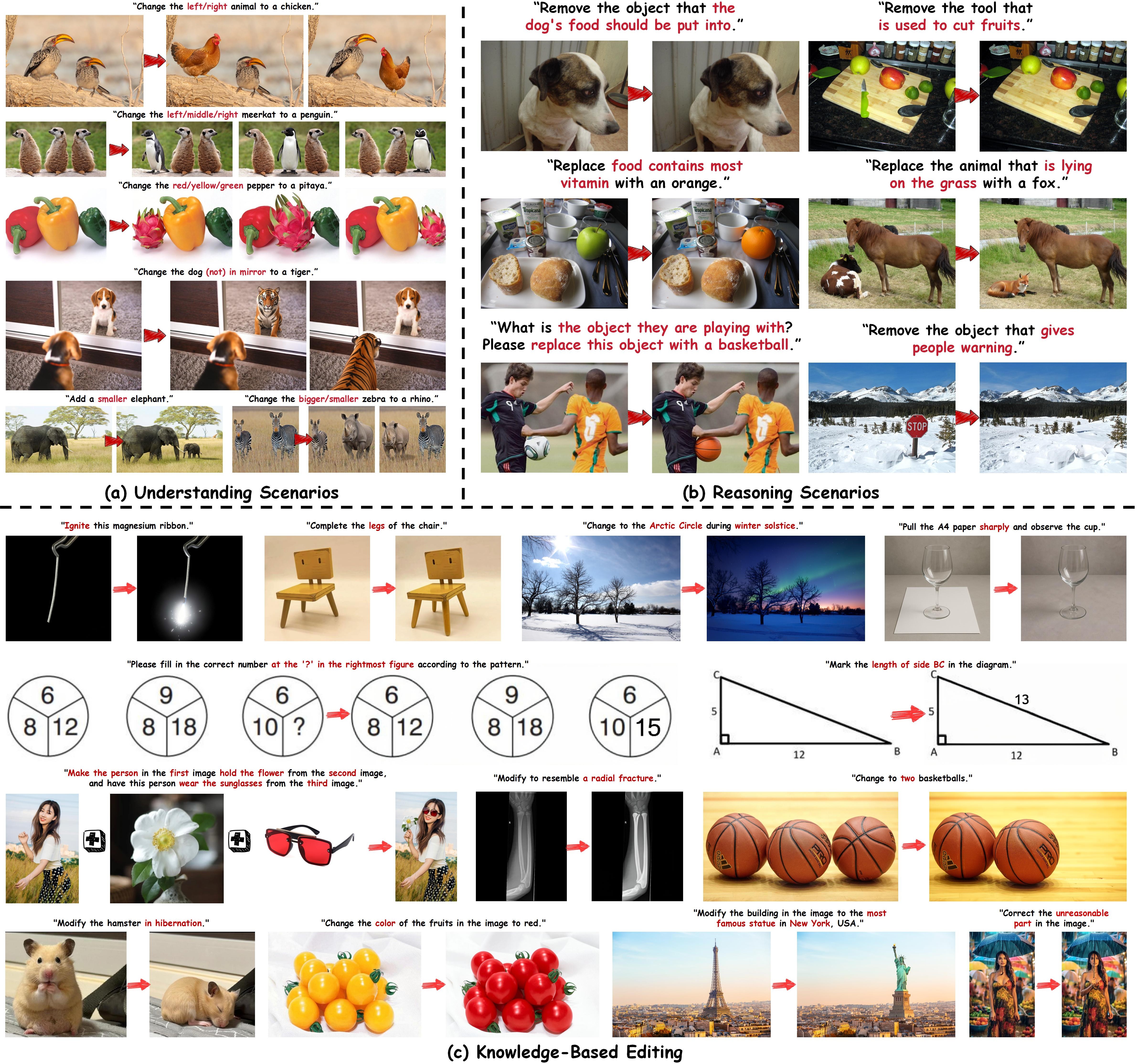}
\caption{Qualitative image editing results on (a) Understanding Scenarios and (b) Reasoning Scenarios in Reason-Edit, and (c) knowledge-based editing tasks in KRIS-Bench.}
\label{fig:reason_edit_qualitative}
\end{figure}

\paperbold{Image Editing.}
We evaluate image editing on Reason-Edit~\citep{smartedit} and KRIS-Bench~\citep{KRIS-Bench}. Reason-Edit provides explicit target cues in Understanding Scenarios and indirect target descriptions in Reasoning Scenarios. KRIS-Bench assesses factual, conceptual, and procedural knowledge in generation and editing. Table~\ref{tab:reason_edit} shows that OmniHarness leads on all four metrics in Understanding Scenarios and achieves the highest SSIM of 0.80 and CLIP Score of 21.32 in Reasoning Scenarios. On KRIS-Bench, it achieves the highest Conceptual score of 81.89 and the second-highest overall score of 77.33, as shown in Table~\ref{tab:krisbench}. Qualitative results are shown in Figure~\ref{fig:reason_edit_qualitative}.

\begin{table}[!htb]
\caption{Quantitative comparison on Reason-Edit~\citep{smartedit}.}
\label{tab:reason_edit}
\centering

\begingroup
\fontsize{8pt}{10pt}\selectfont
\setlength{\tabcolsep}{1.0pt}
\renewcommand{\arraystretch}{0.90}
\setlength{\extrarowheight}{0.2pt}
\setlength{\aboverulesep}{0.6pt}
\setlength{\belowrulesep}{0.6pt}
\setlength{\cmidrulesep}{0.5pt}
\renewcommand{\tabularxcolumn}[1]{m{#1}}

\definecolor{reOmniHarness}{HTML}{C9DDEC}
\definecolor{reGainColor}{HTML}{940001}
\definecolor{reLossColor}{HTML}{006400}

\newcommand{\reHead}[1]{%
    {\fontsize{6pt}{7.4pt}\selectfont\bfseries\mbox{#1}}%
}
\newcommand{\reCell}[2]{%
    \mbox{#1\kern0.25pt%
    {\normalfont\fontsize{6pt}{7.4pt}\selectfont #2}}%
}
\newcommand{\reGain}[2]{%
    \reCell{#1}{%
        \textcolor{reGainColor}{\ensuremath{\uparrow}#2}}%
}
\newcommand{\reLoss}[2]{%
    \reCell{#1}{%
        \textcolor{reLossColor}{\ensuremath{\downarrow}#2}}%
}
\newcommand{\reApprox}[1]{%
    \reCell{#1}{\textcolor{gray}{\ensuremath{\approx}0.00}}%
}

\begin{tabularx}{\linewidth}{
    l
    *{8}{>{\centering\arraybackslash}X}
}
\toprule
\multicolumn{1}{c}{\multirow[c]{2}{*}{\paperbold{Method}}}
& \multicolumn{4}{c}{\paperbold{Understanding Scenarios}}
& \multicolumn{4}{c}{\paperbold{Reasoning Scenarios}} \\
\cmidrule(lr){2-5}
\cmidrule(lr){6-9}
& \reHead{PSNR\(\uparrow\)}
& \reHead{SSIM\(\uparrow\)}
& \reHead{LPIPS\(\downarrow\)}
& \reHead{CLIP\(\uparrow\)}
& \reHead{PSNR\(\uparrow\)}
& \reHead{SSIM\(\uparrow\)}
& \reHead{LPIPS\(\downarrow\)}
& \reHead{CLIP\(\uparrow\)} \\
\midrule

InstructPix2Pix~\citep{InstructPix2Pix}
& 21.58
& 0.72
& 0.09
& 22.76
& 24.23
& 0.71
& 0.08
& 19.41 \\

MagicBrush~\citep{MagicBrush}
& \reLoss{18.12}{3.46}
& \reLoss{0.68}{0.04}
& \reGain{0.14}{0.05}
& \reLoss{22.62}{0.14}
& \reLoss{22.10}{2.13}
& \reLoss{0.69}{0.01}
& \reGain{0.11}{0.03}
& \reGain{19.76}{0.34} \\

InstructDiffusion~\citep{InstructDiffusion}
& \reGain{23.26}{1.68}
& \reGain{0.74}{0.02}
& \reLoss{\underline{0.07}}{0.02}
& \reGain{23.08}{0.32}
& \reLoss{21.45}{2.78}
& \reLoss{0.67}{0.04}
& \reGain{0.12}{0.03}
& \reGain{19.52}{0.11} \\

SmartEdit-7B~\citep{smartedit}
& \reGain{22.05}{0.47}
& \reGain{0.73}{0.01}
& \reApprox{0.09}
& \reGain{23.61}{0.85}
& \reGain{25.26}{1.02}
& \reGain{0.74}{0.04}
& \reLoss{\underline{0.06}}{0.03}
& \reGain{\underline{20.95}}{1.54} \\

SmartEdit-13B~\citep{smartedit}
& \reGain{\underline{23.60}}{2.02}
& \reGain{\underline{0.75}}{0.03}
& \reLoss{\underline{0.07}}{0.02}
& \reGain{23.54}{0.77}
& \reGain{\paperbold{25.76}}{1.52}
& \reGain{\underline{0.75}}{0.04}
& \reLoss{\paperbold{0.05}}{0.03}
& \reGain{20.78}{1.36} \\

InsightEdit~\citep{InsightEdit}
& \reGain{23.59}{2.01}
& \reGain{\underline{0.75}}{0.03}
& \reLoss{\underline{0.07}}{0.02}
& \reGain{\underline{23.73}}{0.97}
& \reGain{\underline{25.71}}{1.48}
& \reGain{\underline{0.75}}{0.04}
& \reLoss{\paperbold{0.05}}{0.03}
& \reGain{20.87}{1.45} \\

\midrule

\rowcolor{reOmniHarness}
\paperbold{OmniHarness}
& \reGain{\paperbold{23.89}}{2.32}
& \reGain{\paperbold{0.86}}{0.14}
& \reLoss{\paperbold{0.05}}{0.04}
& \reGain{\paperbold{24.55}}{1.79}
& \reLoss{23.87}{0.36}
& \reGain{\paperbold{0.80}}{0.09}
& \reLoss{\paperbold{0.05}}{0.03}
& \reGain{\paperbold{21.32}}{1.91} \\

\bottomrule
\end{tabularx}

\endgroup
\end{table}

\begin{table}[!htb]
\caption{Quantitative comparison on KRIS-Bench~\citep{KRIS-Bench}.}
\label{tab:krisbench}
\centering

\begingroup
\fontsize{8pt}{10pt}\selectfont
\setlength{\tabcolsep}{1.0pt}
\renewcommand{\arraystretch}{0.90}
\setlength{\extrarowheight}{0.2pt}
\setlength{\aboverulesep}{0.6pt}
\setlength{\belowrulesep}{0.6pt}
\setlength{\cmidrulesep}{0.5pt}
\renewcommand{\tabularxcolumn}[1]{m{#1}}

\definecolor{krSymbOmni}{HTML}{E5EDF5}
\definecolor{krOmniHarness}{HTML}{C9DDEC}
\definecolor{krGainColor}{HTML}{940001}
\definecolor{krLossColor}{HTML}{006400}

\newcommand{\krHead}[1]{\mbox{\paperbold{#1}}}

\newcommand{\krCell}[2]{%
    \mbox{#1\kern0.25pt%
    {\normalfont\fontsize{6pt}{7.4pt}\selectfont #2}}%
}
\newcommand{\krGain}[2]{%
    \krCell{#1}{%
        \textcolor{krGainColor}{\ensuremath{\uparrow}#2}}%
}
\newcommand{\krLoss}[2]{%
    \krCell{#1}{%
        \textcolor{krLossColor}{\ensuremath{\downarrow}#2}}%
}

\begin{tabularx}{\linewidth}{
    l
    *{4}{>{\centering\arraybackslash}X}
}
\toprule
\multicolumn{1}{c}{\paperbold{Method}}
& \krHead{Factual\(\uparrow\)}
& \krHead{Conceptual\(\uparrow\)}
& \krHead{Procedural\(\uparrow\)}
& \krHead{Overall\(\uparrow\)} \\
\midrule

\multicolumn{5}{l}{\textit{Open-Source Models}} \\

InsPix2Pix~\citep{InstructPix2Pix}
& 23.33
& 25.59
& 17.28
& 22.82 \\

MagicBrush~\citep{MagicBrush}
& \krGain{41.84}{18.51}
& \krGain{39.24}{13.65}
& \krGain{26.54}{9.26}
& \krGain{37.15}{14.33} \\

Emu2~\citep{emu2}
& \krGain{45.40}{22.07}
& \krGain{37.54}{11.95}
& \krGain{34.91}{17.63}
& \krGain{39.70}{16.88} \\

OmniGen~\citep{OmniGen}
& \krGain{33.11}{9.78}
& \krGain{28.02}{2.43}
& \krGain{23.89}{6.61}
& \krGain{28.85}{6.03} \\

AnyEdit~\citep{AnyEdit}
& \krGain{39.26}{15.93}
& \krGain{41.88}{16.29}
& \krGain{31.74}{14.46}
& \krGain{38.55}{15.73} \\

Step1X-Edit~\citep{Step1X-Edit}
& \krGain{45.52}{22.19}
& \krGain{48.01}{22.42}
& \krGain{31.82}{14.54}
& \krGain{43.29}{20.47} \\

BAGEL~\citep{bagel}
& \krGain{47.71}{24.38}
& \krGain{52.17}{26.58}
& \krGain{40.23}{22.95}
& \krGain{47.76}{24.94} \\

BAGEL-Think~\citep{bagel}
& \krGain{55.77}{32.44}
& \krGain{59.44}{33.85}
& \krGain{39.26}{21.98}
& \krGain{53.36}{30.54} \\

\midrule

\multicolumn{5}{l}{\textit{Closed-Source Models}} \\

Gemini 2.0 Flash Experimental
& 65.26
& 59.65
& 62.90
& 62.41 \\

Doubao
& \krLoss{63.30}{1.96}
& \krGain{62.23}{2.58}
& \krLoss{54.17}{8.73}
& \krLoss{60.70}{1.71} \\

GPT-Image-1
& \krGain{\paperbold{79.80}}{14.54}
& \krGain{\underline{81.37}}{21.72}
& \krGain{\paperbold{78.32}}{15.42}
& \krGain{\paperbold{80.09}}{17.68} \\

\midrule

\multicolumn{5}{l}{\textit{Collaborative AI Systems}} \\

\rowcolor{krSymbOmni}
SymbOmni~\citep{symbomni}
& 73.33
& 72.28
& 70.29
& 72.18 \\

\rowcolor{krOmniHarness}
\paperbold{OmniHarness}
& \krGain{\underline{74.81}}{1.48}
& \krGain{\paperbold{81.89}}{9.61}
& \krGain{\underline{73.22}}{2.93}
& \krGain{\underline{77.33}}{5.15} \\

\bottomrule
\end{tabularx}

\endgroup
\end{table}

\subsection{Ablation Study}

\paperbold{Symbolic Policy Learning Ablations.}
The component ablations in Table~\ref{tab:ablation_comfybench} demonstrate that abstraction, adaptation, and composition improve generalization across tasks. Table~\ref{tab:symbolic_retrieval_ablation} shows higher Resolve with lower downstream token usage and execution time, supporting improvements in both selection quality and computational efficiency. These advantages arise from how knowledge is represented and reused beyond storing parameterized workflows.

\begin{table}[!htb]
\caption{Quantitative ablation results on ComfyBench~\citep{comfybench}.
}
\label{tab:ablation_comfybench}
\centering

\begingroup
\fontsize{8pt}{10pt}\selectfont
\setlength{\tabcolsep}{1.0pt}
\renewcommand{\arraystretch}{0.90}
\setlength{\extrarowheight}{0.2pt}
\setlength{\aboverulesep}{0.6pt}
\setlength{\belowrulesep}{0.6pt}
\setlength{\cmidrulesep}{0.5pt}

\renewcommand{\tabularxcolumn}[1]{m{#1}}

\newcommand{\abHead}[1]{%
    {\fontsize{6pt}{7.4pt}\selectfont\bfseries\mbox{#1}}%
}

\definecolor{abrowone}{HTML}{F6F8FA}
\definecolor{abrowtwo}{HTML}{EFF4F8}
\definecolor{abrowthree}{HTML}{E7EFF5}
\definecolor{abrowfour}{HTML}{E0EBF3}
\definecolor{abrowfive}{HTML}{D8E6F1}
\definecolor{abrowsix}{HTML}{D1E2EE}
\definecolor{abrowseven}{HTML}{C9DDEC}

\begin{tabularx}{\linewidth}{
    l
    *{8}{>{\centering\arraybackslash}X}
}
\toprule

\multicolumn{1}{c}{\multirow[c]{2}{*}{\paperbold{Method}}}
& \multicolumn{2}{c}{\paperbold{Vanilla}}
& \multicolumn{2}{c}{\paperbold{Complex}}
& \multicolumn{2}{c}{\paperbold{Creative}}
& \multicolumn{2}{c}{\paperbold{Total}} \\

\cmidrule(lr){2-3}
\cmidrule(lr){4-5}
\cmidrule(lr){6-7}
\cmidrule(lr){8-9}

& \abHead{Pass\(\uparrow\)}
& \abHead{Res.\(\uparrow\)}
& \abHead{Pass\(\uparrow\)}
& \abHead{Res.\(\uparrow\)}
& \abHead{Pass\(\uparrow\)}
& \abHead{Res.\(\uparrow\)}
& \abHead{Pass\(\uparrow\)}
& \abHead{Res.\(\uparrow\)} \\
\midrule

\rowcolor{abrowone}
w/o Instance-Agnostic Abstraction
& \gecell{\paperbold{100.0}}{\textcolor{gray}{\ensuremath{=}0.0}}
& \gecell{91.0}{\gedown{6.0}}
& \gecell{95.0}{\gedown{5.0}}
& \gecell{65.0}{\gedown{18.3}}
& \gecell{95.0}{\gedown{5.0}}
& \gecell{70.0}{\gedown{25.0}}
& \gecell{97.5}{\gedown{2.5}}
& \gecell{79.0}{\gedown{13.5}} \\

\rowcolor{abrowtwo}
w/o Applicability Conditions
& \gecell{\paperbold{100.0}}{\textcolor{gray}{\ensuremath{=}0.0}}
& \gecell{95.0}{\gedown{2.0}}
& \gecell{\underline{98.3}}{\gedown{1.7}}
& \gecell{75.0}{\gedown{8.3}}
& \gecell{\underline{97.5}}{\gedown{2.5}}
& \gecell{82.5}{\gedown{12.5}}
& \gecell{\underline{99.0}}{\gedown{1.0}}
& \gecell{86.5}{\gedown{6.0}} \\

\rowcolor{abrowthree}
w/o Adaptive Instantiation
& \gecell{\underline{99.0}}{\gedown{1.0}}
& \gecell{90.0}{\gedown{7.0}}
& \gecell{95.0}{\gedown{5.0}}
& \gecell{70.0}{\gedown{13.3}}
& \gecell{92.5}{\gedown{7.5}}
& \gecell{75.0}{\gedown{20.0}}
& \gecell{96.5}{\gedown{3.5}}
& \gecell{81.0}{\gedown{11.5}} \\

\rowcolor{abrowfour}
w/o Workflow Composition
& \gecell{\paperbold{100.0}}{\textcolor{gray}{\ensuremath{=}0.0}}
& \gecell{\underline{96.0}}{\gedown{1.0}}
& \gecell{96.7}{\gedown{3.3}}
& \gecell{63.3}{\gedown{20.0}}
& \gecell{\underline{97.5}}{\gedown{2.5}}
& \gecell{85.0}{\gedown{10.0}}
& \gecell{98.5}{\gedown{1.5}}
& \gecell{84.0}{\gedown{8.5}} \\
\midrule

\rowcolor{abrowfive}
w/o Self-Directed Inquiry
& \gecell{\paperbold{100.0}}{\textcolor{gray}{\ensuremath{=}0.0}}
& \gecell{\underline{96.0}}{\gedown{1.0}}
& \gecell{\paperbold{100.0}}{\textcolor{gray}{\ensuremath{=}0.0}}
& \gecell{\underline{81.7}}{\gedown{1.6}}
& \gecell{\paperbold{100.0}}{\textcolor{gray}{\ensuremath{=}0.0}}
& \gecell{72.5}{\gedown{22.5}}
& \gecell{\paperbold{100.0}}{\textcolor{gray}{\ensuremath{=}0.0}}
& \gecell{87.0}{\gedown{5.5}} \\

\rowcolor{abrowsix}
w/o Online Policy Updates
& \gecell{\paperbold{100.0}}{\textcolor{gray}{\ensuremath{=}0.0}}
& \gecell{\underline{96.0}}{\gedown{1.0}}
& \gecell{\paperbold{100.0}}{\textcolor{gray}{\ensuremath{=}0.0}}
& \gecell{76.7}{\gedown{6.6}}
& \gecell{\paperbold{100.0}}{\textcolor{gray}{\ensuremath{=}0.0}}
& \gecell{\underline{87.5}}{\gedown{7.5}}
& \gecell{\paperbold{100.0}}{\textcolor{gray}{\ensuremath{=}0.0}}
& \gecell{\underline{88.5}}{\gedown{4.0}} \\
\midrule

\rowcolor{abrowseven}
\paperbold{OmniHarness}
& \paperbold{100.0}
& \paperbold{97.0}
& \paperbold{100.0}
& \paperbold{83.3}
& \paperbold{100.0}
& \paperbold{95.0}
& \paperbold{100.0}
& \paperbold{92.5} \\

\bottomrule
\end{tabularx}
\endgroup
\end{table}

\begin{table}[!htb]
\caption{Quantitative ablation study of symbolic retrieval.
$\dagger$ denotes agents initialized with the same library learned through OmniHarness's self-directed inquiry.}
\label{tab:symbolic_retrieval_ablation}
\centering

\begingroup
\fontsize{8pt}{10pt}\selectfont
\setlength{\tabcolsep}{1.0pt}
\renewcommand{\arraystretch}{0.90}
\setlength{\extrarowheight}{0.2pt}
\setlength{\aboverulesep}{0.6pt}
\setlength{\belowrulesep}{0.6pt}
\setlength{\cmidrulesep}{0.5pt}

\renewcommand{\tabularxcolumn}[1]{m{#1}}

\newcommand{\srHead}[1]{%
    {\fontsize{6pt}{7.4pt}\selectfont\bfseries\mbox{#1}}%
}

\newcommand{\srYes}{\textcolor{geloss}{\ding{51}}}
\newcommand{\srNo}{\textcolor{gegain}{\ding{55}}}

\definecolor{srrowone}{HTML}{F6F8FA}
\definecolor{srrowtwo}{HTML}{EFF4F8}
\definecolor{srrowthree}{HTML}{E7EFF5}
\definecolor{srrowfour}{HTML}{E0EBF3}
\definecolor{srrowfive}{HTML}{D8E6F1}
\definecolor{srrowsix}{HTML}{D1E2EE}
\definecolor{srrowseven}{HTML}{C9DDEC}

\begin{adjustbox}{max width=\linewidth}
\begin{tabular}{@{}cc*{9}{c}@{}}
\toprule
\multicolumn{2}{c}{\paperbold{Method}}
& \multicolumn{3}{c}{\paperbold{Vanilla}}
& \multicolumn{3}{c}{\paperbold{Complex}}
& \multicolumn{3}{c}{\paperbold{Creative}} \\

\cmidrule(lr){1-2}
\cmidrule(lr){3-5}
\cmidrule(lr){6-8}
\cmidrule(lr){9-11}

\srHead{Symbolic Retrieval}
& \srHead{Reliability Tier}
& \srHead{Res.\(\uparrow\)}
& \srHead{Tokens (K)\(\downarrow\)}
& \srHead{Time (s)\(\downarrow\)}
& \srHead{Res.\(\uparrow\)}
& \srHead{Tokens (K)\(\downarrow\)}
& \srHead{Time (s)\(\downarrow\)}
& \srHead{Res.\(\uparrow\)}
& \srHead{Tokens (K)\(\downarrow\)}
& \srHead{Time (s)\(\downarrow\)} \\
\midrule

\rowcolor{srrowone}
\srNo & \srNo
& \gecell{94.0}{\gedown{3.0}}
& \gecell{17.68}{\geup{9.86}}
& \gecell{70.6}{\geup{32.8}}
& \gecell{73.3}{\gedown{10.0}}
& \gecell{41.92}{\geup{26.56}}
& \gecell{143.8}{\geup{81.7}}
& \gecell{77.5}{\gedown{17.5}}
& \gecell{47.36}{\geup{29.81}}
& \gecell{170.2}{\geup{93.5}} \\

\rowcolor{srrowtwo}
\srYes & \srNo
& \gecell{\paperbold{97.0}}{\textcolor{gray}{\ensuremath{=}0.0}}
& \gecell{8.35}{\geup{0.53}}
& \gecell{40.3}{\geup{2.5}}
& \gecell{80.0}{\gedown{3.3}}
& \gecell{17.38}{\geup{2.02}}
& \gecell{73.3}{\geup{11.2}}
& \gecell{\underline{90.0}}{\gedown{5.0}}
& \gecell{19.46}{\geup{1.91}}
& \gecell{87.6}{\geup{10.9}} \\

\rowcolor{srrowthree}
\srNo & \srYes
& \gecell{95.0}{\gedown{2.0}}
& \gecell{15.42}{\geup{7.60}}
& \gecell{61.9}{\geup{24.1}}
& \gecell{78.3}{\gedown{5.0}}
& \gecell{36.87}{\geup{21.51}}
& \gecell{120.4}{\geup{58.3}}
& \gecell{85.0}{\gedown{10.0}}
& \gecell{40.94}{\geup{23.39}}
& \gecell{145.7}{\geup{69.0}} \\

\rowcolor{srrowfour}
\srYes & \srYes
& \paperbold{97.0}
& \paperbold{7.82}
& \paperbold{37.8}
& 83.3
& \underline{15.36}
& \underline{62.1}
& \paperbold{95.0}
& \paperbold{17.55}
& \paperbold{76.7} \\
\midrule

\rowcolor{srrowfive}
\multicolumn{2}{l}{ComfyAgent\(^{\dagger}\)~\citep{comfybench}}
& \gecell{69.0}{\gedown{28.0}}
& \gecell{21.87}{\geup{14.05}}
& \gecell{85.7}{\geup{47.9}}
& \gecell{46.7}{\gedown{36.6}}
& \gecell{48.62}{\geup{33.26}}
& \gecell{169.4}{\geup{107.3}}
& \gecell{42.5}{\gedown{52.5}}
& \gecell{55.34}{\geup{37.79}}
& \gecell{205.8}{\geup{129.1}} \\

\rowcolor{srrowsix}
\multicolumn{2}{l}{ComfyMind\(^{\dagger}\)~\citep{comfymind}}
& \gecell{94.0}{\gedown{3.0}}
& \gecell{9.64}{\geup{1.82}}
& \gecell{44.8}{\geup{7.0}}
& \gecell{\paperbold{86.7}}{\geup{3.4}}
& \gecell{\paperbold{15.18}}{\gedown{0.18}}
& \gecell{64.5}{\geup{2.4}}
& \gecell{75.0}{\gedown{20.0}}
& \gecell{21.83}{\geup{4.28}}
& \gecell{96.2}{\geup{19.5}} \\

\rowcolor{srrowseven}
\multicolumn{2}{l}{SymbOmni\(^{\dagger}\)~\citep{symbomni}}
& \gecell{\underline{96.0}}{\gedown{1.0}}
& \gecell{\underline{8.16}}{\geup{0.34}}
& \gecell{\underline{38.5}}{\geup{0.7}}
& \gecell{\underline{85.0}}{\geup{1.7}}
& \gecell{16.47}{\geup{1.11}}
& \gecell{\paperbold{60.8}}{\gedown{1.3}}
& \gecell{77.5}{\gedown{17.5}}
& \gecell{\underline{19.26}}{\geup{1.71}}
& \gecell{\underline{83.4}}{\geup{6.7}} \\

\bottomrule
\end{tabular}
\end{adjustbox}
\endgroup
\end{table}

\begin{figure}[!htb]
    \centering
    \includegraphics[width=\textwidth]{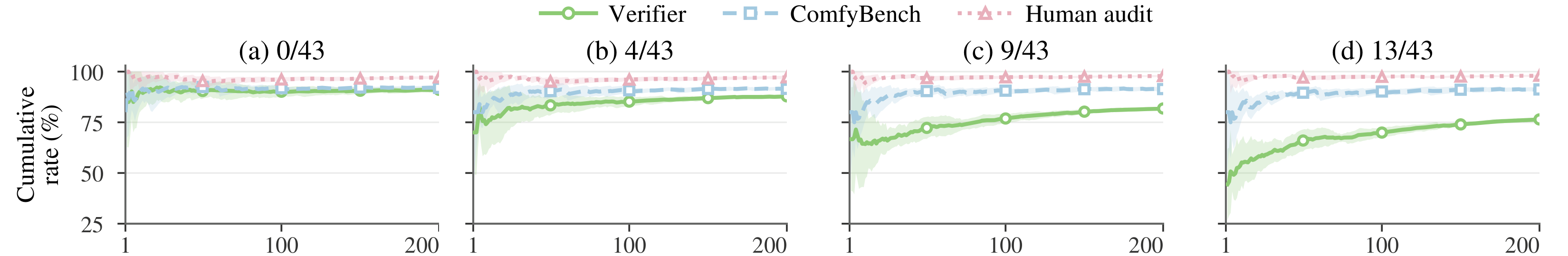}
    \caption{Reliability of verifier-guided policy learning under policy corruption. Panel labels indicate the number of corrupted policies out of 43. Curves and shaded regions represent means and standard deviations across five randomized task orders.}
    \label{fig:verifier_reliability}
\end{figure}

\paperbold{Reliability of Verifier-Guided Policy Learning.}
We evaluate four levels of policy corruption across five randomized orders of 200 ComfyBench tasks. Figure~\ref{fig:verifier_reliability} reports means and standard deviations across runs. The \emph{Verifier} curve measures the cumulative proportion of policy reuse attempts judged successful, while \emph{ComfyBench} reports the cumulative official Resolve rate. \emph{Human audit} measures the cumulative accuracy of verifier decisions against independent human judgments. Stronger corruption reduces the proportion of reuse attempts judged successful, yet human-audited verifier accuracy remains high and official Resolve rates remain stable. These results support accurate verification and robust policy learning under noisy stored knowledge. Limited variation across randomized task orders further indicates that performance is robust to the tested execution orders.

\begin{figure}[!htb]
    \centering
    \includegraphics[width=\textwidth]{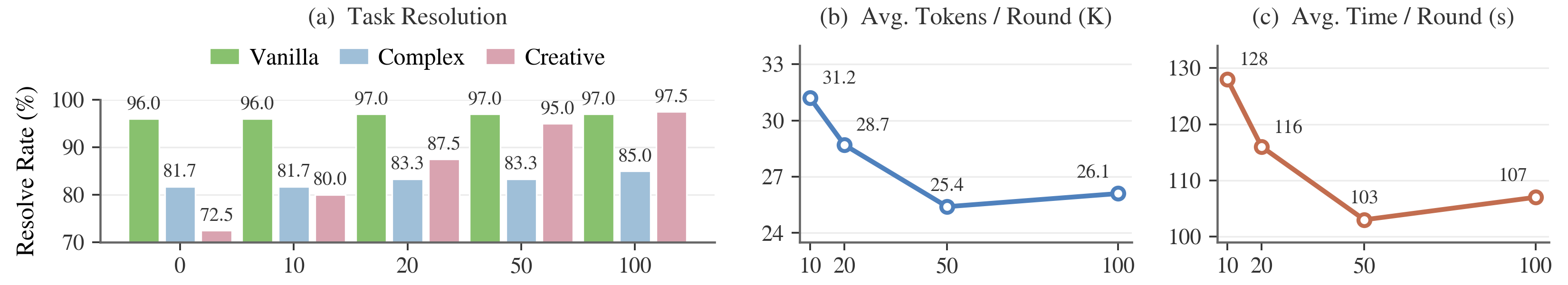}
    \caption{Effect of the self-directed inquiry budget on performance and cost. The horizontal axis indicates the number of inquiry rounds.}
    \label{fig:play_rounds_ablation}
\end{figure}

\paperbold{Efficiency Analysis.}
Figure~\ref{fig:play_rounds_ablation} shows that 50 inquiry rounds increase Creative Resolve from 72.5\% to 95.0\%, averaging 25.4K tokens and 103\,s per round. Increasing the budget to 100 rounds more than doubles the inquiry cost but yields only modest additional gains, supporting 50 rounds as a practical tradeoff in the current exploration environment. Table~\ref{tab:symbolic_retrieval_ablation} shows that, compared with disabling both mechanisms, symbolic retrieval with reliability prioritization improves Resolve while reducing downstream token usage by over 55\% and execution time by over 46\% across all subsets.

\paperbold{Additional Component Ablations.}
Self-directed inquiry targets specific generation capabilities, providing broader capability coverage that benefits Creative tasks more directly. Table~\ref{tab:play_fine_grained_ablation} supports the contributions of capability novelty and the competence frontier to this benefit. Complex tasks additionally require workflow composition and execution coordination, which the inquiry curriculum does not explicitly practice. Table~\ref{tab:ablation_comfybench} shows that Complex performance is more sensitive to online updates than to prior inquiry, while Table~\ref{tab:execution_fine_grained_ablation} highlights the importance of planning, intermediate verification, and localized recovery.

\begin{table}[!htb]
\centering
\begingroup

\setlength{\tabcolsep}{0.7pt}
\renewcommand{\arraystretch}{0.90}
\setlength{\extrarowheight}{0.2pt}
\setlength{\aboverulesep}{0.6pt}
\setlength{\belowrulesep}{0.6pt}
\setlength{\cmidrulesep}{0.5pt}
\renewcommand{\tabularxcolumn}[1]{m{#1}}

\definecolor{paRowOne}{HTML}{EFF4F8}
\definecolor{paRowTwo}{HTML}{E7EFF5}
\definecolor{paRowThree}{HTML}{D8E6F1}
\definecolor{paRowFull}{HTML}{C9DDEC}
\definecolor{paDownColor}{HTML}{006400}

\newcommand{\paHead}[1]{%
    {\fontsize{6pt}{7.4pt}\selectfont\bfseries\mbox{#1}}%
}
\newcommand{\paCell}[2]{%
    \mbox{#1\kern0.25pt%
    {\normalfont\fontsize{6pt}{7.4pt}\selectfont #2}}%
}
\newcommand{\paDown}[1]{%
    \textcolor{paDownColor}{\ensuremath{\downarrow}#1}%
}
\newcommand{\paEqual}{%
    \textcolor{gray}{\ensuremath{=}0.0}%
}

\begin{minipage}[t]{0.475\textwidth}
\vspace{0pt}
\caption{Task selection ablations on the Creative subset.\strut}
\label{tab:play_fine_grained_ablation}
\centering
\fontsize{8pt}{10pt}\selectfont

\begin{tabularx}{\linewidth}{
    @{}>{\raggedright\arraybackslash}m{0.29\linewidth}*{4}{>{\centering\arraybackslash}X}@{}
}
\toprule
\multicolumn{1}{c}{\multirow[c]{2}{*}{\paperbold{Method}}}
& \multicolumn{2}{c}{\paperbold{Creative T2I}}
& \multicolumn{2}{c}{\paperbold{Creative I2I}} \\
\cmidrule(lr){2-3}
\cmidrule(lr){4-5}
& \paHead{Pass\(\uparrow\)}
& \paHead{Res.\(\uparrow\)}
& \paHead{Pass\(\uparrow\)}
& \paHead{Res.\(\uparrow\)} \\
\midrule

\rowcolor{paRowOne}
w/o Capability Novelty
& \paCell{\paperbold{100.0}}{\paEqual}
& \paCell{\underline{66.7}}{\paDown{16.6}}
& \paCell{\paperbold{100.0}}{\paEqual}
& \paCell{\underline{86.7}}{\paDown{6.6}} \\

\rowcolor{paRowThree}
w/o Competence Frontier
& \paCell{\paperbold{100.0}}{\paEqual}
& \paCell{\paperbold{83.3}}{\paEqual}
& \paCell{\paperbold{100.0}}{\paEqual}
& \paCell{80.0}{\paDown{13.3}} \\

\rowcolor{paRowFull}
\paperbold{OmniHarness}
& \paperbold{100.0}
& \paperbold{83.3}
& \paperbold{100.0}
& \paperbold{93.3} \\

\bottomrule
\end{tabularx}
\end{minipage}\hfill
\begin{minipage}[t]{0.505\textwidth}
\vspace{0pt}
\caption{Feedback-guided execution ablations on the Complex subset.\strut}
\label{tab:execution_fine_grained_ablation}
\centering
\fontsize{8pt}{10pt}\selectfont

\begin{tabularx}{\linewidth}{
    @{}>{\raggedright\arraybackslash}m{0.29\linewidth}*{4}{>{\centering\arraybackslash}X}@{}
}
\toprule
\multicolumn{1}{c}{\multirow[c]{2}{*}{\paperbold{Method}}}
& \multicolumn{2}{c}{\paperbold{Complex T2I}}
& \multicolumn{2}{c}{\paperbold{Complex I2I}} \\
\cmidrule(lr){2-3}
\cmidrule(lr){4-5}
& \paHead{Pass\(\uparrow\)}
& \paHead{Res.\(\uparrow\)}
& \paHead{Pass\(\uparrow\)}
& \paHead{Res.\(\uparrow\)} \\
\midrule

\rowcolor{paRowOne}
w/o Planning
& \paCell{80.0}{\paDown{20.0}}
& \paCell{40.0}{\paDown{50.0}}
& \paCell{87.0}{\paDown{13.0}}
& \paCell{56.5}{\paDown{34.8}} \\

\rowcolor{paRowTwo}
w/o Intermediate Verification
& \paCell{70.0}{\paDown{30.0}}
& \paCell{60.0}{\paDown{30.0}}
& \paCell{73.9}{\paDown{26.1}}
& \paCell{69.6}{\paDown{21.7}} \\

\rowcolor{paRowThree}
w/o Localized Recovery
& \paCell{\underline{90.0}}{\paDown{10.0}}
& \paCell{\underline{80.0}}{\paDown{10.0}}
& \paCell{\underline{95.7}}{\paDown{4.3}}
& \paCell{\underline{82.6}}{\paDown{8.7}} \\

\rowcolor{paRowFull}
\paperbold{OmniHarness}
& \paperbold{100.0}
& \paperbold{90.0}
& \paperbold{100.0}
& \paperbold{91.3} \\

\bottomrule
\end{tabularx}
\end{minipage}

\par
\endgroup
\end{table}

\section{Conclusion}

We introduce OmniHarness, a framework for generalizable visual generation via symbolic policy learning. Before downstream objectives are specified, self-directed inquiry selects practice tasks based on capability novelty and the current competence frontier. Verified workflows are abstracted into symbolic policies for task families. Symbolic retrieval uses workflow metadata and reliability tiers to guide reuse, while execution feedback continually updates the policies with model parameters fixed. Experiments across six benchmarks demonstrate the framework's effectiveness. On ComfyBench's Creative tasks, OmniHarness achieves a 95.0\% Resolve rate, exceeding the strongest baseline by 27.5 percentage points. Self-directed inquiry currently operates within a predefined capability space and primarily acquires policies for individual capability categories. It does not explicitly practice compositions across these categories. Tasks such as ComfyBench's Complex tasks therefore require combining and adapting existing policies, with additional reasoning to resolve dependencies between workflow steps. Future work could extend exploration to broader capabilities and compositional tasks, while enabling agents to define and expand their own capability spaces. Verification also relies on VLM judgments, making policy learning dependent on the model's visual understanding. Incorrect judgments can distort policy reliability estimates, promote unreliable policies, or suppress useful ones, affecting subsequent retrieval and reuse. Future work should explore more reliable verification mechanisms to reduce the impact of uncertain feedback across the system.

\FloatBarrier
\bibliographystyle{assets/plainnat}
\bibliography{main}

\begin{thebibliography}{61}
\providecommand{\natexlab}[1]{#1}
\providecommand{\url}[1]{\texttt{#1}}
\expandafter\ifx\csname urlstyle\endcsname\relax
  \providecommand{\doi}[1]{doi: #1}\else
  \providecommand{\doi}{doi: \begingroup \urlstyle{rm}\Url}\fi

\bibitem[Baranes and Oudeyer(2013)]{BARANES201349}
Adrien Baranes and Pierre-Yves Oudeyer.
\newblock Active learning of inverse models with intrinsically motivated goal
  exploration in robots.
\newblock \emph{Robotics and Autonomous Systems}, 61\penalty0 (1):\penalty0
  49--73, 2013.
\newblock ISSN 0921-8890.
\newblock \doi{10.1016/j.robot.2012.05.008}.

\bibitem[Brooks et~al.(2023)Brooks, Holynski, and Efros]{InstructPix2Pix}
Tim Brooks, Aleksander Holynski, and Alexei~A. Efros.
\newblock {InstructPix2Pix}: Learning to follow image editing instructions.
\newblock In \emph{Proceedings of the IEEE/CVF Conference on Computer Vision
  and Pattern Recognition (CVPR)}, pages 18392--18402, June 2023.
\newblock \doi{10.1109/CVPR52729.2023.01764}.

\bibitem[Brown et~al.(2020)Brown, Mann, Ryder, Subbiah, Kaplan, Dhariwal,
  Neelakantan, Shyam, Sastry, Askell, et~al.]{Few-Shot}
Tom Brown, Benjamin Mann, Nick Ryder, Melanie Subbiah, Jared~D Kaplan, Prafulla
  Dhariwal, Arvind Neelakantan, Pranav Shyam, Girish Sastry, Amanda Askell,
  et~al.
\newblock Language models are few-shot learners.
\newblock In H.~Larochelle, M.~Ranzato, R.~Hadsell, M.F. Balcan, and H.~Lin,
  editors, \emph{Advances in Neural Information Processing Systems}, volume~33,
  pages 1877--1901. Curran Associates, Inc., 2020.

\bibitem[Chen et~al.(2024)Chen, Yu, Ge, Yao, Xie, Wang, Kwok, Luo, Lu, and
  Li]{PixArt}
Junsong Chen, Jincheng Yu, Chongjian Ge, Lewei Yao, Enze Xie, Zhongdao Wang,
  James Kwok, Ping Luo, Huchuan Lu, and Zhenguo Li.
\newblock {PixArt-$\alpha$}: Fast training of diffusion transformer for
  photorealistic text-to-image synthesis.
\newblock In B.~Kim, Y.~Yue, S.~Chaudhuri, K.~Fragkiadaki, M.~Khan, and Y.~Sun,
  editors, \emph{International Conference on Learning Representations}, volume
  2024, pages 57611--57640, 2024.

\bibitem[Chen et~al.(2026{\natexlab{a}})Chen, Ma, Fan, Huang, Sinha, Dai, Wang,
  He, Yang, Li, et~al.]{UniT}
Leon~Liangyu Chen, Haoyu Ma, Zhipeng Fan, Ziqi Huang, Animesh Sinha, Xiaoliang
  Dai, Jialiang Wang, Zecheng He, Jianwei Yang, Chunyuan Li, et~al.
\newblock {UniT}: Unified multimodal chain-of-thought test-time scaling.
\newblock \emph{arXiv preprint arXiv:2602.12279}, 2026{\natexlab{a}}.

\bibitem[Chen et~al.(2026{\natexlab{b}})Chen, Xing, Ye, Geng, Lin, Lai, He,
  Zhai, Gao, and Zhu]{genevolve}
Sixiang Chen, Zhaohu Xing, Tian Ye, Xinyu Geng, Yunlong Lin, Jianyu Lai,
  Xuanhua He, Fuxiang Zhai, Jialin Gao, and Lei Zhu.
\newblock {GenEvolve}: Self-evolving image generation agents via
  tool-orchestrated visual experience distillation.
\newblock \emph{arXiv preprint arXiv:2605.21605}, 2026{\natexlab{b}}.

\bibitem[Chen et~al.(2025)Chen, Wu, Liu, Pan, Liu, Xie, Yu, and
  Ruan]{Janus-Pro}
Xiaokang Chen, Zhiyu Wu, Xingchao Liu, Zizheng Pan, Wen Liu, Zhenda Xie,
  Xingkai Yu, and Chong Ruan.
\newblock {Janus-Pro}: Unified multimodal understanding and generation with
  data and model scaling.
\newblock \emph{arXiv preprint arXiv:2501.17811}, 2025.

\bibitem[Comanici et~al.(2025)Comanici, Bieber, Schaekermann, Pasupat,
  Sachdeva, Dhillon, Blistein, Ram, Zhang, Rosen, et~al.]{gemini25}
Gheorghe Comanici, Eric Bieber, Mike Schaekermann, Ice Pasupat, Noveen
  Sachdeva, Inderjit Dhillon, Marcel Blistein, Ori Ram, Dan Zhang, Evan Rosen,
  et~al.
\newblock Gemini 2.5: Pushing the frontier with advanced reasoning,
  multimodality, long context, and next generation agentic capabilities.
\newblock \emph{arXiv preprint arXiv:2507.06261}, 2025.

\bibitem[Deng et~al.(2025)Deng, Zhu, Li, Gou, Li, Wang, Zhong, Yu, Nie, Song,
  et~al.]{bagel}
Chaorui Deng, Deyao Zhu, Kunchang Li, Chenhui Gou, Feng Li, Zeyu Wang, Shu
  Zhong, Weihao Yu, Xiaonan Nie, Ziang Song, et~al.
\newblock Emerging properties in unified multimodal pretraining.
\newblock \emph{arXiv preprint arXiv:2505.14683}, 2025.

\bibitem[Esser et~al.(2024)Esser, Kulal, Blattmann, Entezari, M\"{u}ller,
  Saini, Levi, Lorenz, Sauer, Boesel, et~al.]{SD3}
Patrick Esser, Sumith Kulal, Andreas Blattmann, Rahim Entezari, Jonas
  M\"{u}ller, Harry Saini, Yam Levi, Dominik Lorenz, Axel Sauer, Frederic
  Boesel, et~al.
\newblock Scaling rectified flow transformers for high-resolution image
  synthesis.
\newblock In Ruslan Salakhutdinov, Zico Kolter, Katherine Heller, Adrian
  Weller, Nuria Oliver, Jonathan Scarlett, and Felix Berkenkamp, editors,
  \emph{Proceedings of the 41st International Conference on Machine Learning},
  volume 235 of \emph{Proceedings of Machine Learning Research}, pages
  12606--12633. PMLR, 21--27 Jul 2024.

\bibitem[Fang et~al.(2025)Fang, Duan, Wang, Huang, Li, Tian, Yan, Yu, Zeng,
  Dai, et~al.]{got}
Rongyao Fang, Chengqi Duan, Kun Wang, Linjiang Huang, Hao Li, Hao Tian, Shilin
  Yan, Weihao Yu, Xingyu Zeng, Jifeng Dai, et~al.
\newblock {GoT}: Unleashing reasoning capability of {MLLM} for visual
  generation and editing.
\newblock In D.~Belgrave, C.~Zhang, H.~Lin, R.~Pascanu, P.~Koniusz,
  M.~Ghassemi, and N.~Chen, editors, \emph{Advances in Neural Information
  Processing Systems}, volume 38, Main Conference, pages 67680--67708. Curran
  Associates, Inc., 2025.
\newblock \doi{10.52202/085713-2270}.

\bibitem[Ge et~al.(2025)Ge, Zhao, Zhu, Ge, Yi, Song, Li, Ding, and Shan]{seedx}
Yuying Ge, Sijie Zhao, Jinguo Zhu, Yixiao Ge, Kun Yi, Lin Song, Chen Li,
  Xiaohan Ding, and Ying Shan.
\newblock {SEED-X}: Multimodal models with unified multi-granularity
  comprehension and generation.
\newblock \emph{arXiv preprint arXiv:2404.14396}, 2025.

\bibitem[Geng et~al.(2024)Geng, Yang, Hang, Li, Gu, Zhang, Bao, Zhang, Li, Hu,
  et~al.]{InstructDiffusion}
Zigang Geng, Binxin Yang, Tiankai Hang, Chen Li, Shuyang Gu, Ting Zhang,
  Jianmin Bao, Zheng Zhang, Houqiang Li, Han Hu, et~al.
\newblock {InstructDiffusion}: A generalist modeling interface for vision
  tasks.
\newblock In \emph{Proceedings of the IEEE/CVF Conference on Computer Vision
  and Pattern Recognition (CVPR)}, pages 12709--12720, June 2024.
\newblock \doi{10.1109/CVPR52733.2024.01208}.

\bibitem[Ghosh et~al.(2023)Ghosh, Hajishirzi, and Schmidt]{geneval}
Dhruba Ghosh, Hannaneh Hajishirzi, and Ludwig Schmidt.
\newblock {GenEval}: An object-focused framework for evaluating text-to-image
  alignment.
\newblock In A.~Oh, T.~Naumann, A.~Globerson, K.~Saenko, M.~Hardt, and
  S.~Levine, editors, \emph{Advances in Neural Information Processing Systems},
  volume~36, pages 52132--52152. Curran Associates, Inc., 2023.
\newblock \doi{10.52202/075280-2270}.

\bibitem[Guo et~al.(2025)Guo, Xu, Wang, Lin, Zhou, Zhang, Su, and
  Chen]{comfymind}
Litao Guo, Xinli Xu, Luozhou Wang, Jiantao Lin, Jinsong Zhou, Zixin Zhang,
  Bolan Su, and Yingcong Chen.
\newblock {ComfyMind}: Toward general-purpose generation via tree-based
  planning and reactive feedback.
\newblock In D.~Belgrave, C.~Zhang, H.~Lin, R.~Pascanu, P.~Koniusz,
  M.~Ghassemi, and N.~Chen, editors, \emph{Advances in Neural Information
  Processing Systems}, volume 38, Main Conference, pages 45128--45164. Curran
  Associates, Inc., 2025.
\newblock \doi{10.52202/085713-1503}.

\bibitem[Han et~al.(2026)Han, Ji, Xin, Shi, Qiu, Ye, Xia, Liu, Chen, Zhou,
  et~al.]{paper2figure}
Siwei Han, Haonian Ji, Siyang Xin, Juanquan Shi, Shi Qiu, Xinyu Ye, Peng Xia,
  Jiaqi Liu, Zhaorun Chen, Yiyang Zhou, et~al.
\newblock {Paper2Figure}: A multi-agent collaborative system for figure
  generation towards academic research paper.
\newblock In \emph{Proceedings of the IEEE/CVF Conference on Computer Vision
  and Pattern Recognition (CVPR)}, pages 29157--29166, June 2026.

\bibitem[Hao et~al.(2026)Hao, Wang, Luo, Zhang, Zhou, Lin, Wang, Dong, and
  Chen]{hao-etal-2026-recreate}
Zhezheng Hao, Hong Wang, Jian Luo, Jianqing Zhang, Yuyan Zhou, Qiang Lin, Can
  Wang, Hande Dong, and Jiawei Chen.
\newblock {R}e{C}reate: Reasoning and creating domain agents driven by
  experience.
\newblock In Maria Liakata, Viviane~P. Moreira, Jiajun Zhang, and David
  Jurgens, editors, \emph{Proceedings of the 64th Annual Meeting of the
  {A}ssociation for {C}omputational {L}inguistics (Volume 1: Long Papers)},
  pages 31018--31046, San Diego, California, United States, July 2026.
  Association for Computational Linguistics.
\newblock ISBN 979-8-89176-390-6.
\newblock \doi{10.18653/v1/2026.acl-long.1432}.

\bibitem[Huang et~al.(2025)Huang, Ma, Zhao, Wu, Ji, Zhang, Hu, Sun, and
  Ji]{comfygpt}
Oucheng Huang, Yuhang Ma, Zeng Zhao, Mingrui Wu, Jiayi Ji, Rongsheng Zhang,
  Zhipeng Hu, Xiaoshuai Sun, and Rongrong Ji.
\newblock {ComfyGPT}: A self-optimizing multi-agent system for comprehensive
  {ComfyUI} workflow generation.
\newblock \emph{arXiv preprint arXiv:2503.17671}, 2025.

\bibitem[Huang et~al.(2024)Huang, Xie, Wang, Yuan, Cun, Ge, Zhou, Dong, Huang,
  Zhang, et~al.]{smartedit}
Yuzhou Huang, Liangbin Xie, Xintao Wang, Ziyang Yuan, Xiaodong Cun, Yixiao Ge,
  Jiantao Zhou, Chao Dong, Rui Huang, Ruimao Zhang, et~al.
\newblock {SmartEdit}: Exploring complex instruction-based image editing with
  multimodal large language models.
\newblock In \emph{Proceedings of the IEEE/CVF Conference on Computer Vision
  and Pattern Recognition (CVPR)}, pages 8362--8371, June 2024.
\newblock \doi{10.1109/CVPR52733.2024.00799}.

\bibitem[Kamath et~al.(2025)Kamath, Chang, Krishna, Zettlemoyer, Hu, and
  Ghazvininejad]{geneval2}
Amita Kamath, Kai-Wei Chang, Ranjay Krishna, Luke Zettlemoyer, Yushi Hu, and
  Marjan Ghazvininejad.
\newblock {GenEval} 2: Addressing benchmark drift in text-to-image evaluation.
\newblock \emph{arXiv preprint arXiv:2512.16853}, 2025.

\bibitem[Kidd et~al.(2012)Kidd, Piantadosi, and
  Aslin]{10.1371/journal.pone.0036399}
Celeste Kidd, Steven~T. Piantadosi, and Richard~N. Aslin.
\newblock The {Goldilocks} effect: Human infants allocate attention to visual
  sequences that are neither too simple nor too complex.
\newblock \emph{PLOS ONE}, 7\penalty0 (5):\penalty0 e36399, May 2012.
\newblock \doi{10.1371/journal.pone.0036399}.

\bibitem[Kou et~al.(2025)Kou, Jin, Liu, Liu, Ma, Jia, Chen, Jiang, and
  Deng]{orthus}
Siqi Kou, Jiachun Jin, Zhihong Liu, Chang Liu, Ye~Ma, Jian Jia, Quan Chen, Peng
  Jiang, and Zhijie Deng.
\newblock Orthus: Autoregressive interleaved image-text generation with
  modality-specific heads.
\newblock \emph{arXiv preprint arXiv:2412.00127}, 2025.

\bibitem[Labs et~al.(2025)Labs, Batifol, Blattmann, Boesel, Consul, Diagne,
  Dockhorn, English, English, Esser, et~al.]{flux.1-dev}
Black~Forest Labs, Stephen Batifol, Andreas Blattmann, Frederic Boesel, Saksham
  Consul, Cyril Diagne, Tim Dockhorn, Jack English, Zion English, Patrick
  Esser, et~al.
\newblock {FLUX.1 Kontext}: Flow matching for in-context image generation and
  editing in latent space.
\newblock \emph{arXiv preprint arXiv:2506.15742}, 2025.

\bibitem[Lewis et~al.(2020)Lewis, Perez, Piktus, Petroni, Karpukhin, Goyal,
  K\"{u}ttler, Lewis, Yih, Rockt\"{a}schel, et~al.]{RAG}
Patrick Lewis, Ethan Perez, Aleksandra Piktus, Fabio Petroni, Vladimir
  Karpukhin, Naman Goyal, Heinrich K\"{u}ttler, Mike Lewis, Wen-tau Yih, Tim
  Rockt\"{a}schel, et~al.
\newblock Retrieval-augmented generation for knowledge-intensive {NLP} tasks.
\newblock In H.~Larochelle, M.~Ranzato, R.~Hadsell, M.F. Balcan, and H.~Lin,
  editors, \emph{Advances in Neural Information Processing Systems}, volume~33,
  pages 9459--9474. Curran Associates, Inc., 2020.

\bibitem[Li et~al.(2026{\natexlab{a}})Li, Wu, Pan, Hui, Hu, Jiang, Sheng, Liu,
  Gong, and Liu]{coDrawAgents}
Chunhan Li, Qifeng Wu, Jia-Hui Pan, Ka-Hei Hui, Jingyu Hu, Yuming Jiang, Bin
  Sheng, Xihui Liu, Wenjuan Gong, and Zhengzhe Liu.
\newblock {coDrawAgents}: A multi-agent dialogue framework for compositional
  image generation.
\newblock In \emph{Proceedings of the IEEE/CVF Conference on Computer Vision
  and Pattern Recognition (CVPR) Findings}, pages 9802--9812, June
  2026{\natexlab{a}}.
\newblock \doi{10.1109/CVPRF71909.2026.00938}.

\bibitem[Li et~al.(2024)Li, Kamko, Akhgari, Sabet, Xu, and
  Doshi]{playgroundv25}
Daiqing Li, Aleks Kamko, Ehsan Akhgari, Ali Sabet, Linmiao Xu, and Suhail
  Doshi.
\newblock Playground v2.5: Three insights towards enhancing aesthetic quality
  in text-to-image generation.
\newblock \emph{arXiv preprint arXiv:2402.17245}, 2024.

\bibitem[Li et~al.(2026{\natexlab{b}})Li, Liu, Liu, Zhou, Wu, Chen, Xie, Wu,
  and Sun]{comfyclaw}
Zongxia Li, Dawei Liu, Fuxiao Liu, Yuhang Zhou, Xiyang Wu, Jingxi Chen, Jing
  Xie, Xiaomin Wu, and Lichao Sun.
\newblock {COMFYCLAW}: Self-evolving skill harnesses for image generation
  workflows.
\newblock \emph{arXiv preprint arXiv:2607.01709}, 2026{\natexlab{b}}.

\bibitem[Liang et~al.(2025)Liang, Wu, Li, Wei, He, Guo, and Xie]{MM-R1}
Qian Liang, Yujia Wu, Kuncheng Li, Jiwei Wei, Shiyuan He, Jinyu Guo, and Ning
  Xie.
\newblock {MM-R1}: Unleashing the power of unified multimodal large language
  models for personalized image generation.
\newblock \emph{arXiv preprint arXiv:2508.11433}, 2025.

\bibitem[Liu et~al.(2025{\natexlab{a}})Liu, Yan, Zaharia, and Abbeel]{LWM}
Hao Liu, Wilson Yan, Matei Zaharia, and Pieter Abbeel.
\newblock World model on million-length video and language with blockwise
  {RingAttention}.
\newblock In Y.~Yue, A.~Garg, N.~Peng, F.~Sha, and R.~Yu, editors,
  \emph{International Conference on Learning Representations}, volume 2025,
  pages 45953--45977, 2025{\natexlab{a}}.

\bibitem[Liu et~al.(2026)Liu, Li, Kuang, Liu, Mei, Wen, and Liu]{symbomni}
Jinxiu Liu, Jianru Li, Tanqing Kuang, Xuanming Liu, Kangfu Mei, Yandong Wen,
  and Weiyang Liu.
\newblock {SymbOmni}: Evolving agentic omni models via symbolic concept
  learning.
\newblock In \emph{Proceedings of the European Conference on Computer Vision
  (ECCV)}, volume 17048 of \emph{Lecture Notes in Computer Science}, pages
  188--206. Springer Nature Switzerland, 2026.
\newblock \doi{10.1007/978-3-032-37372-4_11}.

\bibitem[Liu et~al.(2025{\natexlab{b}})Liu, Han, Xing, Yin, Wang, Cheng, Liao,
  Wang, Fu, Han, et~al.]{Step1X-Edit}
Shiyu Liu, Yucheng Han, Peng Xing, Fukun Yin, Rui Wang, Wei Cheng, Jiaqi Liao,
  Yingming Wang, Honghao Fu, Chunrui Han, et~al.
\newblock {Step1X-Edit}: A practical framework for general image editing.
\newblock \emph{arXiv preprint arXiv:2504.17761}, 2025{\natexlab{b}}.

\bibitem[Ma et~al.(2025)Ma, Liu, Chen, Liu, Wu, Wu, Pan, Xie, Zhang, Yu,
  et~al.]{JanusFlow}
Yiyang Ma, Xingchao Liu, Xiaokang Chen, Wen Liu, Chengyue Wu, Zhiyu Wu, Zizheng
  Pan, Zhenda Xie, Haowei Zhang, Xingkai Yu, et~al.
\newblock {JanusFlow}: Harmonizing autoregression and rectified flow for
  unified multimodal understanding and generation.
\newblock In \emph{Proceedings of the IEEE/CVF Conference on Computer Vision
  and Pattern Recognition (CVPR)}, pages 7739--7751, June 2025.
\newblock \doi{10.1109/CVPR52734.2025.00725}.

\bibitem[Niu et~al.(2026)Niu, Ning, Zheng, Jin, Lin, Jin, Liao, Feng, Meng,
  Ning, et~al.]{WISE}
Yuwei Niu, Munan Ning, Mengren Zheng, Weiyang Jin, Bin Lin, Peng Jin, Jiaqi
  Liao, Chaoran Feng, Fanqing Meng, Kun-Peng Ning, et~al.
\newblock {WISE}: World knowledge-informed semantic evaluation for
  text-to-image generation.
\newblock In Tong Zhang, Miroslav Dudik, Martin Jaggi, Alekh Agarwal, Sharon
  Li, Dale Schuurmans, Jerry Zhu, Felix Berkenkamp, Hanze Dong, and Alberto
  Bietti, editors, \emph{Proceedings of the 43rd International Conference on
  Machine Learning}, volume 306 of \emph{Proceedings of Machine Learning
  Research}, pages 93570--93590. PMLR, 06--11 Jul 2026.

\bibitem[Peng et~al.(2026)Peng, Meng, Cai, Zhuang, Yang, Fang, Wu, Lin, Wu, and
  Bai]{peng2026unifiedmultimodalautoregressivemodeling}
Wujian Peng, Lingchen Meng, Yuxuan Cai, Xianwei Zhuang, Yuhuan Yang, Rongyao
  Fang, Chenfei Wu, Junyang Lin, Zuxuan Wu, and Shuai Bai.
\newblock Unified multimodal autoregressive modeling with shared context-visual
  tokenizer is key to unification.
\newblock \emph{arXiv preprint arXiv:2606.18249}, 2026.

\bibitem[Podell et~al.(2024)Podell, English, Lacey, Blattmann, Dockhorn,
  M\"{u}ller, Penna, and Rombach]{SDXL}
Dustin Podell, Zion English, Kyle Lacey, Andreas Blattmann, Tim Dockhorn, Jonas
  M\"{u}ller, Joe Penna, and Robin Rombach.
\newblock {SDXL}: Improving latent diffusion models for high-resolution image
  synthesis.
\newblock In B.~Kim, Y.~Yue, S.~Chaudhuri, K.~Fragkiadaki, M.~Khan, and Y.~Sun,
  editors, \emph{International Conference on Learning Representations}, volume
  2024, pages 1862--1874, 2024.

\bibitem[Ramesh et~al.(2022)Ramesh, Dhariwal, Nichol, Chu, and Chen]{DALLE}
Aditya Ramesh, Prafulla Dhariwal, Alex Nichol, Casey Chu, and Mark Chen.
\newblock Hierarchical text-conditional image generation with {CLIP} latents.
\newblock \emph{arXiv preprint arXiv:2204.06125}, 2022.

\bibitem[Su et~al.(2026)Su, Wei, Cen, Wang, Chen, Yuan, and
  Chu]{su2026generationenhancesunderstandingunified}
Zihan Su, Hongyang Wei, Kangrui Cen, Yong Wang, Guanhua Chen, Chun Yuan, and
  Xiangxiang Chu.
\newblock Generation enhances understanding in unified multimodal models via
  multi-representation generation.
\newblock \emph{arXiv preprint arXiv:2601.21406}, 2026.

\bibitem[Sun et~al.(2024{\natexlab{a}})Sun, Jiang, Chen, Zhang, Peng, Luo, and
  Yuan]{llamagen}
Peize Sun, Yi~Jiang, Shoufa Chen, Shilong Zhang, Bingyue Peng, Ping Luo, and
  Zehuan Yuan.
\newblock Autoregressive model beats diffusion: {Llama} for scalable image
  generation.
\newblock \emph{arXiv preprint arXiv:2406.06525}, 2024{\natexlab{a}}.

\bibitem[Sun et~al.(2024{\natexlab{b}})Sun, Cui, Zhang, Zhang, Yu, Wang, Rao,
  Liu, Huang, and Wang]{emu2}
Quan Sun, Yufeng Cui, Xiaosong Zhang, Fan Zhang, Qiying Yu, Yueze Wang,
  Yongming Rao, Jingjing Liu, Tiejun Huang, and Xinlong Wang.
\newblock Generative multimodal models are in-context learners.
\newblock In \emph{Proceedings of the IEEE/CVF Conference on Computer Vision
  and Pattern Recognition (CVPR)}, pages 14398--14409, June 2024{\natexlab{b}}.
\newblock \doi{10.1109/CVPR52733.2024.01365}.

\bibitem[Wang et~al.(2026)Wang, Zhang, Ye, Deng, Wang, Hu, Guo, Liu, and
  Guo]{MAS2}
Kun Wang, Guibin Zhang, ManKit Ye, Xinyu Deng, Dongxia Wang, Xiaobin Hu,
  Jinyang Guo, Yang Liu, and Yufei Guo.
\newblock {MAS}$^{2}$: Self-generative, self-configuring, self-rectifying
  multi-agent systems.
\newblock In C.~Vondrick, B.~Hariharan, C.~Raffel, L.~Pinto, D.~Yang, and
  A.~Faust, editors, \emph{International Conference on Learning
  Representations}, volume 2026, pages 113586--113613, 2026.

\bibitem[Wang et~al.(2024)Wang, Zhang, Luo, Sun, Cui, Wang, Zhang, Wang, Li,
  Yu, et~al.]{emu3}
Xinlong Wang, Xiaosong Zhang, Zhengxiong Luo, Quan Sun, Yufeng Cui, Jinsheng
  Wang, Fan Zhang, Yueze Wang, Zhen Li, Qiying Yu, et~al.
\newblock Emu3: Next-token prediction is all you need.
\newblock \emph{arXiv preprint arXiv:2409.18869}, 2024.

\bibitem[Wang et~al.(2023)Wang, Wei, Schuurmans, Le, Chi, Narang, Chowdhery,
  and Zhou]{cot-sc}
Xuezhi Wang, Jason Wei, Dale Schuurmans, Quoc~V Le, Ed~H. Chi, Sharan Narang,
  Aakanksha Chowdhery, and Denny Zhou.
\newblock Self-consistency improves chain of thought reasoning in language
  models.
\newblock In \emph{International Conference on Learning Representations}, 2023.

\bibitem[Wang(1916)]{wangyangming}
Yangming Wang.
\newblock \emph{The philosophy of Wang Yang-ming}.
\newblock Open Court Publishing Company, 1916.

\bibitem[Wei et~al.(2022)Wei, Wang, Schuurmans, Bosma, Ichter, Xia, Chi, Le,
  and Zhou]{CoT}
Jason Wei, Xuezhi Wang, Dale Schuurmans, Maarten Bosma, Brian Ichter, Fei Xia,
  Ed~Chi, Quoc~V Le, and Denny Zhou.
\newblock Chain-of-thought prompting elicits reasoning in large language
  models.
\newblock In S.~Koyejo, S.~Mohamed, A.~Agarwal, D.~Belgrave, K.~Cho, and A.~Oh,
  editors, \emph{Advances in Neural Information Processing Systems}, volume~35,
  pages 24824--24837. Curran Associates, Inc., 2022.
\newblock \doi{10.52202/068431-1800}.

\bibitem[Wen et~al.(2026)Wen, Ye, Zhang, Yang, and Zhu]{wen2024realtime}
Yule Wen, Yixin Ye, Yanzhe Zhang, Diyi Yang, and Hao Zhu.
\newblock Real-time reasoning agents in evolving environments.
\newblock In C.~Vondrick, B.~Hariharan, C.~Raffel, L.~Pinto, D.~Yang, and
  A.~Faust, editors, \emph{International Conference on Learning
  Representations}, volume 2026, pages 126266--126292, 2026.

\bibitem[Wu et~al.(2025{\natexlab{a}})Wu, Li, Zhou, Lin, Gao, Yan, Yin, Bai,
  Xu, Chen, et~al.]{qwenimage}
Chenfei Wu, Jiahao Li, Jingren Zhou, Junyang Lin, Kaiyuan Gao, Kun Yan,
  {Sheng-ming} Yin, Shuai Bai, Xiao Xu, Yilei Chen, et~al.
\newblock {Qwen-Image} technical report.
\newblock \emph{arXiv preprint arXiv:2508.02324}, 2025{\natexlab{a}}.

\bibitem[Wu et~al.(2025{\natexlab{b}})Wu, Chen, Wu, Ma, Liu, Pan, Liu, Xie, Yu,
  Ruan, et~al.]{janus}
Chengyue Wu, Xiaokang Chen, Zhiyu Wu, Yiyang Ma, Xingchao Liu, Zizheng Pan, Wen
  Liu, Zhenda Xie, Xingkai Yu, Chong Ruan, et~al.
\newblock Janus: Decoupling visual encoding for unified multimodal
  understanding and generation.
\newblock In \emph{Proceedings of the IEEE/CVF Conference on Computer Vision
  and Pattern Recognition (CVPR)}, pages 12966--12977, June 2025{\natexlab{b}}.
\newblock \doi{10.1109/CVPR52734.2025.01210}.

\bibitem[Wu et~al.(2025{\natexlab{c}})Wu, Zhang, Chen, Tang, Li, Fang, Zhu,
  Xie, Yin, Yi, et~al.]{VILA-U}
Yecheng Wu, Zhuoyang Zhang, Junyu Chen, Haotian Tang, Dacheng Li, Yunhao Fang,
  Ligeng Zhu, Enze Xie, Hongxu Yin, Li~Yi, et~al.
\newblock {VILA-U}: a unified foundation model integrating visual understanding
  and generation.
\newblock In Y.~Yue, A.~Garg, N.~Peng, F.~Sha, and R.~Yu, editors,
  \emph{International Conference on Learning Representations}, volume 2025,
  pages 93620--93638, 2025{\natexlab{c}}.

\bibitem[Wu et~al.(2025{\natexlab{d}})Wu, Li, Hu, Ye, Zeng, Yu, Zhu, Schiele,
  Yang, and Yang]{KRIS-Bench}
Yongliang Wu, Zonghui Li, Xinting Hu, Xinyu Ye, Xianfang Zeng, Gang Yu, Wenbo
  Zhu, Bernt Schiele, Ming-Hsuan Yang, and Xu~Yang.
\newblock {KRIS-Bench}: Benchmarking next-level intelligent image editing
  models.
\newblock In D.~Belgrave, C.~Zhang, H.~Lin, R.~Pascanu, P.~Koniusz,
  M.~Ghassemi, and N.~Chen, editors, \emph{Advances in Neural Information
  Processing Systems}, volume 38, Main Conference. Curran Associates, Inc.,
  2025{\natexlab{d}}.
\newblock \doi{10.52202/085713-5242}.

\bibitem[Xiao et~al.(2025)Xiao, Wang, Zhou, Yuan, Xing, Yan, Li, Wang, Huang,
  and Liu]{OmniGen}
Shitao Xiao, Yueze Wang, Junjie Zhou, Huaying Yuan, Xingrun Xing, Ruiran Yan,
  Chaofan Li, Shuting Wang, Tiejun Huang, and Zheng Liu.
\newblock {OmniGen}: Unified image generation.
\newblock In \emph{Proceedings of the IEEE/CVF Conference on Computer Vision
  and Pattern Recognition (CVPR)}, pages 13294--13304, June 2025.
\newblock \doi{10.1109/CVPR52734.2025.01241}.

\bibitem[Xie et~al.(2025{\natexlab{a}})Xie, Mao, Bai, Zhang, Wang, Lin, Gu,
  Chen, Yang, and Shou]{show-o}
Jinheng Xie, Weijia Mao, Zechen Bai, David~Junhao Zhang, Weihao Wang,
  Kevin~Qinghong Lin, Yuchao Gu, Zhijie Chen, Zhenheng Yang, and Mike~Zheng
  Shou.
\newblock Show-o: One single transformer to unify multimodal understanding and
  generation.
\newblock In Y.~Yue, A.~Garg, N.~Peng, F.~Sha, and R.~Yu, editors,
  \emph{International Conference on Learning Representations}, volume 2025,
  pages 28240--28264, 2025{\natexlab{a}}.

\bibitem[Xie et~al.(2025{\natexlab{b}})Xie, Yang, and Shou]{Show-o2}
Jinheng Xie, Zhenheng Yang, and Mike~Zheng Shou.
\newblock Show-o2: Improved native unified multimodal models.
\newblock In D.~Belgrave, C.~Zhang, H.~Lin, R.~Pascanu, P.~Koniusz,
  M.~Ghassemi, and N.~Chen, editors, \emph{Advances in Neural Information
  Processing Systems}, volume 38, Main Conference, pages 47490--47518. Curran
  Associates, Inc., 2025{\natexlab{b}}.
\newblock \doi{10.52202/085713-1584}.

\bibitem[Xu et~al.(2025)Xu, Kong, Wang, Pan, Lin, and Liu]{InsightEdit}
Yingjing Xu, Jie Kong, Jiazhi Wang, Xiao Pan, Bo~Lin, and Qiang Liu.
\newblock {InsightEdit}: Towards better instruction following for image
  editing.
\newblock In \emph{Proceedings of the IEEE/CVF Conference on Computer Vision
  and Pattern Recognition (CVPR)}, pages 2694--2703, June 2025.
\newblock \doi{10.1109/CVPR52734.2025.00257}.

\bibitem[Xue et~al.(2025)Xue, Lu, Huang, Wang, Ouyang, and Bai]{comfybench}
Xiangyuan Xue, Zeyu Lu, Di~Huang, Zidong Wang, Wanli Ouyang, and Lei Bai.
\newblock {ComfyBench}: Benchmarking {LLM-based} agents in {ComfyUI} for
  autonomously designing collaborative {AI} systems.
\newblock In \emph{Proceedings of the IEEE/CVF Conference on Computer Vision
  and Pattern Recognition (CVPR)}, pages 24614--24624, June 2025.
\newblock \doi{10.1109/CVPR52734.2025.02292}.

\bibitem[Yu et~al.(2025)Yu, Chow, Yue, Pan, Wu, Wan, Li, Tang, Zhang, and
  Zhuang]{AnyEdit}
Qifan Yu, Wei Chow, Zhongqi Yue, Kaihang Pan, Yang Wu, Xiaoyang Wan, Juncheng
  Li, Siliang Tang, Hanwang Zhang, and Yueting Zhuang.
\newblock {AnyEdit}: Mastering unified high-quality image editing for any idea.
\newblock In \emph{Proceedings of the IEEE/CVF Conference on Computer Vision
  and Pattern Recognition (CVPR)}, pages 26125--26135, June 2025.
\newblock \doi{10.1109/CVPR52734.2025.02433}.

\bibitem[Zhang et~al.(2026{\natexlab{a}})Zhang, Wang, Chen, Zhou, Wang, and
  Yan]{AgenTracer}
Guibin Zhang, Junhao Wang, Junjie Chen, Wangchunshu Zhou, Kun Wang, and
  Shuicheng Yan.
\newblock {AgenTracer}: Who is inducing failure in the {LLM} agentic systems?
\newblock In C.~Vondrick, B.~Hariharan, C.~Raffel, L.~Pinto, D.~Yang, and
  A.~Faust, editors, \emph{International Conference on Learning
  Representations}, volume 2026, pages 11377--11399, 2026{\natexlab{a}}.

\bibitem[Zhang et~al.(2023)Zhang, Mo, Chen, Sun, and Su]{MagicBrush}
Kai Zhang, Lingbo Mo, Wenhu Chen, Huan Sun, and Yu~Su.
\newblock {MagicBrush}: A manually annotated dataset for instruction-guided
  image editing.
\newblock In A.~Oh, T.~Naumann, A.~Globerson, K.~Saenko, M.~Hardt, and
  S.~Levine, editors, \emph{Advances in Neural Information Processing Systems},
  volume~36, pages 31428--31449. Curran Associates, Inc., 2023.
\newblock \doi{10.52202/075280-1365}.

\bibitem[Zhang et~al.(2026{\natexlab{b}})Zhang, Li, Zhang, Fu, Song, Bian,
  Zhang, Yang, and Wang]{PixelCraft}
Shuoshuo Zhang, Zijian Li, Yizhen Zhang, Jingjing Fu, Lei Song, Jiang Bian, Jun
  Zhang, Yujiu Yang, and Rui Wang.
\newblock {PixelCraft}: A multi-agent system for high-fidelity visual reasoning
  on structured images.
\newblock In C.~Vondrick, B.~Hariharan, C.~Raffel, L.~Pinto, D.~Yang, and
  A.~Faust, editors, \emph{International Conference on Learning
  Representations}, volume 2026, pages 77608--77632, 2026{\natexlab{b}}.

\bibitem[Zhang et~al.(2026{\natexlab{c}})Zhang, Huang, and Wang]{IntentEdit}
Yuxuan Zhang, Shijia Huang, and Liwei Wang.
\newblock {IntentEdit}: Multi-agent reasoning for intent-driven complex image
  editing.
\newblock In \emph{Proceedings of the IEEE/CVF Conference on Computer Vision
  and Pattern Recognition (CVPR) Findings}, pages 8776--8785, June
  2026{\natexlab{c}}.
\newblock \doi{10.1109/CVPRF71909.2026.00842}.

\bibitem[Zhao et~al.(2026)Zhao, Zhang, Wang, Deng, Zhao, Zhi, Huang, He, Che,
  Qin, et~al.]{zhao-etal-2026-safety}
Weixiang Zhao, Yichen Zhang, Yingshuo Wang, Yang Deng, Yanyan Zhao, Xuda Zhi,
  Yongbo Huang, Hao He, Wanxiang Che, Bing Qin, et~al.
\newblock On safety risks in experience-driven self-evolving agents.
\newblock In Maria Liakata, Viviane~P. Moreira, Jiajun Zhang, and David
  Jurgens, editors, \emph{Findings of the {A}ssociation for {C}omputational
  {L}inguistics: {ACL} 2026}, pages 42145--42169, San Diego, California, United
  States, July 2026. Association for Computational Linguistics.
\newblock ISBN 979-8-89176-395-1.
\newblock \doi{10.18653/v1/2026.findings-acl.2091}.

\bibitem[Zhu et~al.(2026)Zhu, Sun, Zhu, Ma, Zhang, Wang, Yang, Zhang, Lan,
  Zhao, et~al.]{IEA}
Zichen Zhu, Yuheng Sun, Mingxuan Zhu, Wenjie Ma, Situo Zhang, Zhexiang Wang,
  Ziyue Yang, Danyang Zhang, Kunyao Lan, Zihan Zhao, et~al.
\newblock {IEA}: Amateur-friendly conversational image editing agent via three
  stages of multitask alignment.
\newblock In \emph{Proceedings of the IEEE/CVF Conference on Computer Vision
  and Pattern Recognition (CVPR) Findings}, pages 8672--8683, June 2026.
\newblock \doi{10.1109/CVPRF71909.2026.00832}.

\end{thebibliography}

\clearpage
\beginappendix
\section{Appendix}
\label{sec:appendix}

\begingroup

\rmfamily
\fontsize{10pt}{14pt}\selectfont
\setcounter{tocdepth}{3}

\renewcommand{\cftsubsecfont}{\rmfamily}
\renewcommand{\cftsubsecpagefont}{\rmfamily}
\renewcommand{\cftsubsubsecfont}{\rmfamily}
\renewcommand{\cftsubsubsecpagefont}{\rmfamily}

\setlength{\cftsubsecnumwidth}{3em}
\setlength{\cftsubsubsecnumwidth}{3.4em}

\setlength{\cftbeforesubsecskip}{4pt}
\setlength{\cftbeforesubsubsecskip}{0pt}

\etocsettocstyle
{%
    \subsection*{Appendix Contents}
    \vspace{6pt}
}
{}

\localtableofcontents

\endgroup

\subsection{Additional Experiments}
\label{app:additional_experiments}

\subsubsection{Policy Evolution and Execution Analysis}
\label{app:policy_evolution_execution}

\begin{figure}[!htbp]
\centering
\includegraphics[width=\textwidth]{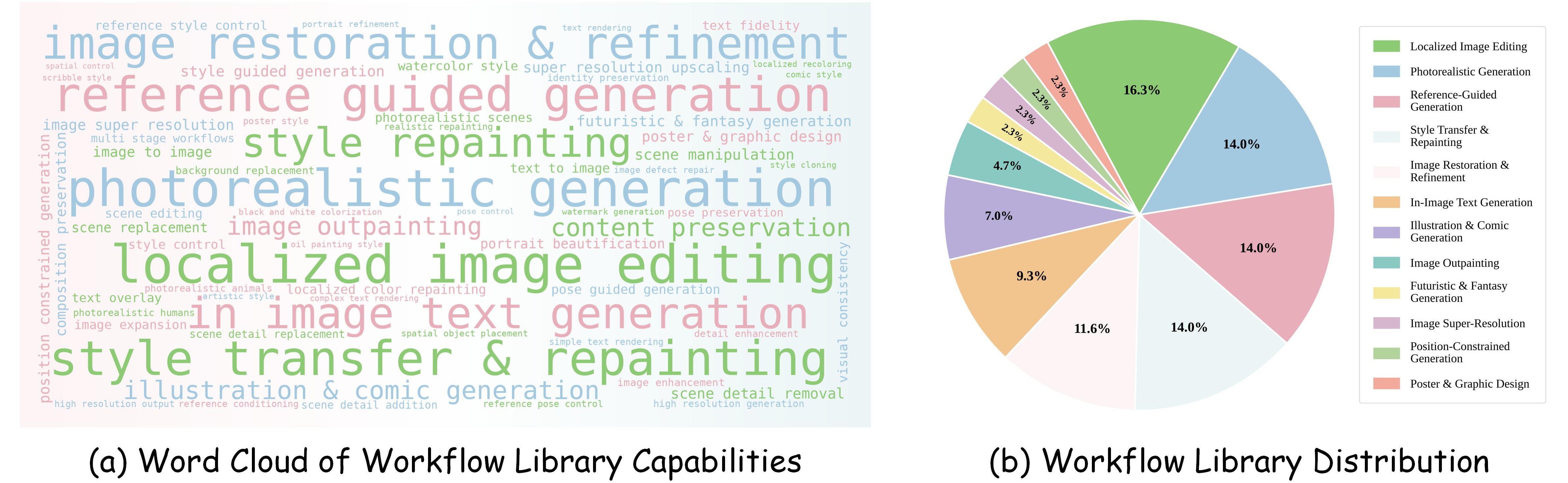}
\caption{
Visualization of the workflow library after 50 iterations of self-directed inquiry.
(a) Word cloud summarizing the capabilities represented by stored policies.
(b) Distribution of workflow entries across capability categories.
}
\label{fig:4}
\end{figure}

After 50 rounds of self-directed inquiry, OmniHarness acquires 43 symbolic policies spanning all twelve predefined capability categories, as shown in Figure~\ref{fig:4}. Localized Image Editing accounts for the largest share at 16.3\%. Photorealistic Generation, Reference-Guided Generation, and Style Transfer \& Repainting each account for 14.0\%.

\begin{figure}[!htbp]
\centering
\includegraphics[width=\textwidth]{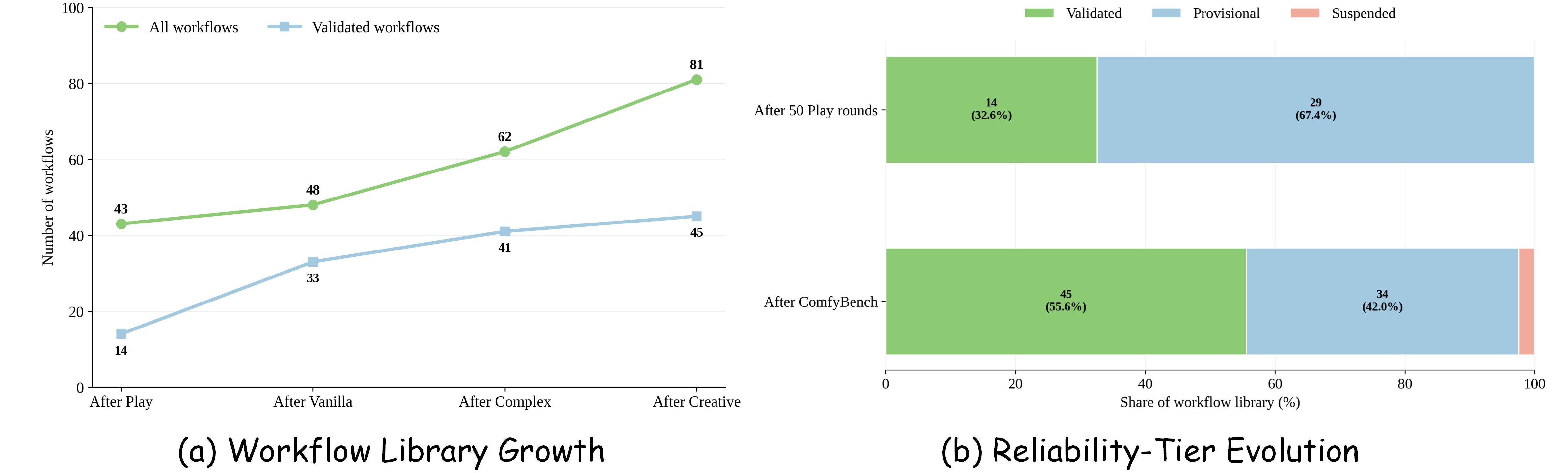}
\caption{
Symbolic policy learning during downstream execution on ComfyBench~\citep{comfybench}.
(a) Growth in total and validated workflow entries in the workflow library, starting after 50 inquiry iterations and continuing through the Vanilla, Complex, and Creative subsets.
(b) Changes in the proportions of \textit{Validated}, \textit{Provisional}, and \textit{Suspended} workflows before and after ComfyBench execution.
}
\label{fig:5}
\end{figure}

During downstream execution, the library expands while accumulating evidence about policy reliability. Figure~\ref{fig:5} shows growth from 43 to 48, 62, and 81 entries after the Vanilla, Complex, and Creative subsets, respectively. Vanilla adds five entries but increases the number of \emph{Validated} policies from 14 to 33, strengthening evidence for existing policies. Complex and Creative introduce more policies, with validation progressing more slowly than library growth. By the end of evaluation, 45 policies are \emph{Validated} and two are \emph{Suspended}, allowing subsequent retrieval to distinguish their reliability.

\begin{figure}[!htbp]
\centering
\includegraphics[width=\textwidth]{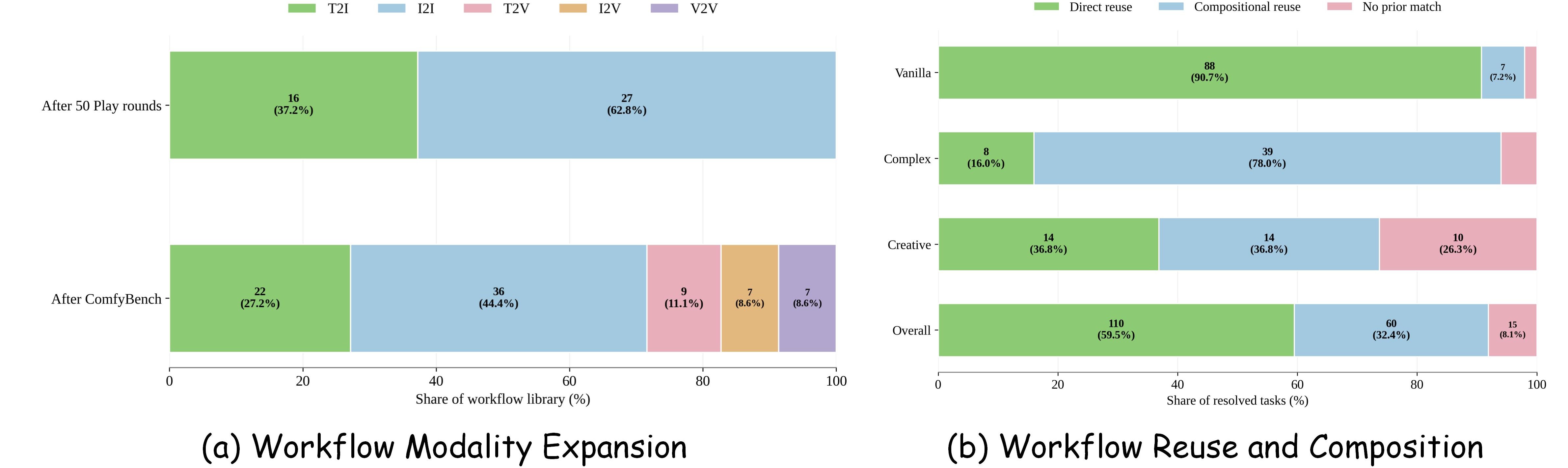}
\caption{
Evolution and reuse of symbolic policies during downstream execution on ComfyBench~\citep{comfybench}.
(a) Expansion of the image-only workflow library acquired after 50 inquiry iterations to include T2I, I2I, T2V, I2V, and V2V workflows after ComfyBench execution.
(b) Distribution of direct reuse, compositional reuse, and no-prior-match cases among resolved Vanilla, Complex, and Creative tasks, together with the overall distribution.
}
\label{fig:6}
\end{figure}

Figure~\ref{fig:6} shows that downstream execution also extends the initially image-only library with 23 video workflows. Among 185 resolved tasks, 110 use direct reuse and 60 use compositional reuse, yielding a combined reuse rate of 91.9\%. Direct reuse dominates Vanilla tasks, while composition accounts for 78.0\% of resolved Complex tasks. Creative tasks have the highest proportion without a prior policy match at 26.3\%, indicating a greater need to construct new workflows.

\begin{figure}[!htbp]
\centering
\includegraphics[width=\textwidth]{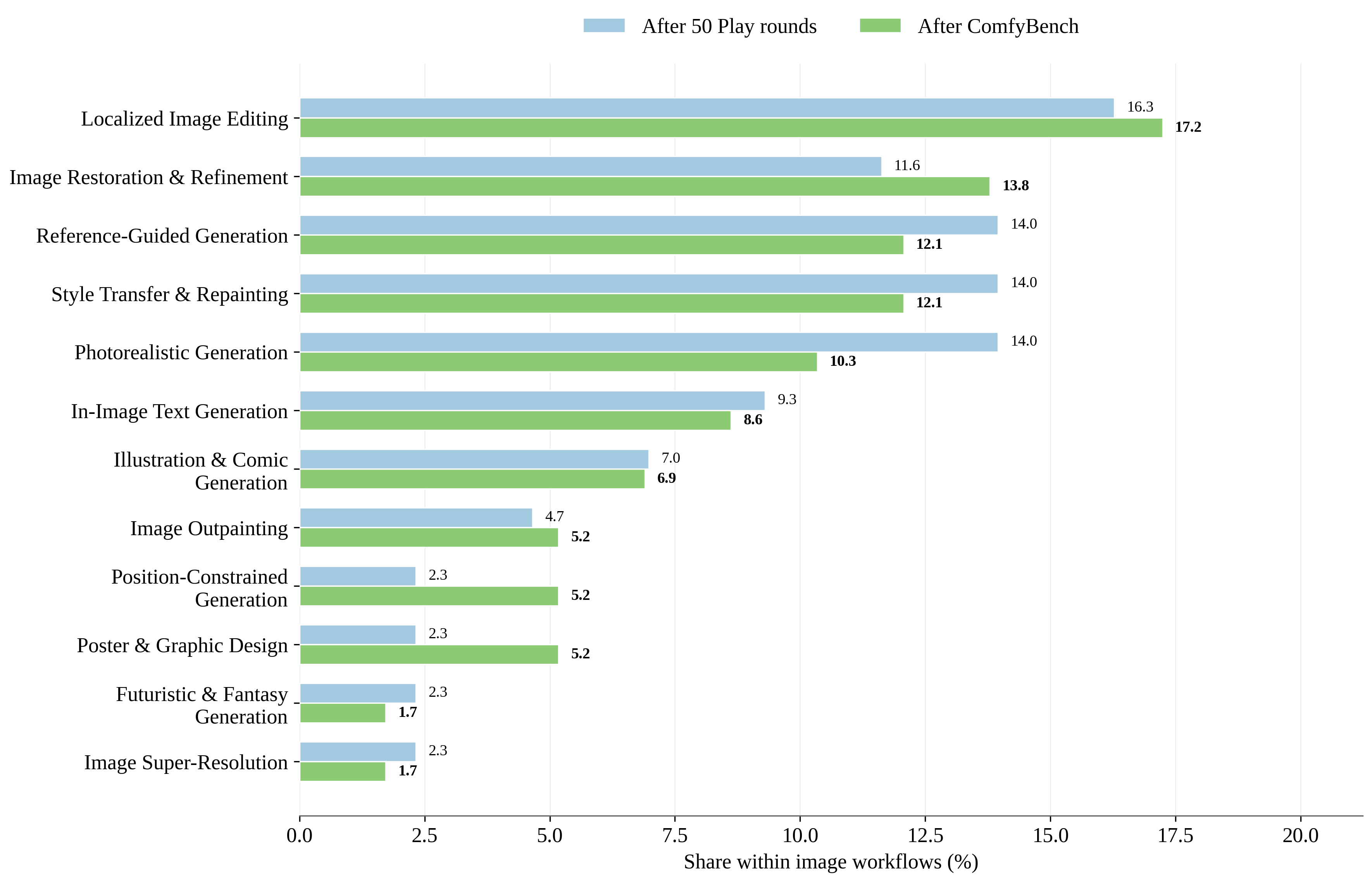}
\caption{
Capability distribution shift in the image workflow library during downstream execution on ComfyBench~\citep{comfybench}.
The bars compare workflow proportions across capability categories after 50 inquiry iterations and after ComfyBench execution.
Percentages are normalized over T2I and I2I workflows only.
The comparison shows how the capabilities represented by stored symbolic policies change during downstream execution.
}
\label{fig:7}
\end{figure}

Within the image workflow library, Figure~\ref{fig:7} reveals changes in capability distribution as the number of entries grows from 43 to 58. Localized Image Editing increases from seven to ten policies, while Image Restoration \& Refinement increases from five to eight. Spatially constrained generation and graphic design also gain coverage. Relative shares reflect growth across the entire image library. For example, Reference-Guided Generation expands from six to seven policies even though its share decreases from 14.0\% to 12.1\%.

\begin{figure}[!htbp]
\centering
\includegraphics[width=\textwidth]{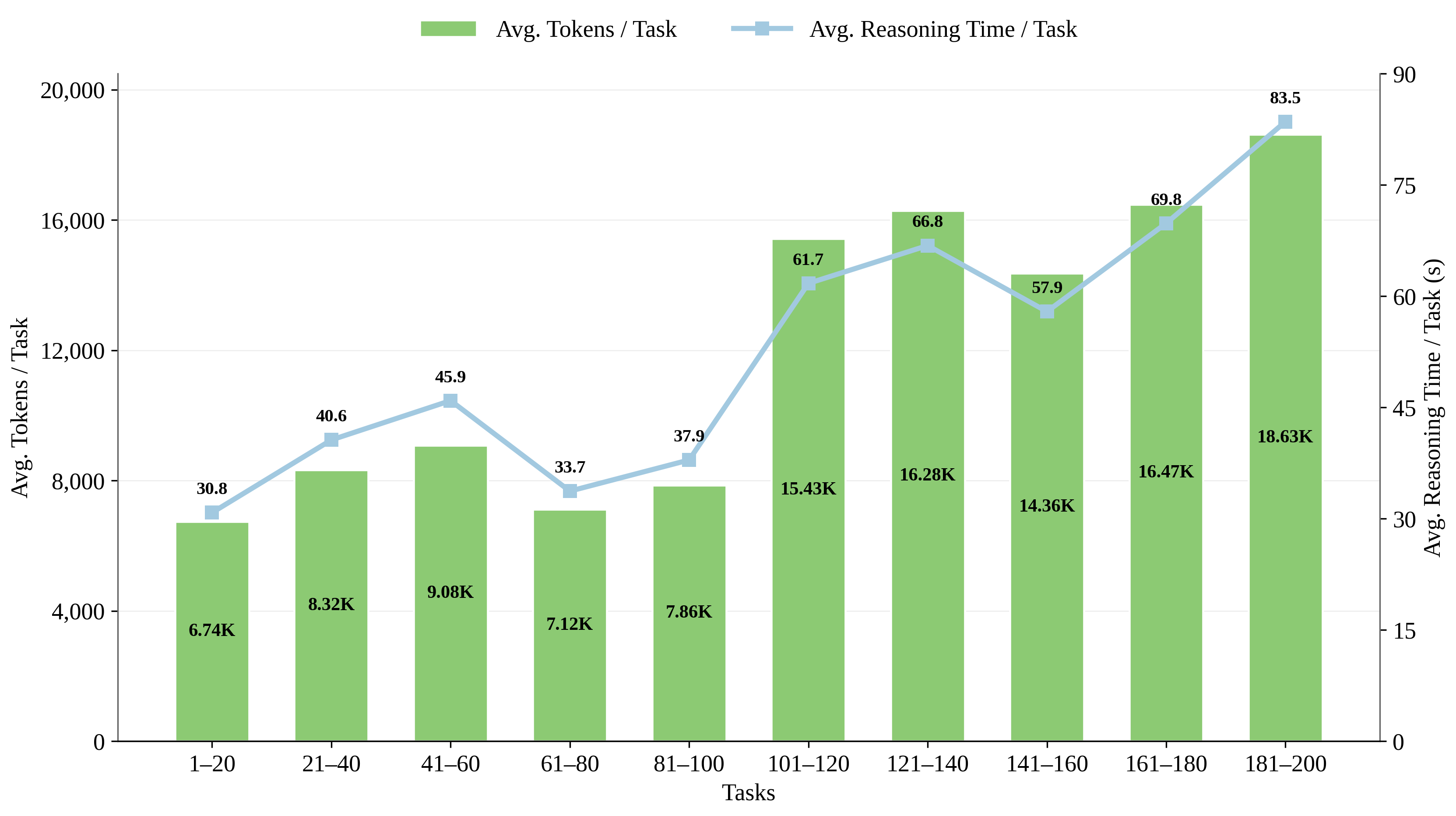}
\caption{
Reasoning efficiency of OmniHarness on the 200 ComfyBench tasks, reported over consecutive groups of 20 tasks.
The bars show the average token consumption per task, while the line shows the average reasoning time per task.
Tasks 1--100, 101--160, and 161--200 correspond to the Vanilla, Complex, and Creative subsets, respectively.
Reasoning time includes agent-side planning, workflow synthesis, verification, and recovery, but excludes visual generation model execution.
}
\label{fig:8}
\end{figure}

Figure~\ref{fig:8} reports reasoning costs across successive task groups. OmniHarness averages 12.03K tokens and 52.9 seconds per task over the 200 ComfyBench tasks. Costs are lower on Vanilla and generally higher on Complex and Creative, consistent with greater demands for workflow composition, verification, and recovery. The final task group reaches 18.63K tokens and 83.5 seconds per task. These measurements cover agent-side planning, workflow synthesis, verification, and recovery, excluding visual generation model execution.

\subsubsection{Backbone Ablation}

Table~\ref{tab:backbone_ablation} evaluates OmniHarness across reasoning models and planner configurations. All five configurations achieve a 100.0\% Pass rate. With GPT-4o, OmniHarness exceeds ComfyMind by 6.5 percentage points in Total Resolve. With Gemini-2.5-Flash, it exceeds SymbOmni by 4.0 points. Both comparisons show larger gains on Creative tasks, although Complex performance remains below the respective baselines. Across OmniHarness configurations, Total Resolve ranges from 89.5\% to 95.0\%, with Codex GPT-5.5 achieving the highest score. The improvements under matched reasoning backbones demonstrate the contribution of the OmniHarness framework. Its performance across multiple models further supports effectiveness beyond a single backbone.

\begin{table}[!htbp]
\caption{
Backbone ablation of OmniHarness on ComfyBench~\citep{comfybench}.
The first three rows report representative existing agents, followed by
OmniHarness instantiated with different reasoning backbones and planner
configurations. Pass and Resolve are reported as percentages, with higher
values indicating better performance.
}
\label{tab:backbone_ablation}
\centering

\begingroup
\fontsize{8pt}{10pt}\selectfont
\setlength{\tabcolsep}{1.0pt}
\renewcommand{\arraystretch}{0.90}
\setlength{\extrarowheight}{0.2pt}
\setlength{\aboverulesep}{0.6pt}
\setlength{\belowrulesep}{0.6pt}
\setlength{\cmidrulesep}{0.5pt}
\renewcommand{\tabularxcolumn}[1]{m{#1}}

\newcommand{\metrichead}[1]{%
    {\fontsize{6pt}{7.4pt}\selectfont\bfseries\mbox{#1}}%
}

\begin{tabularx}{\linewidth}{
    l
    *{8}{>{\centering\arraybackslash}X}
}
\toprule

\multicolumn{1}{c}{
    \multirow[c]{2}{*}{\paperbold{Agent}}
}
&
\multicolumn{2}{c}{\paperbold{Vanilla}}
&
\multicolumn{2}{c}{\paperbold{Complex}}
&
\multicolumn{2}{c}{\paperbold{Creative}}
&
\multicolumn{2}{c}{\paperbold{Total}}
\\

\cmidrule(lr){2-3}
\cmidrule(lr){4-5}
\cmidrule(lr){6-7}
\cmidrule(lr){8-9}

&
\metrichead{Pass\(\uparrow\)}
&
\metrichead{Res.\(\uparrow\)}
&
\metrichead{Pass\(\uparrow\)}
&
\metrichead{Res.\(\uparrow\)}
&
\metrichead{Pass\(\uparrow\)}
&
\metrichead{Res.\(\uparrow\)}
&
\metrichead{Pass\(\uparrow\)}
&
\metrichead{Res.\(\uparrow\)}
\\
\midrule

GPT-4o + ComfyAgent~\citep{comfybench}
& 67.0
& 46.0
& 48.3
& 21.7
& 40.0
& 15.0
& 56.0
& 32.5
\\

GPT-4o + ComfyMind~\citep{comfymind}
& \bestscore{100.0}
& 92.0
& \bestscore{100.0}
& \secondscore{85.0}
& \bestscore{100.0}
& 57.5
& \bestscore{100.0}
& 83.0
\\

Gemini-2.5-Flash + SymbOmni~\citep{symbomni}
& \bestscore{100.0}
& 95.0
& \bestscore{100.0}
& 83.3
& \bestscore{100.0}
& 67.5
& \bestscore{100.0}
& 86.0
\\

\midrule

\rowcolor{omniharnessbg!10!white}
Gemini-2.5-Flash + OmniHarness
& \bestscore{100.0}
& 96.0
& \bestscore{100.0}
& 81.7
& \bestscore{100.0}
& 87.5
& \bestscore{100.0}
& 90.0
\\

\rowcolor{omniharnessbg!22!white}
GPT-4o + OmniHarness
& \bestscore{100.0}
& 95.0
& \bestscore{100.0}
& 76.7
& \bestscore{100.0}
& \secondscore{95.0}
& \bestscore{100.0}
& 89.5
\\

\rowcolor{omniharnessbg!45!white}
GPT-5.5 + OmniHarness
& \bestscore{100.0}
& \secondscore{97.0}
& \bestscore{100.0}
& 81.7
& \bestscore{100.0}
& \bestscore{97.5}
& \bestscore{100.0}
& \secondscore{92.5}
\\

\rowcolor{omniharnessbg!70!white}
Codex GPT-4o + OmniHarness
& \bestscore{100.0}
& \secondscore{97.0}
& \bestscore{100.0}
& 83.3
& \bestscore{100.0}
& \secondscore{95.0}
& \bestscore{100.0}
& \secondscore{92.5}
\\

\rowcolor{omniharnessbg}
\paperbold{Codex GPT-5.5 + OmniHarness}
& \bestscore{100.0}
& \bestscore{98.0}
& \bestscore{100.0}
& \bestscore{88.3}
& \bestscore{100.0}
& \bestscore{97.5}
& \bestscore{100.0}
& \bestscore{95.0}
\\

\bottomrule
\end{tabularx}
\endgroup
\end{table}

\subsubsection{Source Image Pool Isolation Study}

To assess whether the benefits of inquiry depend on ComfyBench source images, we replace the original pool with 33 independently generated images from the same generative capability space.
As shown in Table~\ref{tab:source_pool_independence}, OmniHarness with \(\mathcal{X}_{\mathrm{Gen}}\) achieves a Total Resolve rate of 91.5\%, compared with 92.5\% using \(\mathcal{X}_{\mathrm{ComfyBench}}\).
Vanilla and Complex results remain unchanged, while Creative Resolve decreases.
These results suggest that inquiry can acquire useful symbolic policies from independently generated images, with limited dependence on the original ComfyBench source pool in this evaluation.

\begin{table}[!htbp]
\caption{
Source image pool isolation study on ComfyBench~\citep{comfybench}.
\(\mathcal{X}_{\mathrm{ComfyBench}}\) denotes the original ComfyBench source image pool,
while \(\mathcal{X}_{\mathrm{Gen}}\) replaces it with independently generated images
from the same generative capability space.
}
\label{tab:source_pool_independence}
\centering

\begingroup
\fontsize{8pt}{10pt}\selectfont
\setlength{\tabcolsep}{1.0pt}
\renewcommand{\arraystretch}{0.90}
\setlength{\extrarowheight}{0.2pt}
\setlength{\aboverulesep}{0.6pt}
\setlength{\belowrulesep}{0.6pt}
\setlength{\cmidrulesep}{0.5pt}
\renewcommand{\tabularxcolumn}[1]{m{#1}}

\newcommand{\metrichead}[1]{%
    {\fontsize{6pt}{7.4pt}\selectfont\bfseries\mbox{#1}}%
}

\begin{tabularx}{\linewidth}{
    l
    *{8}{>{\centering\arraybackslash}X}
}
\toprule

\multicolumn{1}{c}{
    \multirow[c]{2}{*}{\paperbold{Agent}}
}
& \multicolumn{2}{c}{\paperbold{Vanilla}}
& \multicolumn{2}{c}{\paperbold{Complex}}
& \multicolumn{2}{c}{\paperbold{Creative}}
& \multicolumn{2}{c}{\paperbold{Total}}
\\

\cmidrule(lr){2-3}
\cmidrule(lr){4-5}
\cmidrule(lr){6-7}
\cmidrule(lr){8-9}

& \metrichead{Pass\(\uparrow\)}
& \metrichead{Res.\(\uparrow\)}
& \metrichead{Pass\(\uparrow\)}
& \metrichead{Res.\(\uparrow\)}
& \metrichead{Pass\(\uparrow\)}
& \metrichead{Res.\(\uparrow\)}
& \metrichead{Pass\(\uparrow\)}
& \metrichead{Res.\(\uparrow\)}
\\
\midrule

\rowcolor{omniharnessbg}
\paperbold{OmniHarness with \(\mathcal{X}_{\mathrm{ComfyBench}}\)}
& \bestscore{100.0}
& \bestscore{97.0}
& \bestscore{100.0}
& \bestscore{83.3}
& \bestscore{100.0}
& \bestscore{95.0}
& \bestscore{100.0}
& \bestscore{92.5}
\\

\rowcolor{symbomnibg}
\paperbold{OmniHarness with \(\mathcal{X}_{\mathrm{Gen}}\)}
& \bestscore{100.0}
& \bestscore{97.0}
& \bestscore{100.0}
& \bestscore{83.3}
& \bestscore{100.0}
& \secondscore{90.0}
& \bestscore{100.0}
& \secondscore{91.5}
\\

\bottomrule
\end{tabularx}
\endgroup
\end{table}

\subsubsection{Cross-Modal Generalization Evaluation}

\begin{figure}[!htbp]
\centering
\includegraphics[width=0.8\textwidth]{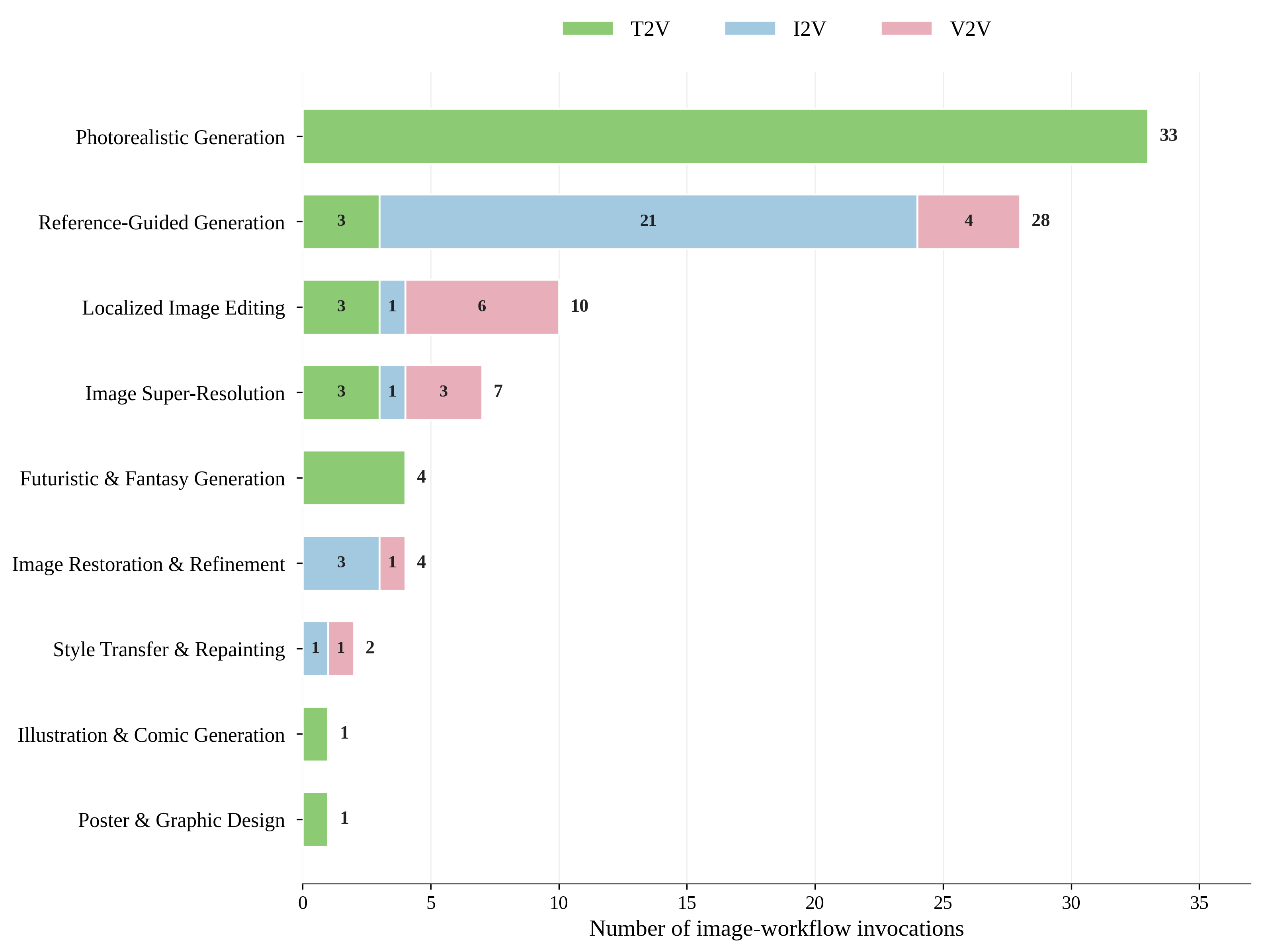}
\caption{
Reuse of symbolic policies learned through image-only inquiry during
cross-modal generalization on ComfyBench~\citep{comfybench}.
}
\label{fig:cross_modal_image_workflow_reuse_revised}
\end{figure}

\begin{figure}[!htbp]
\centering
\includegraphics[width=0.9\textwidth]{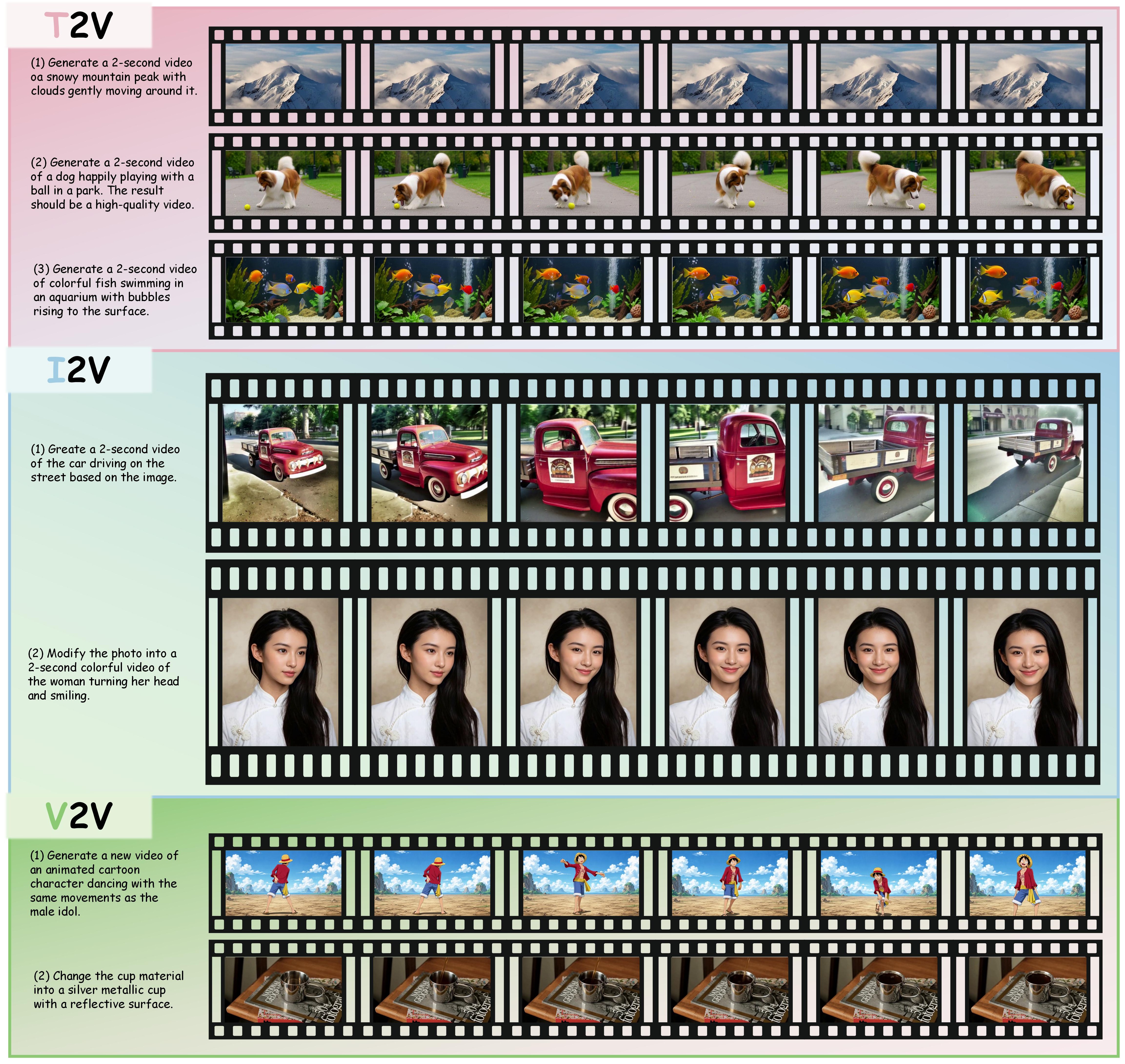}
\caption{
Qualitative cross-modal generalization results of OmniHarness on
representative video-generation tasks from ComfyBench~\citep{comfybench}.
}
\label{fig:2v}
\end{figure}

We evaluate OmniHarness on ComfyBench video tasks using the frozen policy snapshot $\mathcal{K}_{\mathrm{inquiry}}$, learned through image-only inquiry without exposure to video tasks. Table~\ref{tab:cross_modal_generalization} shows a 100.0\% Pass rate across T2V, I2V, and V2V, with an overall Resolve rate of 85.9\%, exceeding SymbOmni by 5.1 percentage points. The largest gain occurs on V2V, where Resolve reaches 80.0\%, an improvement of 13.3 points. Figure~\ref{fig:2v} illustrates content generation and source-content preservation across these modalities. Figure~\ref{fig:cross_modal_image_workflow_reuse_revised} records 90 invocations of image policies, including 33 for Photorealistic Generation and 28 for Reference-Guided Generation. These policies provide content generation, reference conditioning, and editing operations, while video-specific components handle temporal processing. Together, the results support cross-modal generalization through adaptation and composition of image policies within video workflows.

\begin{table}[!htbp]
\caption{
Quantitative cross-modal generalization results on the video-generation
subset of ComfyBench~\citep{comfybench}.
T2V, I2V, and V2V denote text-to-video, image-to-video, and video-to-video
tasks, respectively.
}
\label{tab:cross_modal_generalization}
\centering

\begingroup
\fontsize{8pt}{10pt}\selectfont
\setlength{\tabcolsep}{1.0pt}
\renewcommand{\arraystretch}{0.90}
\setlength{\extrarowheight}{0.2pt}
\setlength{\aboverulesep}{0.6pt}
\setlength{\belowrulesep}{0.6pt}
\setlength{\cmidrulesep}{0.5pt}
\renewcommand{\tabularxcolumn}[1]{m{#1}}

\newcommand{\metrichead}[1]{%
    {\fontsize{6pt}{7.4pt}\selectfont\bfseries\mbox{#1}}%
}

\begin{tabularx}{\linewidth}{
    l
    *{8}{>{\centering\arraybackslash}X}
}
\toprule

\multicolumn{1}{c}{
    \multirow[c]{2}{*}{\paperbold{Agent}}
}
&
\multicolumn{2}{c}{\paperbold{T2V}}
&
\multicolumn{2}{c}{\paperbold{I2V}}
&
\multicolumn{2}{c}{\paperbold{V2V}}
&
\multicolumn{2}{c}{\paperbold{Total}}
\\

\cmidrule(lr){2-3}
\cmidrule(lr){4-5}
\cmidrule(lr){6-7}
\cmidrule(lr){8-9}

&
\metrichead{Pass\(\uparrow\)}
&
\metrichead{Res.\(\uparrow\)}
&
\metrichead{Pass\(\uparrow\)}
&
\metrichead{Res.\(\uparrow\)}
&
\metrichead{Pass\(\uparrow\)}
&
\metrichead{Res.\(\uparrow\)}
&
\metrichead{Pass\(\uparrow\)}
&
\metrichead{Res.\(\uparrow\)}
\\
\midrule

\rowcolor{comfyagentbg}
GPT-4o + ComfyAgent~\citep{comfybench}
& \secondscore{68.4}
& 47.4
& \secondscore{48.0}
& 20.0
& \secondscore{40.0}
& 13.3
& \secondscore{56.4}
& 32.1
\\

\rowcolor{comfymindbg}
GPT-4o + ComfyMind~\citep{comfymind}
& \bestscore{100.0}
& 71.1
& \bestscore{100.0}
& 76.0
& \bestscore{100.0}
& 53.3
& \bestscore{100.0}
& 69.2
\\

\rowcolor{symbomnibg}
Gemini-2.5-Flash + SymbOmni~\citep{symbomni}
& \bestscore{100.0}
& \secondscore{81.6}
& \bestscore{100.0}
& \secondscore{88.0}
& \bestscore{100.0}
& \secondscore{66.7}
& \bestscore{100.0}
& \secondscore{80.8}
\\

\rowcolor{omniharnessbg}
\paperbold{GPT-4o + OmniHarness}
& \bestscore{100.0}
& \bestscore{84.2}
& \bestscore{100.0}
& \bestscore{92.0}
& \bestscore{100.0}
& \bestscore{80.0}
& \bestscore{100.0}
& \bestscore{85.9}
\\

\bottomrule
\end{tabularx}
\endgroup
\end{table}

\subsubsection{User Study}

\begin{figure}[!htbp]
    \centering
    \includegraphics[width=\textwidth]{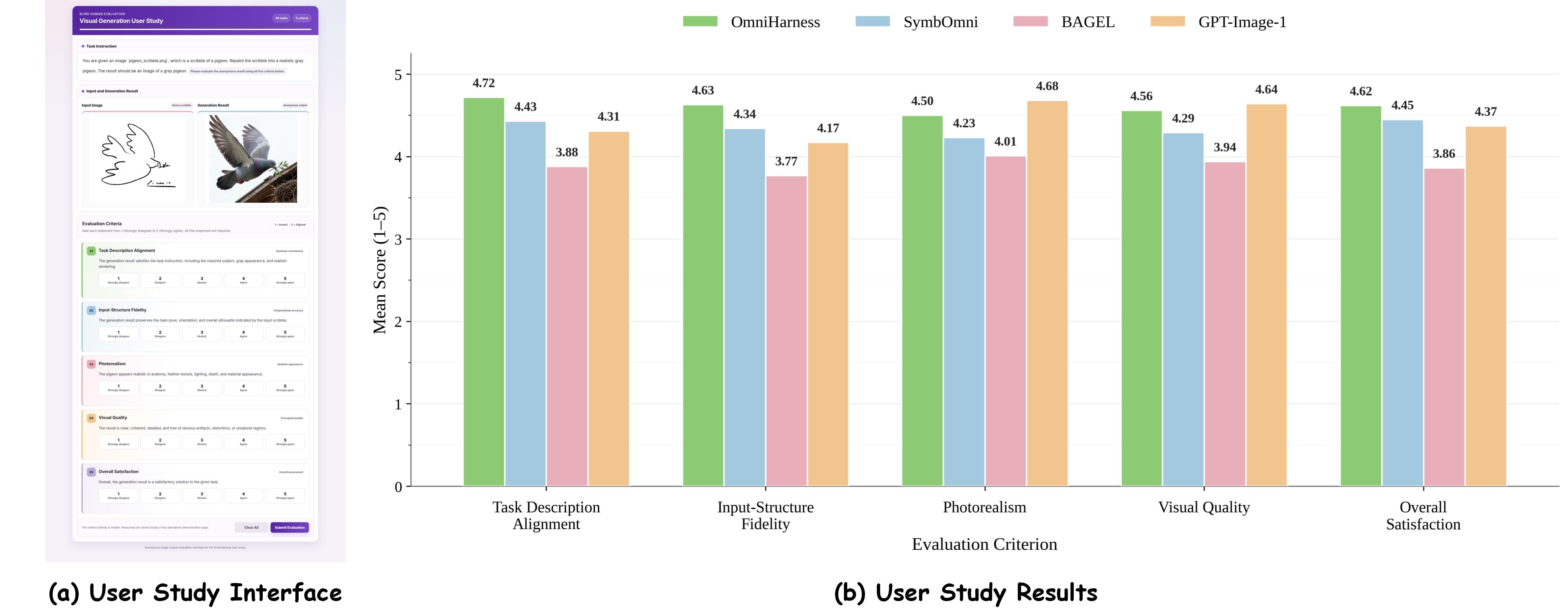}
    \caption{
        User study on 40 ComfyBench tasks.
        (a) Evaluation interface.
        (b) Mean ratings from 36 participants on a five-point Likert scale.
    }
    \label{fig:user_study}
\end{figure}

We conduct a blind user study with 36 participants on 40 ComfyBench tasks~\citep{comfybench}. Participants rate anonymized outputs from OmniHarness, SymbOmni, BAGEL, and GPT-Image-1 on a five-point Likert scale. Figure~\ref{fig:user_study} presents the interface and mean ratings across five criteria. OmniHarness achieves the highest ratings for Task Description Alignment, Input-Structure Fidelity, and Overall Satisfaction at 4.72, 4.63, and 4.62, respectively. It also obtains the highest average across all five criteria at 4.61, while GPT-Image-1 leads in Photorealism and Visual Quality. These results highlight the strengths of OmniHarness in satisfying task requirements and preserving input structure, alongside favorable overall user satisfaction.

\subsubsection{Modality Breakdown of Ablation Results}
\label{app:modality_ablation_results}

Tables~\ref{tab:play_fine_grained_ablation}
and~\ref{tab:execution_fine_grained_ablation}
report T2I and I2I results for the Creative and Complex ablations.
Table~\ref{tab:full_modality_ablations} provides the complete
breakdown across all five modalities.
Creative contains 40 tasks, comprising 6 T2I, 15 I2I, 5 T2V,
3 I2V, and 11 V2V tasks.
Complex contains 60 tasks, comprising 10 T2I, 23 I2I, 13 T2V,
12 I2V, and 2 V2V tasks.
Within each modality, Pass and Resolve use the full task count
as the denominator, including execution failures.
Total rates are computed by summing the corresponding passing
or resolved task counts across all five modalities and dividing
by 40 or 60.
The full configuration resolves 38 of 40 Creative tasks and
50 of 60 Complex tasks, yielding 95.00\% and 83.33\%, respectively.

\begin{table}[!htbp]
\caption{
Complete modality breakdown of component ablations on ComfyBench.
$n$ denotes the number of tasks.
Pass is reported as a percentage.
Each Resolve entry gives the resolved task count followed by
its percentage in parentheses.
Total aggregates all five modalities.
}
\label{tab:full_modality_ablations}
\centering

\begingroup
\fontsize{8pt}{10pt}\selectfont
\setlength{\tabcolsep}{1.0pt}
\renewcommand{\arraystretch}{0.90}
\setlength{\extrarowheight}{0.2pt}
\setlength{\aboverulesep}{0.6pt}
\setlength{\belowrulesep}{0.6pt}
\setlength{\cmidrulesep}{0.5pt}
\renewcommand{\tabularxcolumn}[1]{m{#1}}

\definecolor{maRowOne}{HTML}{EFF4F8}
\definecolor{maRowTwo}{HTML}{E7EFF5}
\definecolor{maRowThree}{HTML}{D8E6F1}
\definecolor{maRowFull}{HTML}{C9DDEC}

\newcommand{\maHead}[1]{%
    {\fontsize{6pt}{7.4pt}\selectfont\bfseries\mbox{#1}}%
}

\begin{adjustbox}{max width=\linewidth}
\begin{tabular}{@{}>{\raggedright\arraybackslash}m{62pt}*{12}{c}@{}}
\toprule

\multicolumn{13}{l}{
    \textit{(a) Task selection ablations on the Creative subset}
}
\\

\multicolumn{1}{c}{
    \multirow[c]{2}{*}{\paperbold{Method}}
}
& \multicolumn{2}{c}{\paperbold{T2I} ($n=6$)}
& \multicolumn{2}{c}{\paperbold{I2I} ($n=15$)}
& \multicolumn{2}{c}{\paperbold{T2V} ($n=5$)}
& \multicolumn{2}{c}{\paperbold{I2V} ($n=3$)}
& \multicolumn{2}{c}{\paperbold{V2V} ($n=11$)}
& \multicolumn{2}{c}{\paperbold{Total} ($n=40$)}
\\

\cmidrule(lr){2-3}
\cmidrule(lr){4-5}
\cmidrule(lr){6-7}
\cmidrule(lr){8-9}
\cmidrule(lr){10-11}
\cmidrule(lr){12-13}

& \maHead{Pass\(\uparrow\)} & \maHead{Res.\(\uparrow\)}
& \maHead{Pass\(\uparrow\)} & \maHead{Res.\(\uparrow\)}
& \maHead{Pass\(\uparrow\)} & \maHead{Res.\(\uparrow\)}
& \maHead{Pass\(\uparrow\)} & \maHead{Res.\(\uparrow\)}
& \maHead{Pass\(\uparrow\)} & \maHead{Res.\(\uparrow\)}
& \maHead{Pass\(\uparrow\)} & \maHead{Res.\(\uparrow\)}
\\
\midrule

\rowcolor{maRowOne}
w/o Capability Novelty
& \mbox{100.00} & \mbox{4 (66.67)}
& \mbox{100.00} & \mbox{13 (86.67)}
& \mbox{100.00} & \mbox{4 (80.00)}
& \mbox{100.00} & \mbox{3 (100.00)}
& \mbox{100.00} & \mbox{11 (100.00)}
& \mbox{100.00} & \mbox{35 (87.50)}
\\

\rowcolor{maRowThree}
w/o Competence Frontier
& \mbox{100.00} & \mbox{5 (83.33)}
& \mbox{100.00} & \mbox{12 (80.00)}
& \mbox{100.00} & \mbox{5 (100.00)}
& \mbox{100.00} & \mbox{2 (66.67)}
& \mbox{100.00} & \mbox{10 (90.91)}
& \mbox{100.00} & \mbox{34 (85.00)}
\\

\rowcolor{maRowFull}
\paperbold{OmniHarness}
& \mbox{\paperbold{100.00}} & \mbox{\paperbold{5 (83.33)}}
& \mbox{\paperbold{100.00}} & \mbox{\paperbold{14 (93.33)}}
& \mbox{\paperbold{100.00}} & \mbox{\paperbold{5 (100.00)}}
& \mbox{\paperbold{100.00}} & \mbox{\paperbold{3 (100.00)}}
& \mbox{\paperbold{100.00}} & \mbox{\paperbold{11 (100.00)}}
& \mbox{\paperbold{100.00}} & \mbox{\paperbold{38 (95.00)}}
\\
\midrule

\multicolumn{13}{l}{
    \textit{(b) Feedback-guided execution ablations on the Complex subset}
}
\\

\multicolumn{1}{c}{
    \multirow[c]{2}{*}{\paperbold{Method}}
}
& \multicolumn{2}{c}{\paperbold{T2I} ($n=10$)}
& \multicolumn{2}{c}{\paperbold{I2I} ($n=23$)}
& \multicolumn{2}{c}{\paperbold{T2V} ($n=13$)}
& \multicolumn{2}{c}{\paperbold{I2V} ($n=12$)}
& \multicolumn{2}{c}{\paperbold{V2V} ($n=2$)}
& \multicolumn{2}{c}{\paperbold{Total} ($n=60$)}
\\

\cmidrule(lr){2-3}
\cmidrule(lr){4-5}
\cmidrule(lr){6-7}
\cmidrule(lr){8-9}
\cmidrule(lr){10-11}
\cmidrule(lr){12-13}

& \maHead{Pass\(\uparrow\)} & \maHead{Res.\(\uparrow\)}
& \maHead{Pass\(\uparrow\)} & \maHead{Res.\(\uparrow\)}
& \maHead{Pass\(\uparrow\)} & \maHead{Res.\(\uparrow\)}
& \maHead{Pass\(\uparrow\)} & \maHead{Res.\(\uparrow\)}
& \maHead{Pass\(\uparrow\)} & \maHead{Res.\(\uparrow\)}
& \maHead{Pass\(\uparrow\)} & \maHead{Res.\(\uparrow\)}
\\
\midrule

\rowcolor{maRowOne}
w/o Planning
& \mbox{80.00}  & \mbox{4 (40.00)}
& \mbox{86.96}  & \mbox{13 (56.52)}
& \mbox{84.62}  & \mbox{7 (53.85)}
& \mbox{91.67}  & \mbox{8 (66.67)}
& \mbox{100.00} & \mbox{1 (50.00)}
& \mbox{86.67}  & \mbox{33 (55.00)}
\\

\rowcolor{maRowTwo}
w/o Intermediate Verification
& \mbox{70.00}  & \mbox{6 (60.00)}
& \mbox{73.91}  & \mbox{16 (69.57)}
& \mbox{76.92}  & \mbox{8 (61.54)}
& \mbox{75.00}  & \mbox{9 (75.00)}
& \mbox{100.00} & \mbox{2 (100.00)}
& \mbox{75.00}  & \mbox{41 (68.33)}
\\

\rowcolor{maRowThree}
w/o Localized Recovery
& \mbox{90.00}  & \mbox{8 (80.00)}
& \mbox{95.65}  & \mbox{19 (82.61)}
& \mbox{92.31}  & \mbox{8 (61.54)}
& \mbox{91.67}  & \mbox{10 (83.33)}
& \mbox{100.00} & \mbox{1 (50.00)}
& \mbox{93.33}  & \mbox{46 (76.67)}
\\

\rowcolor{maRowFull}
\paperbold{OmniHarness}
& \mbox{\paperbold{100.00}} & \mbox{\paperbold{9 (90.00)}}
& \mbox{\paperbold{100.00}} & \mbox{\paperbold{21 (91.30)}}
& \mbox{\paperbold{100.00}} & \mbox{\paperbold{8 (61.54)}}
& \mbox{\paperbold{100.00}} & \mbox{\paperbold{10 (83.33)}}
& \mbox{\paperbold{100.00}} & \mbox{\paperbold{2 (100.00)}}
& \mbox{\paperbold{100.00}} & \mbox{\paperbold{50 (83.33)}}
\\

\bottomrule
\end{tabular}
\end{adjustbox}
\endgroup
\end{table}

\subsubsection{Localized Image Editing on ComfyBench}

\begin{figure}[!htbp]
    \centering
    \includegraphics[width=\textwidth]{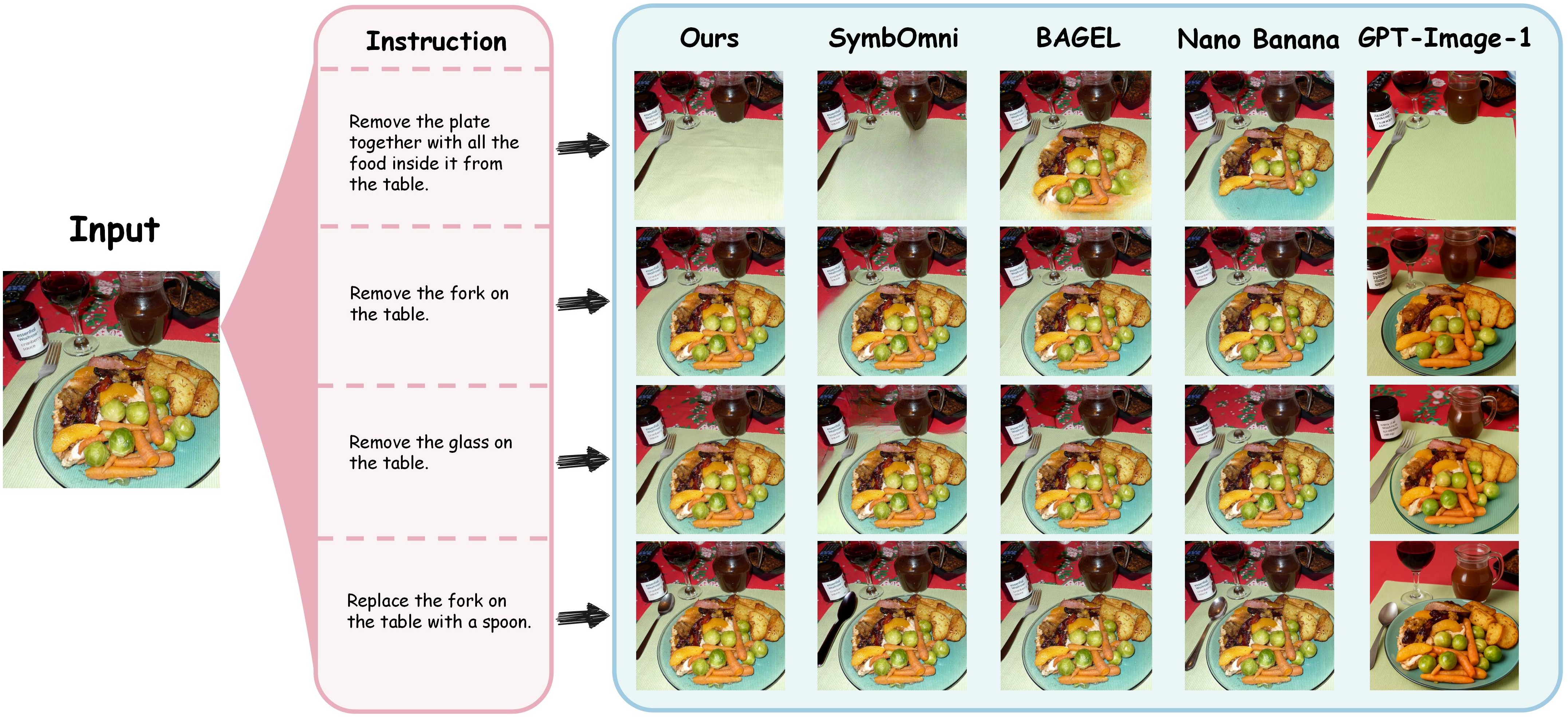}
    \caption{Qualitative comparison on localized image editing tasks from ComfyBench~\citep{comfybench}. Four independent instructions assess target localization, instruction following, and preservation of unrelated content.}
    \label{fig:10}
\end{figure}

Figure~\ref{fig:10} compares OmniHarness with SymbOmni, BAGEL, Nano Banana, and GPT-Image-1 on four independent image editing tasks from ComfyBench~\citep{comfybench}, using the same source image. The instructions require removing specific objects or replacing the fork with a spoon. OmniHarness performs the requested edits while preserving unrelated objects and the scene layout. Some competing outputs retain the target object or introduce broader appearance changes. These examples illustrate the framework's ability to localize targets and constrain edits to the requested content.

\subsubsection{GPT-Score Evaluation on Reason-Edit}

\begin{figure}[!htbp]
    \centering
    \includegraphics[width=\textwidth]{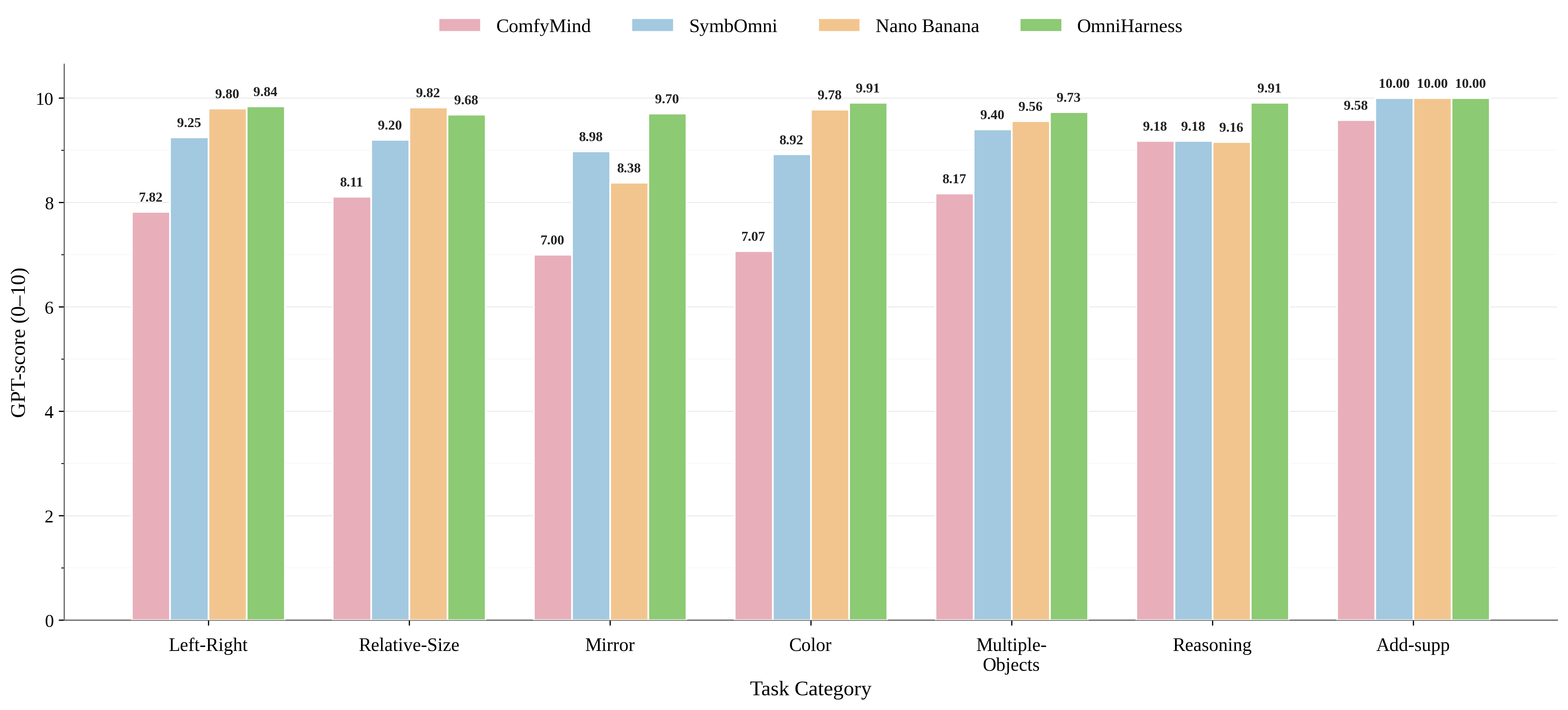}
    \caption{GPT-score comparison across seven image editing categories on Reason-Edit~\citep{smartedit}. Scores range from 0 to 10, with higher values indicating better performance.}
    \label{fig:gpt score}
\end{figure}

Figure~\ref{fig:gpt score} complements the main results with GPT-scores across seven image editing categories on Reason-Edit~\citep{smartedit}. OmniHarness ranks first in five categories and shares the highest score on Add-supp. Its largest improvements over the strongest competing scores occur in Mirror and Reasoning, with gains of 0.72 and 0.73 points, respectively. On Relative-Size, it scores 9.68, below Nano Banana's 9.82. These results support its effectiveness in interpreting spatial relations and reasoning about editing targets.

\subsection{Additional Details and Clarifications}
\label{app:additional_details}

\subsubsection{Generative Capability Space}
\label{app:generative_capability_space}

The capability space $\mathcal{G}$ contains six T2I and six I2I categories,
listed in Table~\ref{tab:app_capability_space}.
By default, self-directed inquiry starts with empty workflow and failure
libraries and runs for 50 rounds.
Each round proposes ten candidate tasks using $\mathcal{G}$,
the current scene context $c_t$, and the library state
$\mathcal{S}_t=(\mathcal{L}_t,\mathcal{F}_t)$.
Task selection follows the capability novelty and competence frontier
scores defined in the main paper.
The curriculum covers individual generation and editing capabilities
and does not explicitly practice compositional workflow tasks.
In the default setting, I2I practice uses the original ComfyBench
source images and excludes video inputs.
Benchmark task instructions, target outputs, reference workflows,
annotations, and evaluation labels are withheld.

\begin{table}[!htbp]
\caption{
Generative capability space $\mathcal{G}$ for self-directed inquiry.
}
\label{tab:app_capability_space}
\centering
\begingroup
\fontsize{8pt}{9.5pt}\selectfont
\setlength{\tabcolsep}{4pt}
\renewcommand{\arraystretch}{1.06}
\renewcommand{\tabularxcolumn}[1]{m{#1}}

\begin{tabularx}{\linewidth}{
    >{\raggedright\arraybackslash}m{0.34\linewidth}
    X
}
\toprule
\paperbold{Capability Category} & \paperbold{Description} \\
\midrule

\multicolumn{2}{l}{\textit{Text-to-Image (T2I)}} \\

Photorealistic Generation
& Generates realistic subjects and indoor or outdoor scenes. \\

Futuristic \& Fantasy Generation
& Generates futuristic, science-fiction, and fantasy content. \\

Illustration \& Comic Generation
& Generates illustrations, cartoons, comics, and visual narratives. \\

Poster \& Graphic Design
& Generates posters, covers, and other graphics with specified layouts. \\

In-Image Text Generation
& Renders specified text with controlled content, appearance, and placement. \\

Position-Constrained Generation
& Places objects or visual elements in specified image regions. \\

\midrule
\multicolumn{2}{l}{\textit{Image-to-Image (I2I)}} \\

Localized Image Editing
& Modifies selected content while preserving unrelated regions. \\

Style Transfer \& Repainting
& Changes visual style while preserving content and structure. \\

Reference-Guided Generation
& Generates content conditioned on reference pose, structure,
style, or semantics. \\

Image Super-Resolution
& Improves resolution and detail while preserving content and structure. \\

Image Outpainting
& Extends image boundaries while maintaining visual consistency. \\

Image Restoration \& Refinement
& Repairs degradation and artifacts and restores visual detail. \\

\bottomrule
\end{tabularx}
\endgroup
\end{table}

\subsubsection{Schema of the Failure Library}
\label{app:failure_library_schema}

The failure library $\mathcal{F}_t$ stores execution failure records
and corrective strategies, as specified in
Table~\ref{tab:app_failure_library}.
Failure records preserve the conditions and evidence needed for diagnosis.
Corrective strategies summarize causes, remedies, and their applicable
scope to guide subsequent planning and localized recovery.
The curator consolidates both libraries every five inquiry rounds,
merging redundant entries and refining reusable corrections.

\begin{table}[!htbp]
\caption{Schema of the failure library $\mathcal{F}_t$.}
\label{tab:app_failure_library}
\centering
\begingroup
\fontsize{8pt}{9.5pt}\selectfont
\setlength{\tabcolsep}{4pt}
\renewcommand{\arraystretch}{1.06}
\renewcommand{\tabularxcolumn}[1]{m{#1}}

\begin{tabularx}{\linewidth}{
    >{\raggedright\arraybackslash}m{0.27\linewidth}
    X
}
\toprule
\paperbold{Field} & \paperbold{Description} \\
\midrule

\multicolumn{2}{l}{\textit{Execution Failure Record}} \\

Task $\tau_t$
& Task description, modality, required capabilities,
and source image when applicable. \\

Failure Context
& Input conditions and generation constraints associated with the failure. \\

Failed Workflow $\pi_t$
& Executed graph, including nodes, parameters, connections,
and execution settings. \\

Failure Evidence $o_t$
& Execution status, source images, intermediate and final outputs,
execution feedback, and verifier judgments. \\

\midrule
\multicolumn{2}{l}{\textit{Corrective Strategies}} \\

Root Cause
& Workflow or generation factor responsible for the failure. \\

Workflow Antipattern
& Workflow structures, node configurations, or parameter settings
associated with failure. \\

Remedy
& Correction or alternative workflow strategy addressing
the identified cause. \\

Applicable Scope
& Modalities, capabilities, workflow stages, nodes, or models
to which the correction applies. \\

\bottomrule
\end{tabularx}
\endgroup
\end{table}

\subsubsection{Fields of the Current Scene Context}
\label{app:current_scene_context}

The scene context $c_t$ summarizes the current library and available
source images to guide task proposals.
Table~\ref{tab:app_scene_context} lists its fields.
The context is refreshed after execution and library consolidation.
Coverage counts describe stored workflow entries.
The separate counts $n_t(z,g)$ used for capability novelty record
all practice attempts, including unsuccessful ones.

\begin{table}[!htbp]
\caption{Fields of the current scene context $c_t$.}
\label{tab:app_scene_context}
\centering
\begingroup
\fontsize{8pt}{9.5pt}\selectfont
\setlength{\tabcolsep}{4pt}
\renewcommand{\arraystretch}{1.06}
\renewcommand{\tabularxcolumn}[1]{m{#1}}

\begin{tabularx}{\linewidth}{
    >{\raggedright\arraybackslash}m{0.28\linewidth}
    X
}
\toprule
\paperbold{Field} & \paperbold{Description} \\
\midrule

Modality Coverage
& Numbers of stored T2I and I2I workflows. \\

Capability Coverage
& Number of stored workflows associated with each capability category. \\

Reliability Distribution
& Numbers of \emph{Provisional}, \emph{Validated},
and \emph{Suspended} workflows. \\

Source Image Pool $\mathcal{X}$
& Images available for I2I practice, excluding video inputs. \\

Source Image Metadata
& Image file names, resolutions, and formats used for source selection. \\

\bottomrule
\end{tabularx}
\endgroup
\end{table}

\subsubsection{Overall Execution Procedure}
\label{app:overall_execution_procedure}

Algorithm~\ref{alg:omniharness} summarizes inquiry and the default
downstream evaluation procedure.
Here, $R$ denotes retrieved policies, $\rho$ the execution plan,
$\pi$ the executable workflow, and $o$ the execution record.
The operation $\operatorname{Run}$ retrieves policies, constructs
a verified plan, compiles its Code-as-Policy program, and executes
the workflow with verification and localized recovery.
If no applicable policy is retrieved, planning uses the ComfyUI
knowledge graph to construct a workflow.
Recovery uses at most four retries and preserves verified, unaffected steps.
The operation $\operatorname{Update}$ abstracts verified executions,
updates policy usage and success counts, records failures and corrections,
and refreshes the context.
Practice-attempt counts are maintained separately and incremented
after every inquiry execution, including failures.
The inquiry snapshot is saved before downstream evaluation.
Both libraries are restored from this snapshot before each benchmark
and updated within that benchmark.
Frozen-snapshot evaluations and the ablation without online policy
updates omit these updates.
Exported snapshots used by external agents remain frozen.

\begin{algorithm}[!htbp]
\caption{OmniHarness inquiry and downstream evaluation}
\label{alg:omniharness}
\small
\begin{algorithmic}[1]

\Require Capability space $\mathcal{G}$, node knowledge $\mathcal{K}$,
source images $\mathcal{X}$,
\Statex \hspace{\algorithmicindent}
inquiry budget $N$ with default 50,
and benchmark task streams $\{\mathcal{D}_b\}$

\Ensure Frozen inquiry snapshot $\mathcal{K}_{\mathrm{inquiry}}$
and benchmark libraries $\{\mathcal{S}^{(b)}\}$

\State $\mathcal{S}_1\gets(\emptyset,\emptyset)$,
$c_1\gets\operatorname{InitContext}(\mathcal{X})$
\State Initialize all practice-attempt counts to zero

\For{$t=1,\ldots,N$}

    \State $\mathcal{T}_t\gets
    \operatorname{Propose}(\mathcal{G},c_t,\mathcal{S}_t,10)$

    \State $\tau_t\gets
    \operatorname*{arg\,max}_{\tau\in\mathcal{T}_t}
    \mathcal{N}(\tau)\mathcal{C}(\tau)$

    \State $(\pi_t,o_t)\gets
    \operatorname{Run}(\tau_t,c_t,\mathcal{K},\mathcal{S}_t)$

    \State Increment attempt counts for each context
    and capability pair in $\tau_t$

    \State $(c_{t+1},\mathcal{S}_{t+1})\gets
    \operatorname{Update}(c_t,\mathcal{S}_t,\tau_t,\pi_t,o_t)$

    \If{$t\bmod 5=0$}
        \State $\mathcal{S}_{t+1}\gets
        \operatorname{Consolidate}(\mathcal{S}_{t+1})$

        \State $c_{t+1}\gets
        \operatorname{RefreshContext}(\mathcal{X},\mathcal{S}_{t+1})$
    \EndIf

\EndFor

\State $\mathcal{K}_{\mathrm{inquiry}}\gets
\operatorname{Snapshot}(\mathcal{S}_{N+1})$

\For{each benchmark $\mathcal{D}_b$}

    \State $\mathcal{S}\gets
    \operatorname{Copy}(\mathcal{K}_{\mathrm{inquiry}})$

    \State $c\gets
    \operatorname{RefreshContext}(\mathcal{X},\mathcal{S})$

    \For{each task $\tau$ in $\mathcal{D}_b$}

        \State $(\pi,o)\gets
        \operatorname{Run}(\tau,c,\mathcal{K},\mathcal{S})$

        \State $(c,\mathcal{S})\gets
        \operatorname{Update}(c,\mathcal{S},\tau,\pi,o)$

    \EndFor

    \State $\mathcal{S}^{(b)}\gets\mathcal{S}$

\EndFor

\State \Return
$\mathcal{K}_{\mathrm{inquiry}},\{\mathcal{S}^{(b)}\}$

\Statex

\Function{Run}{
$\tau,c,\mathcal{K},\mathcal{S}=(\mathcal{L},\mathcal{F})$
}

    \State $R\gets
    \operatorname{SymbolicRetrieve}(\tau,\mathcal{L})$

    \State $\rho\gets
    \operatorname{Plan}(\tau,c,\mathcal{K},R,\mathcal{F})$

    \State $\rho\gets
    \operatorname{VerifyAndRefinePlan}(\tau,\rho,\mathcal{K})$

    \State $\pi\gets
    \operatorname{Compile}
    (\operatorname{Write}(\tau,\rho,\mathcal{K},R))$

    \State $(\pi,o)\gets
    \operatorname{ExecuteVerifyRecover}(\tau,\rho,\pi,\mathcal{F})$

    \State \Return $(\pi,o)$

\EndFunction

\end{algorithmic}
\end{algorithm}

\subsubsection{Symbolic Policy Component Ablation Settings}
\label{app:policy_component_ablation_settings}

Table~\ref{tab:ablation_comfybench} evaluates four policy components
and two learning stages.
The variants follow the reasoning model, generative model,
and retry settings in the main evaluation protocol.
The practice budget is 50 rounds except when self-directed inquiry
is removed.
Each ablation disables the specified component while retaining
the remaining mechanisms.

\textit{w/o Instance-Agnostic Abstraction} stores verified executions
as instance-specific parameterized workflows, preserving workflow memory
while disabling input abstraction.
\textit{w/o Applicability Conditions} removes the Preconditions field
$P_w$ while retaining Expected Effects $E_w$ and Dependencies $D_w$.
\textit{w/o Adaptive Instantiation} disables task-specific input binding
and parameter adaptation during policy reuse.
\textit{w/o Workflow Composition} prevents combining multiple policies
while retaining retrieval and adaptation of individual policies.
\textit{w/o Self-Directed Inquiry} omits practice and starts downstream
execution without the inquiry-derived libraries, while retaining
subsequent updates to both libraries.
\textit{w/o Online Policy Updates} uses the libraries acquired through
inquiry but keeps both fixed during downstream evaluation.
Instantiation, adaptation, composition, verification, and recovery
remain active in this last variant.

\subsubsection{Symbolic Retrieval Ablation Settings}
\label{app:symbolic_retrieval_ablation_settings}

Table~\ref{tab:symbolic_retrieval_ablation} independently enables
or disables symbolic retrieval and reliability-tier prioritization.
Symbolic retrieval selects policies through staged metadata matching
before accessing complete workflow templates.
When disabled, retrieval operates directly on parameterized workflows.
With reliability tiers enabled, retrieval prioritizes
\emph{Validated} policies and deprioritizes \emph{Suspended} policies.
Disabling this mechanism removes the tier-based preference from retrieval.
The four configurations evaluate the individual and combined effects
of these mechanisms under the default execution settings.

Agents marked with $\dagger$ are initialized with the same policy
library acquired through OmniHarness's self-directed inquiry.
The exported snapshot contains workflow templates, failure evidence,
and corrective strategies.
Entries are mapped to each host's native knowledge interface while
preserving its model weights and control flow.
ComfyAgent uses Reference memory, ComfyMind uses Semantic Workflow
Interface modules, and SymbOmni uses Symbolic Concepts.
The exported snapshot remains fixed, and each host retrieves and applies
its entries through its existing execution procedures.

Tokens and Time report mean downstream agent-side costs per task
within each subset.
They cover planning, workflow synthesis, verification, and recovery.
Inquiry costs are excluded, and Time also excludes visual generation
model execution.

\subsubsection{Policy Corruption and Verifier Evaluation Settings}
\label{app:verifier_corruption_settings}

Figure~\ref{fig:verifier_reliability} evaluates four corruption conditions
starting from the 43-policy snapshot acquired through 50 inquiry rounds.
Table~\ref{tab:policy_corruption_settings} specifies their composition.
Description interventions change Workflow Description $d_w$
so that the stated function conflicts with the unchanged
executable workflow.
Workflow interventions introduce one fault in an input binding,
node connection, or key parameter.
Tier interventions replace an entry with a defective historical
policy version previously marked \emph{Suspended} and assign it
\emph{Validated} status.
The full intervention pool uses eleven of the fourteen initially
validated policies for description and workflow corruption,
selected to cover different modalities and capabilities.
Two additional entries receive the tier interventions.

\begin{table}[!htbp]
\caption{
Composition of the policy corruption conditions.
Each intervention replaces one entry,
preserving the initial library size of 43.
}
\label{tab:policy_corruption_settings}
\centering
\begingroup
\fontsize{8pt}{9.5pt}\selectfont
\setlength{\tabcolsep}{4pt}
\renewcommand{\arraystretch}{1.06}

\begin{tabularx}{\linewidth}{
    l
    *{3}{>{\centering\arraybackslash}X}
}
\toprule
\paperbold{Corrupted Policies}
& \paperbold{Description}
& \paperbold{Workflow}
& \paperbold{Reliability Tier}
\\
\midrule
$0/43$  & 0 & 0 & 0 \\
$4/43$  & 1 & 1 & 2 \\
$9/43$  & 3 & 4 & 2 \\
$13/43$ & 5 & 6 & 2 \\
\bottomrule
\end{tabularx}
\endgroup
\end{table}

The corrupted sets are nested, and shared entries use the same
modified versions across conditions.
The same two tier interventions appear in every nonzero condition.
Each condition is evaluated on all 200 ComfyBench tasks under
five randomized task orders.
For each order, the conditions share model settings, the initial
failure library, task sequence, execution budget, and update rules.
Each run starts from a fresh copy of its designated initial library.
Verification, recovery, and online policy updates remain active,
so the library may change during execution.

Two reviewers independently assess execution outcomes against
task requirements, logs, and output artifacts.
A third reviewer resolves disagreements.
Reviewers do not see verifier decisions or corruption assignments.
Failed attempts and repaired executions are recorded separately.
Official benchmark evaluations and human judgments are used only
for analysis and do not update the running policy library.

At each completed-task index, \emph{Verifier} reports the cumulative
proportion of policy reuse attempts judged successful during evaluation.
Its numerator counts successful policy reuse attempts recorded by
success-counter increments, excluding successes accumulated
before evaluation.
\emph{ComfyBench} reports the cumulative proportion of completed tasks
resolved by the official evaluator.
\emph{Human audit} reports the proportion of audited verifier decisions
that agree with human judgments, including correctly accepted successes
and correctly rejected failures.
Their denominators are policy reuse attempts, completed tasks,
and audited verifier decisions, respectively.
Curves and shaded regions show the mean and standard deviation
of each cumulative rate across the five task orders.

\subsection{System Prompts}
\label{sec:omniharness_prompts}

The following tables reproduce the two developer instructions and eight
task-input templates used by the implementation. Braced Python expressions
denote values inserted at runtime rather than literal prompt text.
Candidate validation and scoring, context retrieval, structural plan and
workflow checks, compilation, execution, evidence aggregation, symbolic
policy input abstraction, library updates, consolidation, and snapshot export
are implemented programmatically without separate model prompts.
Candidate generation reuses a proposer thread across inquiry iterations.
Execution roles share a task-scoped thread, while localized repair uses a fresh thread
with the same generation-team developer instructions.

\begingroup
\setlength{\intextsep}{4.0pt plus .6pt minus .6pt}
\setlength{\parskip}{0pt}
\captionsetup[table]{skip=2pt}
\begin{table}[H]
\centering
\caption{Shared developer instructions for self-directed inquiry.}
\label{tab:prompt_task_proposer}
\rmfamily\fontsize{9pt}{10.8pt}\selectfont
\fcolorbox{black}{OmniPromptPink}{%
\begin{tabular}{@{}p{\dimexpr\linewidth-2\fboxsep-2\fboxrule\relax}@{}}
\paperbold{Self-Directed Inquiry Task Proposer} \\
\hline
You are the Self-Directed Inquiry Task Proposer inside OmniHarness. Generate
practice tasks before downstream objectives are specified. Each task must use
the supplied capability space, be feasible with current resources, differ from
recent tasks, and be independently verifiable. Use failure evidence and
corrective strategies to avoid known antipatterns. For I2I, use exactly one
supplied image ID and state preservation constraints. List
applicable\_workflow\_ids only when the workflow's preconditions hold for the
proposed task and source image. Exclude suspended workflows; return an empty
list when none apply. Use only the supplied context and library. Do not
inspect downstream task instructions, reference workflows, target outputs, or
benchmark annotations. Return only the requested JSON object.
\\
\end{tabular}%
}
\end{table}

\begin{table}[H]
\centering
\caption{Task-input template for candidate task generation.}
\label{tab:prompt_candidate_generation}
\rmfamily\fontsize{9pt}{10.8pt}\selectfont
\fcolorbox{black}{OmniPromptPink}{%
\begin{tabular}{@{}p{\dimexpr\linewidth-2\fboxsep-2\fboxrule\relax}@{}}
\paperbold{Candidate Task Generation} \\
\hline
Generate exactly \texttt{\{candidate\_count\}} candidate tasks. Return \{"candidates":
[CandidateTask, ...]\}. Each CandidateTask contains task\_id, description,
modality (T2I or I2I), capability\_categories (one to three exact names),
source\_image\_id (null for T2I), generation\_constraints,
preservation\_constraints, success\_criteria, required\_resources,
task\_signature, proposal\_rationale, safety\_flags (empty for acceptable tasks),
and applicable\_workflow\_ids.
\par\vspace{1pt}
\paperbold{State:}\enspace\texttt{\{json.dumps(compact, ensure\_ascii=False, indent=2)\}}
\\
\end{tabular}%
}
\end{table}

\begin{table}[H]
\centering
\caption{Shared developer instructions for feedback-guided execution.}
\label{tab:prompt_generation_team}
\rmfamily\fontsize{9pt}{10.8pt}\selectfont
\fcolorbox{black}{OmniPromptBlue}{%
\begin{tabular}{@{}p{\dimexpr\linewidth-2\fboxsep-2\fboxrule\relax}@{}}
\paperbold{OmniHarness Generation Team} \\
\hline
You are the OmniHarness Generation Team. Operate through explicit roles:
Planner, Plan Verifier, Workflow Writer, Executor observer, Step Verifier,
Goal Verifier, and Failure Diagnoser. Retrieved workflows are symbolic policy
templates for task families; retrieved failures contain evidence and
corrective strategies. Bind the current task's inputs and adapt or compose
templates before execution. Respect preconditions, dependencies and
component-only retrieval roles. Image components require video-specific
processing for video outputs. Return only the requested JSON. Never edit
files, execute ComfyUI, change retry budgets, update memory, or declare
technical success. Code-as-Policy is a safe DSL with one assignment and one
allow-listed ComfyUI node call per line, keyword arguments only, literals or
earlier variables as inputs, and no imports, attributes, control flow, nested
calls, or arbitrary Python.
\\
\end{tabular}%
}
\end{table}

\begin{table}[H]
\centering
\caption{Task-input template for planning.}
\label{tab:prompt_planner}
\rmfamily\fontsize{9pt}{10.8pt}\selectfont
\fcolorbox{black}{OmniPromptBlue}{%
\begin{tabular}{@{}p{\dimexpr\linewidth-2\fboxsep-2\fboxrule\relax}@{}}
\paperbold{Planner} \\
\hline
Create a dependency-consistent ordered plan with explicit verification for
every step. For each selected workflow, check its input roles, preconditions,
and dependencies against the request and planned intermediate inputs. In
task\_analysis.policy\_applicability return
\{workflow\_id:\{satisfied:true,evidence:[specific condition checks]\}\}. Select a
policy only when its conditions can be met by the plan. Choose reuse, adapt,
compose, or build. Return \{plan\_id, task\_analysis, strategy,
selected\_workflow\_ids, steps:[\{step\_id, objective, inputs, outputs,
required\_nodes, verification:\{type,criteria\}, expected\_output\_nodes,
fallback\}], rationale\}.
\par\vspace{1pt}
\paperbold{Request:}\enspace\texttt{\{\_json(request)\}}
\par\vspace{1pt}
\paperbold{Context:}\enspace\texttt{\{\_json(context.compact(include\_prompts=False))\}}
\\
\end{tabular}%
}
\end{table}

\begin{table}[H]
\centering
\caption{Conditional task-input template for plan repair.}
\label{tab:prompt_plan_repair}
\rmfamily\fontsize{9pt}{10.8pt}\selectfont
\fcolorbox{black}{OmniPromptBlue}{%
\begin{tabular}{@{}p{\dimexpr\linewidth-2\fboxsep-2\fboxrule\relax}@{}}
\paperbold{Plan Repair} \\
\hline
Repair this plan and return the same schema.
\par\vspace{1pt}
\makebox[.48\linewidth][l]{\paperbold{Request:}\enspace\texttt{\{\_json(request)\}}}%
\paperbold{Plan:}\enspace\texttt{\{\_json(plan)\}}
\par\vspace{1pt}
\paperbold{Issues:}\enspace\texttt{\{\_json(issues)\}}
\par\vspace{1pt}
\paperbold{Context:}\enspace\texttt{\{\_json(context.compact(include\_prompts=False))\}}
\\
\end{tabular}%
}
\end{table}

\begin{table}[H]
\centering
\caption{Task-input template for workflow writing and policy adaptation.}
\label{tab:prompt_workflow_writer}
\rmfamily\fontsize{9pt}{10.8pt}\selectfont
\fcolorbox{black}{OmniPromptBlue}{%
\begin{tabular}{@{}p{\dimexpr\linewidth-2\fboxsep-2\fboxrule\relax}@{}}
\paperbold{Workflow Writer} \\
\hline
Convert the verified plan to Code-as-Policy. Use \$SOURCE\_MEDIA\_0 placeholders
for media inputs; \$SOURCE\_IMAGE\_0 remains accepted for compatibility.
Retrieved graphs are templates: replace every \$POLICY\_* placeholder with
task-specific text, numeric parameters, output names, or \$SOURCE\_MEDIA\_n
bindings. Check input-role preconditions and compose component\_only image
policies with video-specific nodes. Set \_meta\_title on every Code-as-Policy
node to its owning plan step\_id. Preserve existing assignment variable names
during localized repairs. Each step must emit an inspectable intermediate
output before dependent steps execute; expected\_output\_nodes should identify
actual numeric node IDs of those output nodes, matching their one-based
Code-as-Policy statement positions. A step title refers only to that step's
output nodes; it does not require observing every internal node. Every node
must contribute to at least one declared step output. If no applicable policy
is available, construct a workflow from node knowledge. Describe the reusable
task-family procedure, not the particular objects, prompt, or source filename
of this execution, in workflow\_name, description, preconditions and
expected\_effects. Return \{ir\_code, workflow\_name, description, preconditions,
expected\_effects, dependencies:\{Nodes,Models,Parameters\}\}.
\par\vspace{1pt}
\makebox[.48\linewidth][l]{\paperbold{Request:}\enspace\texttt{\{\_json(request)\}}}%
\paperbold{Plan:}\enspace\texttt{\{\_json(plan)\}}
\par\vspace{1pt}
\paperbold{Context:}\enspace\texttt{\{\_json(context.compact(include\_prompts=True))\}}
\\
\end{tabular}%
}
\end{table}

\begin{table}[H]
\centering
\caption{Conditional task-input template for step verification.}
\label{tab:prompt_step_verifier}
\rmfamily\fontsize{9pt}{10.8pt}\selectfont
\fcolorbox{black}{OmniPromptBlue}{%
\begin{tabular}{@{}p{\dimexpr\linewidth-2\fboxsep-2\fboxrule\relax}@{}}
\paperbold{Step Verifier} \\
\hline
Inspect the supplied step output artifacts using image or video tools. Verify
each listed criterion against these artifacts and the request. Do not infer an
intermediate effect from the final result. Return
\{criteria:[\{criterion,status:pass\textbar{}\allowbreak{}fail\textbar{}\allowbreak{}uncertain,evidence:[string]\}]\}. Copy
each criterion exactly and use uncertain if its evidence cannot be inspected.
\par\vspace{1pt}
\makebox[.48\linewidth][l]{\paperbold{Request:}\enspace\texttt{\{\_json(request)\}}}%
\paperbold{Step:}\enspace\texttt{\{\_json(step)\}}
\par\vspace{1pt}
\paperbold{Step outputs:}\enspace\texttt{\{\_json([item.path for item in result.outputs])\}}
\\
\end{tabular}%
}
\end{table}

\begin{table}[H]
\centering
\caption{Conditional task-input template for goal verification.}
\label{tab:prompt_goal_verifier}
\rmfamily\fontsize{9pt}{10.8pt}\selectfont
\fcolorbox{black}{OmniPromptBlue}{%
\begin{tabular}{@{}p{\dimexpr\linewidth-2\fboxsep-2\fboxrule\relax}@{}}
\paperbold{Goal Verifier} \\
\hline
Inspect every output path with the appropriate image or video tools. Evaluate
every success and preservation criterion. Use uncertain when evidence is
insufficient. Return \{status:pass\textbar{}\allowbreak{}fail\textbar{}\allowbreak{}uncertain, verifier, scores, evidence,
failed\_criteria\}.
\par\vspace{1pt}
\makebox[.48\linewidth][l]{\paperbold{Request:}\enspace\texttt{\{\_json(request)\}}}%
\paperbold{Plan:}\enspace\texttt{\{\_json(plan)\}}
\par\vspace{1pt}
\paperbold{Outputs:}\enspace\texttt{\{\_json([item.path for item in result.outputs])\}}
\\
\end{tabular}%
}
\end{table}

\begin{table}[H]
\centering
\caption{Conditional task-input template for failure diagnosis.}
\label{tab:prompt_failure_diagnoser}
\rmfamily\fontsize{9pt}{10.8pt}\selectfont
\fcolorbox{black}{OmniPromptBlue}{%
\begin{tabular}{@{}p{\dimexpr\linewidth-2\fboxsep-2\fboxrule\relax}@{}}
\paperbold{Failure Diagnoser} \\
\hline
Identify the smallest failed stage. Return
\{failure\_class:planning\_failure\textbar{}\allowbreak{}compilation\_failure\textbar{}\allowbreak{}infrastructure\_failure\textbar{}\allowbreak{}execution\_failure\textbar{}\allowbreak{}verifier\_failure\textbar{}\allowbreak{}semantic\_failure,
failed\_stage, root\_cause, workflow\_antipattern, remedy, applicable\_scope,
retry\_scope\}.
\par\vspace{1pt}
\makebox[.48\linewidth][l]{\paperbold{Request:}\enspace\texttt{\{\_json(request)\}}}%
\paperbold{Plan:}\enspace\texttt{\{\_json(plan)\}}
\par\vspace{1pt}
\makebox[.48\linewidth][l]{\paperbold{Draft:}\enspace\texttt{\{\_json(draft)\}}}%
\paperbold{Error:}\enspace\texttt{\{\_json(error)\}}
\par\vspace{1pt}
\paperbold{Verification:}\enspace\texttt{\{\_json(verification)\}}
\\
\end{tabular}%
}
\end{table}

\begin{table}[H]
\centering
\caption{Conditional task-input template for localized repair.}
\label{tab:prompt_failure_subagent}
\rmfamily\fontsize{9pt}{10.8pt}\selectfont
\fcolorbox{black}{OmniPromptBlue}{%
\begin{tabular}{@{}p{\dimexpr\linewidth-2\fboxsep-2\fboxrule\relax}@{}}
\paperbold{Failure Subagent} \\
\hline
Act as an isolated Failure Subagent. Repair only the diagnosed step and
preserve unrelated workflow logic. Preserve the assignment variable names and
node inputs outside that step and its dependent downstream nodes. Return
\{failed\_step\_id, root\_cause, subworkflow\_ir, revised\_ir\_code, reusable\_name,
rationale\}. The full revised\_ir\_code must integrate the localized component.
\par\vspace{1pt}
\makebox[.48\linewidth][l]{\paperbold{Request:}\enspace\texttt{\{\_json(request)\}}}%
\paperbold{Plan:}\enspace\texttt{\{\_json(plan)\}}
\par\vspace{1pt}
\makebox[.48\linewidth][l]{\paperbold{Draft:}\enspace\texttt{\{\_json(draft)\}}}%
\paperbold{Diagnosis:}\enspace\texttt{\{\_json(diagnosis)\}}
\par\vspace{1pt}
\paperbold{Recovery context:}\enspace\texttt{\{\_json(context.compact(include\_prompts=True))\}}
\\
\end{tabular}%
}
\end{table}
\endgroup

\end{document}